%% file: main.tex
\documentclass[runningheads]{llncs}

\usepackage{eccv}

\usepackage{eccvabbrv}

\usepackage{graphicx}
\usepackage{booktabs}
\usepackage{multirow}
\usepackage{pgfplots}
\usepgfplotslibrary{groupplots}
\usetikzlibrary{calc, positioning, arrows.meta, decorations.pathreplacing}
\pgfplotsset{compat=1.18}

\usepackage[accsupp]{axessibility}  %

\usepackage[table]{xcolor}

\usepackage{hyperref}

\usepackage{orcidlink}

\newcommand{\myparagraph}[1]{\medskip\noindent\textbf{#1}\hspace{0.2em plus 0.3em minus 0.2em}}

\newcommand{\methodlong}{Amortized Hybrid Attribution\xspace}
\newcommand{\methodabbrev}{AHA\xspace}

\begin{document}

\title{Learn to Rank: Visual Attribution by Learning Importance Ranking}

\author{David Schinagl\inst{1}\orcidlink{0009-0009-5643-4143} \and
Christian Fruhwirth-Reisinger\inst{1,2}\orcidlink{0000-0001-9930-9822} \and
Alexander Prutsch\inst{1}\orcidlink{0009-0001-3577-2794} \and
Samuel Schulter\inst{3}\thanks{This work is independent of the author’s employment at Amazon.}\orcidlink{0000-0002-7341-5907} \and
Horst Possegger\inst{1}\orcidlink{0000-0002-5427-9938}}

\authorrunning{D.~Schinagl et al.}

\institute{
    Institute of Visual Computing (IVC), Graz University of Technology, Graz, Austria \email{\{david.schinagl,reisinger,alexander.prutsch,possegger\}@tugraz.at}
    \and
    Nomadic AI, USA
    \and 
    Amazon, USA
}

\maketitle

\input{sec/0_abstract}

\input{sec/1_introduction}

\input{sec/2_related_work}

\input{sec/3_method}

\input{sec/4_experiments}
\input{sec/5_limitations_conclusion}

\section*{Acknowledgements}
This work received funding from the AI Ecosystems programme of the Republic of Austria, Federal Ministry of Innovation, Mobility and Infrastructure (BMIMI) under the project DOMINO (923575). The Austrian Research Promotion Agency (FFG) has been authorised for the programme management.
\bibliographystyle{splncs04}
\bibliography{main}

\newpage
\appendix

\input{sec/A_supp}

\end{document}

%% file: sec/0_abstract.tex
\begin{abstract}

Interpreting the decisions of complex computer vision models is crucial to establish trust and accountability, especially in safety-critical domains.
An established approach to interpretability is generating visual attribution maps that highlight regions of the input most relevant to the model's prediction.
However, existing methods face a three-way trade-off. 
Propagation-based approaches are efficient, but they can be biased and architecture-specific. 
Meanwhile, perturbation-based methods are causally grounded, yet they are expensive and for vision transformers often yield coarse, patch-level explanations.
Learning-based explainers are fast but usually optimize surrogate objectives or distill from heuristic teachers.
We propose a learning scheme that instead optimizes deletion and insertion metrics directly. 
Since these metrics depend on non-differentiable sorting and ranking, we frame them as permutation learning and replace the hard sorting with a differentiable relaxation using Gumbel-Sinkhorn. 
This enables end-to-end training through attribution-guided perturbations of the target model.
During inference, our method produces dense, pixel-level attributions in a single forward pass with optional, few-step gradient refinement.
Our experiments demonstrate consistent quantitative improvements and sharper, boundary-aligned explanations, particularly for transformer-based vision models.
Code and pretrained models are available at \url{https://github.com/dschinagl/AHA}.
\keywords{Visual Attribution \and Amortized Explanation \and Permutation}

\end{abstract}

%% file: sec/1_introduction.tex
\section{Introduction}
\label{sec:introduction}

Over the past decade, deep neural networks~\cite{classif:vit:dosovitskiy21, model:resnet:he16, classif:swin:liu21, model:convnext:liu22} have powered the vast majority of vision models.  
The growing complexity of these inherently opaque models, however, makes interpreting their decisions difficult.  Especially in safety-critical domains such as healthcare or autonomous driving, explaining \textit{why} a model predicts a certain outcome can be as crucial as the accuracy of the prediction itself.  
A well-established approach for explainability in computer vision is the creation of
visual attribution maps, which highlight the input regions most relevant to a model's prediction~\cite{expl:grad:deconvnet:zeiler14, expl:act:gradcam:selvaraju17, expl:grad:vanilla:simonyan13, expl:pert:rise:petsiuk18, expl:vit:gam:chefer21}. 
Such attribution maps provide an intuitive, human-interpretable way to understand model behavior and can help identify potential biases or failure modes~\cite{other:applic0:boyd22, other:applic1:anders22}.

The majority of attribution methods are \textit{post-hoc}, \ie they explain already trained models without modifying their architecture.
These can be broadly categorized into three groups:
\textit{Propagation-based} approaches which rely on propagating internal signals through the target model, such as gradients~\cite{expl:grad:vanilla:simonyan13, expl:grad:deconvnet:zeiler14, expl:grad:integratedgradients:sundararajan17}, activations~\cite{expl:act:cam:zhou16, expl:act:gradcam:selvaraju17, expl:act:scorecam:wang20} or attention weights~\cite{expl:vit:rollout:abnar20, expl:vit:gam:chefer21, expl:vit:transformerinterpretability:chefer21} in transformer-based vision models. 
They are very efficient and align with the model's internal computations but are architecture-specific, can bias toward low-level features like edges~\cite{sanity:saliencymaps:adebayo18, sanity:modifiedbp:sixt20} and may under-represent non-attention pathways in transformer models.
In contrast, \textit{perturbation-based} methods~\cite{expl:pert:rise:petsiuk18, expl:pert:lime:ribeiro16, expl:pert:shap:lundberg17, expl:pert:meaningfulpert:fong17} infer feature relevance from changes in the target model's output when parts of the input are modified.
They provide more causally grounded attributions, as they directly measure the effects of input changes, including higher-order interactions. 
However, they are computationally demanding, requiring many model evaluations to produce a single attribution map.
For transformer-based vision models, they usually operate on the native patch representation~\cite{expl:vit:mda:walker25, expl:vit:vitcx:xie23, expl:vit:tis:englebert23} yielding patch-level explanations that must be upsampled to pixel space, resulting in coarse, spatially imprecise maps, as shown in \Cref{fig:intro}.
The third and least explored category, \textit{learning-based} (amortized) explainers shift the computational burden to an offline training phase by learning a separate explainer model, which is then able to generate attribution maps very efficiently during inference~\cite{expl:amort:stochamort:covert24}.
However, the current training strategies typically rely on either supervision from other attribution methods~\cite{expl:amort:learn2explain:situ21, expl:amort:fastshap:jethani22} and may adopt their biases, or on self-supervised surrogate objectives that are based on restrictive assumptions like a binary signal to noise decomposition of features~\cite{expl:amort:l2x:chen18, expl:amort:vert:bhalla23}.

Motivated by these limitations, it is natural to ask whether we can avoid attribution heuristics and instead \textit{optimize attribution quality directly}.
Common and widely-used attribution metrics are the \textit{Deletion} and \textit{Insertion} metrics~\cite{expl:pert:rise:petsiuk18}.
They quantify the change in a model's output when features are removed or added back in order of importance.
Recent metric-driven approaches~\cite{expl:vit:lessismore:chen24, expl:vit:mda:walker25} for vision transformers pursue this direction, but do so via a perturbation-based, per-sample optimization on the patch representation.
Consequently, they retain the high computational cost and cannot refine the coarse spatial resolution.

In this work, we propose \emph{\methodlong (\methodabbrev)}: a learning-based approach that bridges these families of attribution methods by integrating their complementary strengths into a unified framework.
We retain the causal grounding of perturbation-based methods by learning from the target model’s outputs under attribution-guided perturbations.
Since we couple explainer and target model through end-to-end backpropagation, we inherit the model-consistent attribution characteristic of propagation-based approaches.
We realize this by directly optimizing the \textit{Deletion} and \textit{Insertion} metrics.
To overcome the non-differentiability of these metrics, caused by the sorting operations they depend on, we leverage the observation that they depend solely on the order of pixel importance rather than their exact values, and treat them as permutation problems.
By replacing the hard sorting and ranking operations with smooth approximations, we enable end-to-end training of the explainer model.
At inference, we preserve the amortized efficiency of learned explainers (single forward pass), but we also keep the door open to the propagation-based regime: 
we can optionally perform a few additional backward steps through the target model to locally refine the explainer’s output on a specific sample.
This effectively combines a learned, metric-aligned prior with a cheap, model-specific propagation signal. 
Due to the end-to-end optimization our explainer is able to generate dense attributions that align closely with object boundaries and visually meaningful structures, even for transformer-based vision models, as shown in \Cref{fig:intro}.\\

\input{figures/intro/fig.tex}

\noindent
Our contributions can be summarized as follows:
{\sloppy
\begin{itemize}
    \item We propose a learning-based attribution framework that bridges perturbation- and propagation-based explainability.
    \item Our amortized explainer directly optimizes differentiable surrogates of \emph{Deletion} and \emph{Insertion}, producing dense pixel-space attributions also for transformer-based vision models.
    \item We formulate metric optimization as a permutation learning problem and enable end-to-end training via a Gumbel-Sinkhorn relaxation of sorting and top-$k$ selection.
    \item To reduce the computational burden and mitigate pixel-level shortcut artifacts, the training procedure leverages region-level permutation, perturbation-step sampling, and grid augmentation.
    \item At inference, the method retains single-pass efficiency and optionally adds a few backprop refinement steps for sample-specific adjustments.
\end{itemize}}

%% file: figures/intro/fig.tex
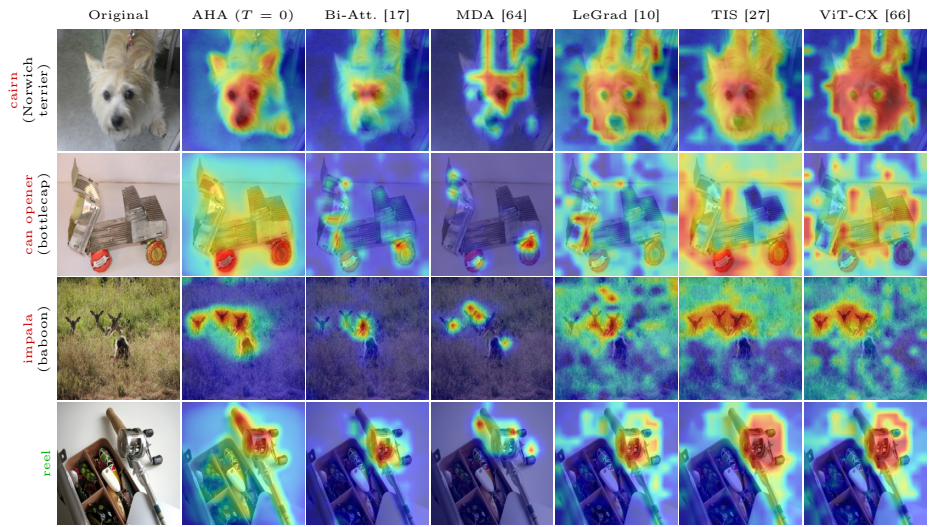
\begin{figure}[tb]
    \centering 
        \input{figures/intro/source.tex}
        \caption{
        \textbf{Qualitative comparison of attribution maps} produced by our method (\methodabbrev) and recent ViT attribution methods for a ViT-B/16 classifier~\cite{classif:vit:dosovitskiy21} on ImageNet~\cite{dataset:imagenet:deng09} samples.
        The \textcolor{green!70!black}{green} / \textcolor{red!80!black}{red} labels mark correct / incorrect predictions (ground truth in parentheses).
        While existing approaches operate on the patch representation and produce coarse, blocky attribution maps, our method generates dense, pixel-level attributions that align closely with object boundaries.
        Note that in this case no additional backprop refinement steps were applied at inference (denoted $T=0$).
    }
    \label{fig:intro}
\end{figure}

%% file: figures/intro/source.tex
\newlength{\introcolsep}
\setlength{\introcolsep}{1pt}
\newlength{\introrowsep}
\setlength{\introrowsep}{1pt}
\newlength{\introhdrsep}
\setlength{\introhdrsep}{3pt}
\newlength{\introimgw}
\setlength{\introimgw}{\dimexpr(\linewidth - 6\introcolsep)/7\relax}

\resizebox{\linewidth}{!}{%
\begin{tikzpicture}[
    every node/.style={inner sep=0pt, outer sep=0pt},
]
  \node[anchor=south, font=\tiny, text width=\introimgw, align=center] (h0) at (0*\introimgw + 0*\introcolsep + 0.5*\introimgw, \introhdrsep) {Original\vphantom{[0]}};
  \node[anchor=south, font=\tiny, text width=\introimgw, align=center] (h1) at (1*\introimgw + 1*\introcolsep + 0.5*\introimgw, \introhdrsep) {\methodabbrev~($T=0$)\vphantom{[0]}};
  \node[anchor=south, font=\tiny, text width=\introimgw, align=center] (h2) at (2*\introimgw + 2*\introcolsep + 0.5*\introimgw, \introhdrsep) {Bi-Att.~\cite{expl:vit:bidirectionalatt:chen23}};
  \node[anchor=south, font=\tiny, text width=\introimgw, align=center] (h3) at (3*\introimgw + 3*\introcolsep + 0.5*\introimgw, \introhdrsep) {MDA~\cite{expl:vit:mda:walker25}};
  \node[anchor=south, font=\tiny, text width=\introimgw, align=center] (h4) at (4*\introimgw + 4*\introcolsep + 0.5*\introimgw, \introhdrsep) {LeGrad~\cite{expl:vit:legrad:bousselham25}};
  \node[anchor=south, font=\tiny, text width=\introimgw, align=center] (h5) at (5*\introimgw + 5*\introcolsep + 0.5*\introimgw, \introhdrsep) {TIS~\cite{expl:vit:tis:englebert23}};
  \node[anchor=south, font=\tiny, text width=\introimgw, align=center] (h6) at (6*\introimgw + 6*\introcolsep + 0.5*\introimgw, \introhdrsep) {ViT-CX~\cite{expl:vit:vitcx:xie23}};
  \node[anchor=south, font=\tiny, rotate=90, text width=\introimgw, align=center] at (-\introhdrsep, -0*\introimgw - 0*\introrowsep - 0.5*\introimgw) {\textcolor{red!80!black}{cairn} \\ (Norwich terrier)};
  \node[anchor=north west] at (0, -0*\introimgw - 0*\introrowsep) {\includegraphics[width=\introimgw]{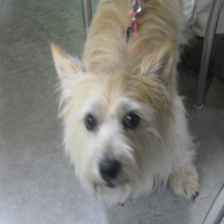}};
  \node[anchor=north west] at (1*\introimgw + 1*\introcolsep, -0*\introimgw - 0*\introrowsep) {\includegraphics[width=\introimgw]{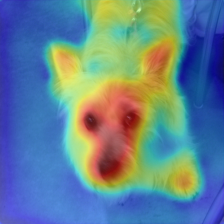}};
  \node[anchor=north west] at (2*\introimgw + 2*\introcolsep, -0*\introimgw - 0*\introrowsep) {\includegraphics[width=\introimgw]{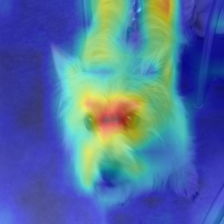}};
  \node[anchor=north west] at (3*\introimgw + 3*\introcolsep, -0*\introimgw - 0*\introrowsep) {\includegraphics[width=\introimgw]{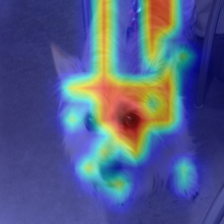}};
  \node[anchor=north west] at (4*\introimgw + 4*\introcolsep, -0*\introimgw - 0*\introrowsep) {\includegraphics[width=\introimgw]{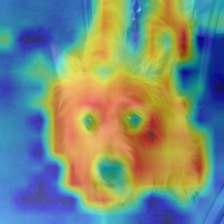}};
  \node[anchor=north west] at (5*\introimgw + 5*\introcolsep, -0*\introimgw - 0*\introrowsep) {\includegraphics[width=\introimgw]{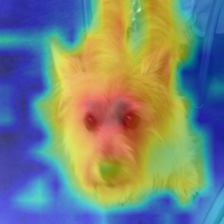}};
  \node[anchor=north west] at (6*\introimgw + 6*\introcolsep, -0*\introimgw - 0*\introrowsep) {\includegraphics[width=\introimgw]{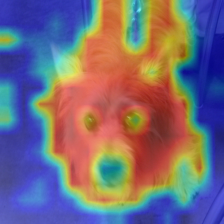}};
  \node[anchor=south, font=\tiny, rotate=90, text width=\introimgw, align=center] at (-\introhdrsep, -1*\introimgw - 1*\introrowsep - 0.5*\introimgw) {\textcolor{red!80!black}{can opener} \\ (bottlecap)};
  \node[anchor=north west] at (0, -1*\introimgw - 1*\introrowsep) {\includegraphics[width=\introimgw]{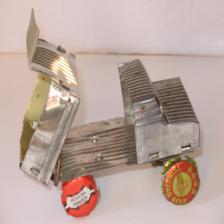}};
  \node[anchor=north west] at (1*\introimgw + 1*\introcolsep, -1*\introimgw - 1*\introrowsep) {\includegraphics[width=\introimgw]{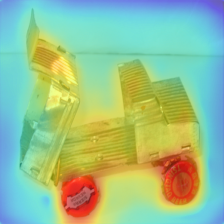}};
  \node[anchor=north west] at (2*\introimgw + 2*\introcolsep, -1*\introimgw - 1*\introrowsep) {\includegraphics[width=\introimgw]{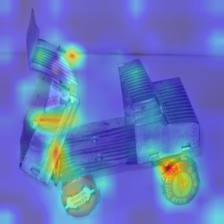}};
  \node[anchor=north west] at (3*\introimgw + 3*\introcolsep, -1*\introimgw - 1*\introrowsep) {\includegraphics[width=\introimgw]{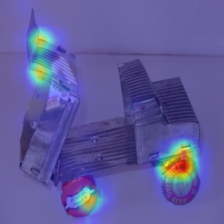}};
  \node[anchor=north west] at (4*\introimgw + 4*\introcolsep, -1*\introimgw - 1*\introrowsep) {\includegraphics[width=\introimgw]{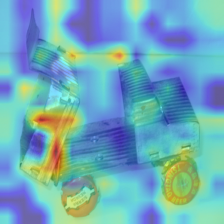}};
  \node[anchor=north west] at (5*\introimgw + 5*\introcolsep, -1*\introimgw - 1*\introrowsep) {\includegraphics[width=\introimgw]{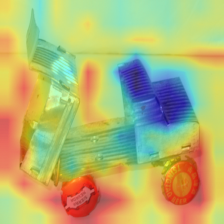}};
  \node[anchor=north west] at (6*\introimgw + 6*\introcolsep, -1*\introimgw - 1*\introrowsep) {\includegraphics[width=\introimgw]{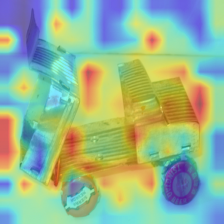}};
  \node[anchor=south, font=\tiny, rotate=90, text width=\introimgw, align=center] at (-\introhdrsep, -2*\introimgw - 2*\introrowsep - 0.5*\introimgw) {\textcolor{red!80!black}{impala} \\ (baboon)};
  \node[anchor=north west] at (0, -2*\introimgw - 2*\introrowsep) {\includegraphics[width=\introimgw]{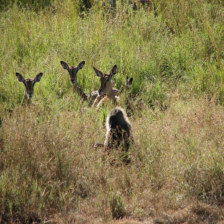}};
  \node[anchor=north west] at (1*\introimgw + 1*\introcolsep, -2*\introimgw - 2*\introrowsep) {\includegraphics[width=\introimgw]{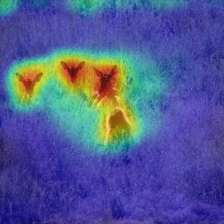}};
  \node[anchor=north west] at (2*\introimgw + 2*\introcolsep, -2*\introimgw - 2*\introrowsep) {\includegraphics[width=\introimgw]{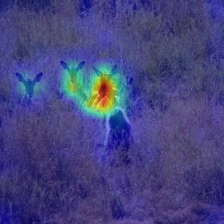}};
  \node[anchor=north west] at (3*\introimgw + 3*\introcolsep, -2*\introimgw - 2*\introrowsep) {\includegraphics[width=\introimgw]{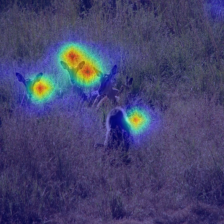}};
  \node[anchor=north west] at (4*\introimgw + 4*\introcolsep, -2*\introimgw - 2*\introrowsep) {\includegraphics[width=\introimgw]{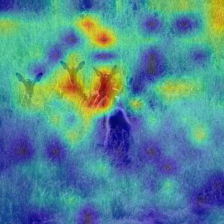}};
  \node[anchor=north west] at (5*\introimgw + 5*\introcolsep, -2*\introimgw - 2*\introrowsep) {\includegraphics[width=\introimgw]{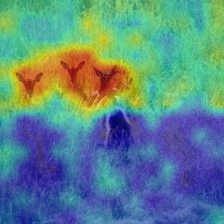}};
  \node[anchor=north west] at (6*\introimgw + 6*\introcolsep, -2*\introimgw - 2*\introrowsep) {\includegraphics[width=\introimgw]{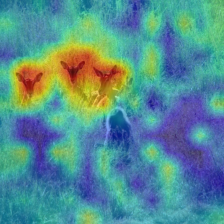}};
  \node[anchor=south, font=\tiny, rotate=90, text width=\introimgw, align=center] at (-\introhdrsep, -3*\introimgw - 3*\introrowsep - 0.5*\introimgw) {\textcolor{green!70!black}{reel}};
  \node[anchor=north west] at (0, -3*\introimgw - 3*\introrowsep) {\includegraphics[width=\introimgw]{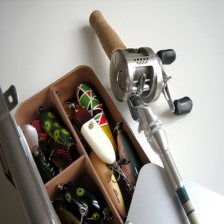}};
  \node[anchor=north west] at (1*\introimgw + 1*\introcolsep, -3*\introimgw - 3*\introrowsep) {\includegraphics[width=\introimgw]{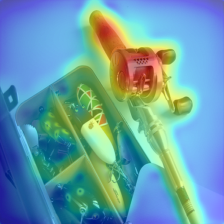}};
  \node[anchor=north west] at (2*\introimgw + 2*\introcolsep, -3*\introimgw - 3*\introrowsep) {\includegraphics[width=\introimgw]{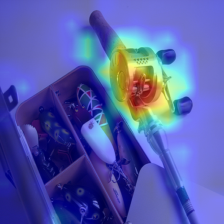}};
  \node[anchor=north west] at (3*\introimgw + 3*\introcolsep, -3*\introimgw - 3*\introrowsep) {\includegraphics[width=\introimgw]{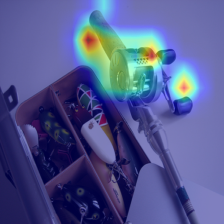}};
  \node[anchor=north west] at (4*\introimgw + 4*\introcolsep, -3*\introimgw - 3*\introrowsep) {\includegraphics[width=\introimgw]{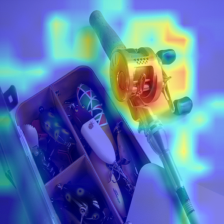}};
  \node[anchor=north west] at (5*\introimgw + 5*\introcolsep, -3*\introimgw - 3*\introrowsep) {\includegraphics[width=\introimgw]{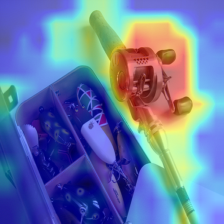}};
  \node[anchor=north west] at (6*\introimgw + 6*\introcolsep, -3*\introimgw - 3*\introrowsep) {\includegraphics[width=\introimgw]{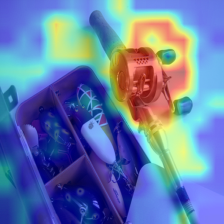}};
\end{tikzpicture}}

%% file: sec/2_related_work.tex
\section{Related Work}
\label{sec:related_work}

Visual attribution methods highlight the most influential input regions for a given model decision.
\textit{Ante-hoc} methods~\cite{expl:antehoc:aim:alshami25} alter the model architecture or training, \eg., prototype-based networks~\cite{expl:antehoc:cbr:li18, expl:antehoc:prototrees:nauta21, expl:antehoc:protopnet:chen19, expl:antehoc:deformableprotopnet:donnelly22}, concept bottleneck models~\cite{expl:antehoc:cbm:koh20, expl:antehoc:lfcbm:oikarinen23}, or alignment-based approaches~\cite{expl:antehoc:bcos:boehle22, expl:antehoc:bcosification:arya24, expl:antehoc:bcostransformers:tran23}.
The vast majority, however, are \textit{post-hoc} methods that explain already trained models without modifying them, typically via \textit{propagation-}, \textit{perturbation-} or \textit{learning-based} methods.

\myparagraph{Propagation-based Methods} propagate attribution signals through the model to determine the relevance of input features.
\textit{Backpropagation}-based methods use gradients of the output~\wrt~the input~\cite{expl:grad:vanilla:simonyan13, expl:grad:deconvnet:zeiler14}.
Various enhancements mitigate noisy gradients, including aggregating multiple gradient signals~\cite{expl:grad:xrai:kapishnikov19, expl:grad:scalespace:xu20, expl:grad:integratedgradients:sundararajan17, expl:grad:gig:kapishnikov21, expl:grad:fullgradient:srinivas19}, preventing negative gradients from flowing back~\cite{expl:grad:striving:springenberg15}, or specific backpropagation rules~\cite{expl:grad:lrp:bach15, expl:grad:deeplift:shrikumar17, expl:grad:excitationbackprop:zhang18}. %
\textit{Forward propagation}-based methods, on the other hand, utilize the activations of intermediate layers to generate attribution maps~\cite{expl:act:cam:zhou16, expl:act:liftcam:jung21}.
Grad-CAM~\cite{expl:act:gradcam:selvaraju17} and its successors~\cite{expl:act:gradcampp:chattopadhay18, expl:act:layercam:jiang21} use the gradient~\wrt~the feature maps to weight the activations in the last convolutional layers.
Specialized approaches to interpret vision transformers~\cite{classif:vit:dosovitskiy21, classif:swin:liu21, selfsup:dino:caron21, selfsup:mae:he22, selfsup:clip:radford21} typically analyze the self-attention mechanism:
Attention-Rollout~\cite{expl:vit:rollout:abnar20} recursively multiplies attention weights across layers, Attention-Flow~\cite{expl:vit:rollout:abnar20} models attention as a flow network, and Transition-Attention~\cite{expl:vit:transatt:yuan21} accumulates attention as a Markov process combined with Integrated Gradients~\cite{expl:grad:integratedgradients:sundararajan17}.
Based on LRP~\cite{expl:grad:lrp:bach15}, Transformer-Attribution~\cite{expl:vit:transformerinterpretability:chefer21} proposes propagation rules with gradient integration for self-attention layers.
To approximate token contributions, Bidirectional Attention~\cite{expl:vit:bidirectionalatt:chen23} jointly analyzes attention and partial derivatives to produce interpretable attribution signals.
Instead of using the gradients to weight the attention maps, LeGrad~\cite{expl:vit:legrad:bousselham25} uses the gradient~\wrt~the attention scores as attribution signals.

The major advantages of these approaches are computational efficiency and their model faithfulness as the attribution is directly derived from the model's internal computations.
However, they are often tied to specific architectures or layer types and the reliance on local gradients can bias the attributions toward low-level edges and textures~\cite{sanity:modifiedbp:sixt20, sanity:saliencymaps:adebayo18}.

\myparagraph{Perturbation-based Methods} directly measure output changes under systematic input perturbations to create causally grounded attributions~\cite{expl:pert:pd:zintgraf17}.
In their pioneering work, Zeiler and Fergus~\cite{expl:grad:deconvnet:zeiler14} monitored the model's output change while a sliding window occludes parts of the input.
LIME~\cite{expl:pert:lime:ribeiro16} uses superpixel perturbations to fit an interpretable surrogate model per sample, whereas SHAP~\cite{expl:pert:shap:lundberg17} leverages cooperative game theory to assign attributions by approximating Shapley values over feature subsets for individual predictions.
Fong~\etal~\cite{expl:pert:meaningfulpert:fong17, expl:pert:extremalpert:fong19} formulate attribution as a per-sample optimization problem, searching for the smallest region that maximally influences the output.
RISE~\cite{expl:pert:rise:petsiuk18} generates attribution maps as an output score-weighted combination of random input masks, while EVA~\cite{expl:pert:eva:fel23} proposes a more systematic and exhaustive exploration of the perturbation space.
While these perturbation-based approaches have predominantly been studied for CNNs, recent adaptations exploit the patch-wise input representation of vision transformers:
ViT-CX~\cite{expl:vit:vitcx:xie23} derives attribution by scoring the causal impact of patch tokens on the model output, while TIS~\cite{expl:vit:tis:englebert23} leverages the variable-length token processing of transformers by removing token subsets.
Contrary to heuristic attribution generation methods, Metric-Driven Attributions~(MDA)~\cite{expl:vit:mda:walker25} aims to directly optimize the metrics used to evaluate the attribution maps. 
In a per-sample optimization process, MDA aims to find the optimal order and magnitude of patch perturbations to maximize the evaluation metrics.
Relatedly, \textit{Less is More}~\cite{expl:vit:lessismore:chen24} builds on a given attribution map as a prior and refines it via greedy submodular selection of few regions, resulting in a small set of explanatory masks.

Since perturbation-based methods directly measure the model responses, they provide causally grounded attributions and can capture higher-order feature interactions.
However, a major drawback is that they are computationally expensive as multiple model evaluations are required per attribution map.
Moreover, methods tailored to vision transformers typically yield only patch-level attributions and therefore rely on upsampling to obtain input-resolution maps, often resulting in coarse and spatially imprecise attribution maps (recall \cref{fig:intro}).

\myparagraph{Learning-based Methods (Amortized Explainers)} 
train a neural explainer model to directly output attributions~\cite{expl:amort:stochamort:covert24}.
This amortizes the cost of per-sample iterative computation by shifting it to an offline training stage performed once per target model, while explanations can be generated efficiently at inference time.
\textit{Prior explanation}-driven approaches use other attribution methods as supervisory signals~\cite{expl:amort:learn2explain:situ21} or approximate Shapley-based priors~\cite{expl:amort:fastshap:jethani22}.
\textit{Self-supervised amortizers} on the other hand, interact with the target model to optimize a proxy objective.
However, these objectives often rely on ideal assumptions like a near-binary signal/noise decomposition to isolate relevant features~\cite{expl:amort:vert:bhalla23}, or assume that only a small, separable feature subset carries the predictive signal~\cite{expl:amort:l2x:chen18}, which may not hold in practice, especially for complex vision models.

\medskip\noindent
In contrast, our learning-based approach is directly guided by recognized evaluation metrics (\textit{Deletion} \& \textit{Insertion}) without any heuristic assumptions about the underlying data distribution or model behavior.
Since the supervisory signal is derived from the model's responses to perturbations, our method remains causally grounded.
Additionally, our end-to-end optimization schema is not limited to patch-level attribution, resulting in denser and better aligned results than existing ViT-specific methods.

%% file: sec/3_method.tex
\section{Amortized Hybrid Attribution}
\label{sec:method}

In the following, we first formalize the problem setting to obtain attribution maps produced for a fixed, pretrained classifier.
We then review the non-differentiable perturbation-based \textit{Deletion} and \textit{Insertion} metrics that serve as our target notion of attribution quality, and present our differentiable relaxation.
This makes the involved sorting and top-$k$ selection operations differentiable, allowing us to optimize the metrics end-to-end.
Finally, we describe practical strategies to make training efficient and robust at high resolution, as well as an optional test-time refinement to adapt the attribution map to sample-specific characteristics for use cases which prioritize explanation quality over throughput.

\myparagraph{Problem Setting and Notation. }
Let $f: \mathbb{R}^{H \times W \times 3} \rightarrow \mathbb{R}^{C}$ denote a pretrained image classifier that maps an input image $\mathbf{I} \in \mathbb{R}^{H \times W \times 3}$ to class probabilities \mbox{$f(\mathbf{I}) = \mathbf{s} \in \mathbb{R}^{C}$}, where $s_c$ corresponds to the predicted probability of class~$c$.
Given a target class $t \in \{1, \dots, C\}$, an attribution method is defined as a mapping $\Phi_f : \mathbb{R}^{H \times W \times 3} \times \{1, \dots, C\} \rightarrow \mathbb{R}^{H \times W}$, which produces an attribution map $\mathbf{A} = \Phi_f(\mathbf{I}, t)$ of the same spatial resolution as the input, where each element $a_{ij}$ reflects the importance of spatial location $(i,j)$ to the target probability $s_t$.
Unless otherwise specified $t = \arg\max_c s_c$.

\myparagraph{Metrics. }
Attribution methods are commonly evaluated by perturbation metrics measuring changes in model output under attribution-guided input perturbations.
Petsiuk~\etal~\cite{expl:pert:rise:petsiuk18} introduced the \textit{Deletion} and \textit{Insertion} metrics, which are widely used for attribution assessment~\cite{expl:vit:legrad:bousselham25, expl:vit:transformerinterpretability:chefer21, expl:vit:gam:chefer21, expl:vps:chen25} and optimization~\cite{expl:vit:mda:walker25}. 
\textit{Deletion} quantifies the drop in the target probability as high-attribution pixels are removed, while \textit{Insertion} measures the probability increase as these pixels are added to a blank reference image.
Formally this is defined as follows:
Let $\mathbf{a} = \mathrm{vec}(\mathbf{A}) \in \mathbb{R}^N$ with $N = H \cdot W$ denote the vectorized attribution map. 
First, a permutation $\pi_{\mathbf{A}}$ over the pixel indices is computed such that it sorts the entries of $\mathbf{a}$ in descending order, $a_{\pi_{\mathbf{A}}(1)} \geq a_{\pi_{\mathbf{A}}(2)} \geq \dots \geq a_{\pi_{\mathbf{A}}(N)}$.
Based on this order a sequence of binary masks $\{\mathbf{M}_{\mathbf{A}}^{(k)}\}_{k=1}^{N}$ is constructed, where each mask $\mathbf{M}_{\mathbf{A}}^{(k)} \in \{0,1\}^{H \times W}$ selects the top-$k$ pixels.
Next, the input image $\mathbf{I}$ is progressively blended with a blank reference image $\mathbf{I}_0$ (e.g., black image, mean colored image or a blurred version of the original image) using these masks, yielding two sequences of perturbed images $\{\mathbf{I}_{\textit{del}}^{(k)}\}_{k=1}^{N}$ and $\{\mathbf{I}_{\textit{ins}}^{(k)}\}_{k=1}^{N}$ where,
\begin{align}
\mathbf{I}_{\textit{del}}^{(k)}(\mathbf{A}, \mathbf{I})
&= \mathbf{I} \odot (1 - \mathbf{M}_{\mathbf{A}}^{(k)})
  + \mathbf{I}_0 \odot \mathbf{M}_{\mathbf{A}}^{(k)},
    \label{eq:deletion_perturbation}
\\
\mathbf{I}_{\textit{ins}}^{(k)}(\mathbf{A}, \mathbf{I})
&= \mathbf{I} \odot \mathbf{M}_{\mathbf{A}}^{(k)}
  + \mathbf{I}_0 \odot (1 - \mathbf{M}_{\mathbf{A}}^{(k)}),
    \label{eq:insertion_perturbation}
\end{align}
with $\odot$ denoting element-wise multiplication.
In $\mathbf{I}_{\textit{del}}^{(k)}$, the top-$k$ pixels of the original image are replaced by the blank reference image, while in $\mathbf{I}_{\textit{ins}}^{(k)}$, the top-$k$ pixels are inserted into the blank reference image.
The attribution quality is then evaluated by computing the area under the curve (AUC) of the target class probability $s_t$ across the sequence of perturbed images:
\begin{align}
\textit{del}_\text{AUC}(\mathbf{A}, \mathbf{I}, t)
&= \frac{1}{N} \sum_{k=1}^{N}
    f(\mathbf{I}_{\textit{del}}^{(k)}(\mathbf{A}, \mathbf{I}))_t,
    \label{eq:deletion_auc}
\\
\textit{ins}_\text{AUC}(\mathbf{A}, \mathbf{I}, t)
&= \frac{1}{N} \sum_{k=1}^{N}
    f(\mathbf{I}_{\textit{ins}}^{(k)}(\mathbf{A}, \mathbf{I}))_t.
    \label{eq:insertion_auc}
\end{align}
Meaningful attribution maps lead to a rapid score drop under \textit{Deletion} (low $\textit{del}_\text{AUC}$) and a rapid score increase under \textit{Insertion} (high $\textit{ins}_\text{AUC}$), indicating that highly ranked pixels indeed have strong influence on the prediction.\\

\subsection{Learn to Attribute}
\label{sec:learn-to-attribute}
In our method we aim to train an attribution model~$\Phi_{f,\theta}$ with parameters~$\theta$ for a given model~$f$, that is able to create dense attribution maps~$\mathbf{A}$ which follow the definition of importance as measured by the \textit{Deletion} and \textit{Insertion} metrics. 
However, the metrics defined in \Cref{eq:deletion_auc,eq:insertion_auc} are non-differentiable~\wrt~the attribution map~$\mathbf{A}$ due to the discrete pixel sorting (permutation)~$\pi_{\mathbf{A}}$ and the corresponding binary top-$k$ masks~$\mathbf{M}_{\mathbf{A}}^{(k)}$, which introduce discontinuities whenever the order changes.
Nevertheless, a key observation is that both metrics depend exclusively on the \textit{relative order} of the attribution values, not their absolute values. 
This fact allows us to bypass the non-differentiability by employing continuous relaxations of the sorting and selection operations, using differentiable sorting techniques~\cite{sort:softsort:prillo20, sort:fastdiffsort:blondel20, sort:gumbelsinkhorn:mena18} to obtain smooth approximations. 
This relaxation yields soft permutation matrices and, in turn, soft top-$k$ masks that enable end-to-end optimization of $\textit{del}_\text{AUC}$ and $\textit{ins}_\text{AUC}$ with respect to the attributions.

\myparagraph{Soft Permutation Matrices. }
The hard permutation $\pi_{\mathbf{A}}$ induced by sorting the attribution scores $\mathbf{a}$ can be represented as permutation matrix $\mathbf{P}_{\pi_{\mathbf{A}}} \in \{0, 1\}^{N \times N}$ such that,
\begin{equation}
    \mathbf{P}_{\pi_{\mathbf{A}}} \mathbf{a} = \begin{bmatrix} a_{\pi_{\mathbf{A}}(1)}, \: a_{\pi_{\mathbf{A}}(2)}, \: \dots, \:  a_{\pi_{\mathbf{A}}(N)}\end{bmatrix}^\top.
\end{equation}
Our approach to obtain a differentiable approximation of this permutation matrix is based on the \textit{Gumbel-Sinkhorn} algorithm~\cite{sort:gumbelsinkhorn:mena18}.
First, we compute a similarity \mbox{matrix~$\mathbf{L} \in \mathbb{R}^{N \times N}$}, which measures the similarity between the attribution scores $\mathbf{a}$ produced by the explainer model and a set of $N$ target positions $\mathbf{p} = \frac{1}{N} [N, N-1, \dots, 1]^\top$ defined as linearly decreasing values, representing the ideal sorted distribution.
Each entry $\mathbf{L}_{i,j}$ is computed as the negative squared distance between the attribution score $a_i$ and the target position $p_j$,
\begin{equation}
    \mathbf{L}_{i,j} = -(a_i - p_j)^2,
    \label{eq:similarity_matrix_entries}
\end{equation}
where larger values correspond to smaller assignment costs and therefore better matches between attribution scores and target ranks.
Subsequently, we add noise sampled from the Gumbel distribution $\mathbf{G} \sim \text{Gumbel}(0, 1)$ to introduce stochasticity, encouraging the exploration of different permutations during training,
\begin{equation}
    \widetilde{\mathbf{L}} = \frac{\mathbf{L} + \mathbf{G}}{\tau},
\end{equation}
where $\tau > 0$ is a temperature parameter that controls the smoothness of the resulting approximation.
This formulation frames the sorting task as an \textit{optimal transport~(OT)} problem, seeking the minimum-cost assignment between the attribution scores and the target positions.
By applying the \textit{Sinkhorn-Knopp} algorithm~\cite{sort:sinkhorn:sinkhorn67} to the matrix~$\widetilde{\mathbf{L}}$, which performs iterative row and column normalization, we obtain a doubly-stochastic matrix \mbox{$\mathbf{P}^{\text{soft}}_{\pi_{\mathbf{A}}} \in [0, 1]^{N \times N}$} that serves as a continuous relaxation of the permutation matrix~$\mathbf{P}_{\pi_{\mathbf{A}}}$.

\myparagraph{Soft Top-$k$ Masks. }
A hard top-$k$ mask $\mathbf{M}_{\mathbf{A}}^{(k)} \in \{0, 1\}^{H \times W}$ selects all pixels whose rank is between $1$ and $k$.
The vectorized form of this mask $\mathbf{m}_{\mathbf{A}}^{(k)} \in \{0, 1\}^{N}$ is created by summing the first $k$ rows of the permutation matrix~$\mathbf{P}_{\pi_{\mathbf{A}}}$.
Similarly, given our soft permutation matrix~$\mathbf{P}^{\text{soft}}_{\pi_{\mathbf{A}}}$, we create a sequence of $N$ soft top-$k$ masks 
$\{\mathbf{m}_{\mathbf{A}}^{\text{soft},(k)}\}_{k=1}^{N}$, where $\mathbf{m}_{\mathbf{A}}^{\text{soft},(k)} \in [0, 1]^{N}$ is obtained by computing the cumulative sum along the rank dimensions (rows) of the soft permutation matrix 
\begin{equation}
    \mathbf{m}_{\mathbf{A}}^{\text{soft},(k)} = \sum_{i=1}^{k} \mathbf{P}^{\text{soft}}_{\pi_{\mathbf{A}}}[i, : ]^\top.
\end{equation}
This results in masks with continuous values between $0$ and $1$, indicating the degree to which each pixel is included in the top-$k$ selection, illustrated in \cref{fig:soft_masking}.

\input{figures/soft_masking/fig.tex}

\myparagraph{Differentiable \textit{Deletion} \& \textit{Insertion}. }
Given the sequence of soft top-$k$ masks, we now replace the hard masks~$\mathbf{M}_{\mathbf{A}}^{(k)}$ in \Cref{eq:deletion_perturbation,eq:insertion_perturbation} with the reshaped soft masks~$\mathbf{m}_{\mathbf{A}}^{\text{soft},(k)}$ to obtain $\mathbf{I}_{\textit{del}}^{\text{soft},(k)}$ and $\mathbf{I}_{\textit{ins}}^{\text{soft},(k)}$.
Consequently, we get differentiable approximations of the \textit{Deletion} and \textit{Insertion} metrics as:
\begin{align}
    \textit{del}_\text{AUC}^{\text{soft}}(\mathbf{A},\, \mathbf{I},\, t)
    &= \frac{1}{N} \sum_{k = 1}^{N}
    f\!\left(\mathbf{I}_{\textit{del}}^{\text{soft},(k)} (\mathbf{A},\mathbf{I})\right)_{\!t},
    \label{eq:deletion_auc_soft}
    \\[2pt]
    \textit{ins}_\text{AUC}^{\text{soft}}(\mathbf{A},\, \mathbf{I},\, t)
    &= \frac{1}{N} \sum_{k = 1}^{N}
    f\!\left(\mathbf{I}_{\textit{ins}}^{\text{soft},(k)} (\mathbf{A},\mathbf{I})\right)_{\!t}.
    \label{eq:insertion_auc_soft}
\end{align}
The loss function for training the attribution model can then be defined as a weighted combination of these two metrics, encouraging the model to produce attribution maps guided directly by attribution quality measures:
\begin{equation}
    \mathcal{L}(\theta) 
    = \lambda_\text{del} \cdot \textit{del}_\text{AUC}^{\text{soft}}(\mathbf{A},\, \mathbf{I},\, t)
    - \lambda_\text{ins} \cdot \textit{ins}_\text{AUC}^{\text{soft}}(\mathbf{A},\, \mathbf{I},\, t),
    \label{eq:final_loss}
\end{equation}
where $\lambda_\text{del}, \lambda_\text{ins}$ are weighting hyperparameters.

\subsection{Efficient and Robust Optimization}
\label{sec:efficient-robust-optimization}
Although the differentiable relaxations allow us to compute gradients with respect to the attribution maps, a naive implementation of the training process that involves computing the full sequence of $N$ perturbations for each image would be computationally prohibitive, especially for high-resolution images.
On the other hand, optimizing the loss described in \Cref{eq:final_loss} on pixel-level would be prone to overfitting, as the attribution model could exploit high-frequency artifacts, resulting in directions targeting the architecture-dependent weaknesses of the model under examination.
To mitigate these issues, we propose strategies making the training process practical and robust:
\begin{itemize}
    \item \textbf{Region-Based Permutation. }
        Instead of sorting all $N$ pixels individually, we partition the image into $K \ll N$ disjoint regions using a regular grid of size $K = G \cdot G$ and mean-pool the attribution values within each region.
        After the differentiable sorting and soft top-$k$ selection on these region-level scores, the resulting masks are upsampled to the original image resolution for perturbation.
        This significantly reduces the number of perturbations during training and reduces the risk of exploiting individual pixel artifacts.
    \item \noindent\textbf{Sampling Perturbation Steps. }
        To further reduce the computational burden during training, we do not evaluate the full sequence of $K$ perturbation steps for each image.
        Instead, we uniformly sample a smaller set of $S \ll K$ steps from the range $\{1, \dots, K\}$,~\ie, we perturb multiple grid regions at once instead of adding them one by one.
        This approximation of the AUC metrics decreases the number of needed forward passes through the model to a fixed budget.
    \item \textbf{Grid Augmentation. }
        While the grid-based region partitioning enhances the training efficiency and robustness, a fixed grid alignment may introduce bias towards specific spatial patterns.
        To alleviate this effect we randomly vary the grid size~$G$ within a predefined range for each training iteration and apply random offsets to the grid's position.
        These augmentations encourage the attribution model to focus on generalizable importance patterns that are not tied to specific grid configurations or scales.
    \item \noindent\textbf{Regularization. }
        In addition to the objective in \Cref{eq:final_loss}, we regularize the attribution model to discourage spurious, isolated peaks, \ie~individual pixels or small clusters of pixels with high scores that are not supported by their neighboring regions.
        Concretely, we convolve the attribution map~$\mathbf{A}$ with an averaging filter and penalize large deviations from the original map using an $L_2$ loss.
\end{itemize}

Taken together, these strategies do not merely make training tractable, they also shape which attribution patterns the objective rewards.
A natural concern is that the order-based metric could be satisfied by diffuse maps, since only the relative ranking matters in \Cref{eq:deletion_auc,eq:insertion_auc}.
The optimization, however, does not act on the ordering alone.
The Sinkhorn assignment matches the continuous attribution values to the target ranks through the distance costs in \Cref{eq:similarity_matrix_entries}, so the degree of separation between region scores controls the sharpness of the resulting soft top-$k$ masks.
Near-tied scores thus yield ambiguous masks that mix foreground and background, whereas clearly separated scores yield decisive masks.
This separation is what the AUC objectives reward: \textit{Deletion} and \textit{Insertion} are driven by a steep score change already at small $k$, which only arises when the perturbation is concentrated on the few truly causal regions.
Decisive masks therefore induce low \textit{Deletion} and high \textit{Insertion}, while diffuse masks perturb the whole image weakly and flatten both curves.
Together with the region-level perturbations, the randomized grid sizes and offsets, and the regularizer penalizing isolated peaks, this drives the explainer toward compact, object-aligned attributions.

\subsection{Optional Test-Time Refinement}

The design of our attribution framework allows for an optional, fast test-time refinement step at low computational cost, that can further enhance the quality of the generated heatmaps on a per-image basis.
The key idea is to briefly fine-tune the explainer for a few gradient steps on the specific test image using the same differentiable AUC objectives used during training.
To make refinement more stable and deterministic, we use soft sorting without Gumbel noise and fix the sorting temperature to the final training value, \ie $\tau=\tau_{\mathrm{final}}$.
After $T$ optimization steps, a final forward pass produces the refined attribution map.
This refinement is particularly useful when explanation quality is prioritized over throughput, with the compute-quality trade-off being adjustable via the number of refinement steps. 
Moreover, if inputs are expected to significantly deviate from the training distribution, this test-time refinement can help to adapt the explainer to the specific characteristics of the new input, improving attribution quality in such cases.

%% file: figures/soft_masking/fig.tex
\begin{figure}[tb]
    \centering
    \resizebox{0.85\linewidth}{!}{
        \input{figures/soft_masking/source.tex}
    }
        \caption{\textbf{Soft deletion masks.} Given an input image $\mathbf{I}$ and its attribution map $\mathbf{A}$, the soft top-$k$ masks $\mathbf{m}_{\mathbf{A}}^{\text{soft},(k)}$ progressively replace the most important regions with the reference image. Unlike hard binary masks, the soft relaxation (\cref{sec:learn-to-attribute}) assigns continuous importance values to each region, enabling gradient-based optimization of the deletion and insertion metrics.
        Note the varying grid resolution $G$ across examples top \emph{vs.} bottom (\cref{sec:efficient-robust-optimization}) which prevents overfitting to a fixed spatial partitioning.
        }
    \label{fig:soft_masking}
\end{figure}
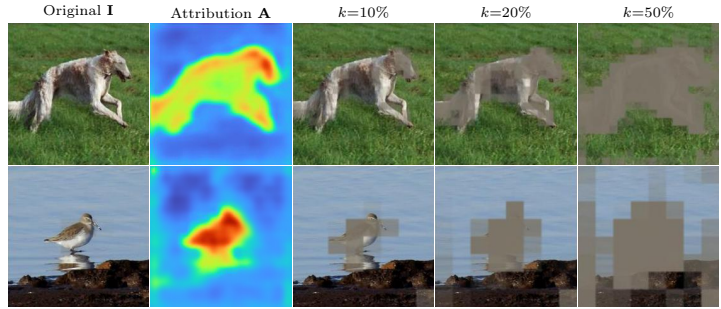

%% file: figures/soft_masking/source.tex
\newlength{\smcolsep}
\setlength{\smcolsep}{1pt}
\newlength{\smrowsep}
\setlength{\smrowsep}{1pt}
\newlength{\smhdrsep}
\setlength{\smhdrsep}{3pt}
\newlength{\smimgw}
\setlength{\smimgw}{\dimexpr(\linewidth - 4\smcolsep)/5\relax}

\begin{tikzpicture}[
    every node/.style={inner sep=0pt, outer sep=0pt},
]
  \node[anchor=south, font=\scriptsize, text width=\smimgw, align=center] at (0*\smimgw + 0*\smcolsep + 0.5*\smimgw, \smhdrsep) {Original $\mathbf{I}$};
  \node[anchor=south, font=\scriptsize, text width=\smimgw, align=center] at (1*\smimgw + 1*\smcolsep + 0.5*\smimgw, \smhdrsep) {Attribution $\mathbf{A}$};
  \node[anchor=south, font=\scriptsize, text width=\smimgw, align=center] at (2*\smimgw + 2*\smcolsep + 0.5*\smimgw, \smhdrsep) {$k{=}10\%$};
  \node[anchor=south, font=\scriptsize, text width=\smimgw, align=center] at (3*\smimgw + 3*\smcolsep + 0.5*\smimgw, \smhdrsep) {$k{=}20\%$};
  \node[anchor=south, font=\scriptsize, text width=\smimgw, align=center] at (4*\smimgw + 4*\smcolsep + 0.5*\smimgw, \smhdrsep) {$k{=}50\%$};
  \node[anchor=north west] at (0, 0) {\includegraphics[width=\smimgw]{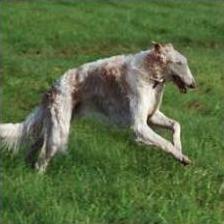}};
  \node[anchor=north west] at (1*\smimgw + 1*\smcolsep, 0) {\includegraphics[width=\smimgw]{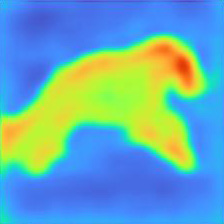}};
  \node[anchor=north west] at (2*\smimgw + 2*\smcolsep, 0) {\includegraphics[width=\smimgw]{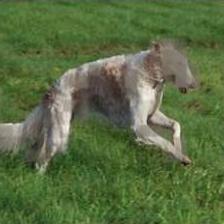}};
  \node[anchor=north west] at (3*\smimgw + 3*\smcolsep, 0) {\includegraphics[width=\smimgw]{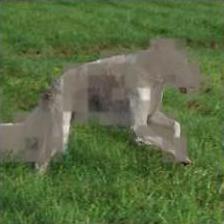}};
  \node[anchor=north west] at (4*\smimgw + 4*\smcolsep, 0) {\includegraphics[width=\smimgw]{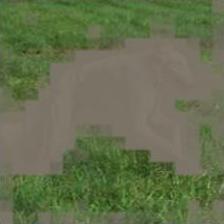}};
  \node[anchor=north west] at (0, -1*\smimgw - 1*\smrowsep) {\includegraphics[width=\smimgw]{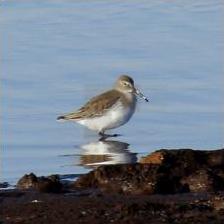}};
  \node[anchor=north west] at (1*\smimgw + 1*\smcolsep, -1*\smimgw - 1*\smrowsep) {\includegraphics[width=\smimgw]{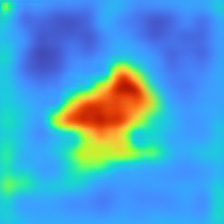}};
  \node[anchor=north west] at (2*\smimgw + 2*\smcolsep, -1*\smimgw - 1*\smrowsep) {\includegraphics[width=\smimgw]{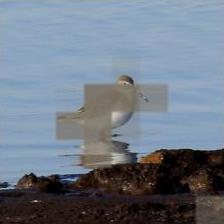}};
  \node[anchor=north west] at (3*\smimgw + 3*\smcolsep, -1*\smimgw - 1*\smrowsep) {\includegraphics[width=\smimgw]{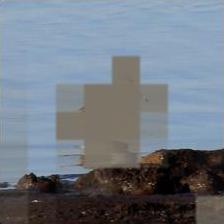}};
  \node[anchor=north west] at (4*\smimgw + 4*\smcolsep, -1*\smimgw - 1*\smrowsep) {\includegraphics[width=\smimgw]{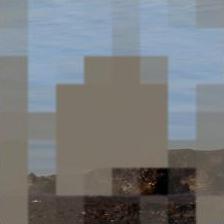}};
\end{tikzpicture}

%% file: sec/4_experiments.tex
\section{Experiments}

\subsection{Implementation Details}
\label{sec:eval_setup}

Unless stated otherwise, we use the following setup for all experiments throughout this section.
We train our attribution model on the ImageNet-1K training split~\cite{dataset:imagenet:deng09} to explain a frozen ImageNet \mbox{ViT-B/16} classifier~\cite{classif:vit:dosovitskiy21}.
Our explainer is an encoder-decoder model with a frozen DINOv3~\cite{model:dinov3:simeoni25} ViT-L/16 backbone and a DPT-inspired decoder~\cite{model:dpt:ranftl21} conditioned on a learned class embedding.
We optimize only the explainer head parameters and keep the DINOv3 backbone frozen.
To create perturbation masks during training, we average-pool the heatmap to a $G{\times}G$ grid and rank regions by mean attribution using Gumbel-Sinkhorn soft sorting with 30 Sinkhorn iterations. 
During training, $G$ is sampled uniformly from $\{7,\dots,28\}$ to reduce grid-alignment bias (region sizes $32$--$8$ px for $224{\times}224$ inputs).
The \textit{Deletion} and \textit{Insertion} curves are approximated using $S=16$ uniformly spaced perturbation steps during training. 
We equally weight the corresponding losses $\lambda_{\text{del}}=\lambda_{\text{ins}}=1$ and use $\lambda_{\text{reg}}=2.5{\times}10^{-3}$ for the pixel regularization term to penalize isolated high-amplitude activations.
The explainer is trained for one epoch with AdamW (lr $3{\times}10^{-4}$, weight decay $10^{-3}$), gradient clipping (norm 1.0), and a one-cycle cosine schedule.
In the optional refinement variant, we update only the explainer head for $T$ steps (lr $10^{-4}$) on the test image using the same loss, fixed $\tau{=}\tau_{\text{final}}$, and deterministic soft sorting (Gumbel noise disabled).

For evaluation we use the full corresponding validation split (50K images, 1{,}000 classes) without Test-Time Refinement ($T=0$), unless explicitly denoted.
We explain the \emph{top-1 predicted class} regardless of the output probability, as well as the \emph{ground truth class} if different, and report results separately for both targets.
All metrics are averaged over three different types of reference images $\mathbf{I}_0$ used for the perturbation-based evaluations to replace masked-out pixels: a black image, a uniform gray image (ImageNet mean), and a blurred version of the input image.
The detailed evaluation results for all metrics, individual reference images, and architectures are reported in the supplementary material.

\subsection{Faithfulness Evaluation}

Besides the \textit{Deletion} and \textit{Insertion} described in \Cref{sec:method}, we evaluate the faithfulness of the generated attribution maps also using the \textit{Positive} \& \textit{Negative} perturbation evaluation~\cite{expl:vit:transformerinterpretability:chefer21}: across the evaluation set, the most-attributed regions (\textit{Positive}) or the least-attributed regions (\textit{Negative}) are progressively removed and the target model's top-1 accuracy is tracked as a function of the percentage of perturbed pixels.
Low \textit{Positive} AUC values and high \textit{Negative} AUC values indicate that highly ranked regions capture important features for the target model, while low-ranked regions are non-informative.
We also report the \textit{Average Drop Percentage} (ADP) and the \textit{Percentage of Increase in Confidence} (PIC) \cite{expl:act:gradcampp:chattopadhay18}: 
for each sample an element-wise product between the normalized attribution map ($\in [0,1]$) and the input image is created to mask the background regions based on the attribution map.
The ADP is then computed as the average percentage drop in the target model's output score between the original image and the masked image.
The lower the ADP, the better the attribution map differentiates between the unimportant background and the important foreground.
The PIC, on the other hand, is defined as the percentage of cases in which this background-masking leads to an increase in the target model's confidence.
Finally, we also report the differences between \textit{Insertion} and \textit{Deletion} AUCs, as well as between \textit{Positive} and \textit{Negative} AUCs, as a measure of the \textit{Faithfulness}~\cite{metric:pixelflip:samek16, metric:faith1:muzellec24}.
To demonstrate that the findings are not mere artefacts of the optimization objective, we provide additional evaluations on MAS~\cite{metric:mas:walker24}, ROAD~\cite{metric:road:rong22} metrics and the ImageNet segmentation benchmark~\cite{dataset:imagenetseg:guillaumin14} in the supplemental material, which also confirm the performance benefits of AHA.

\Cref{tab:main_results_vitb16} (ViT-B/16) and \Cref{tab:main_results_vitb32} (ViT-B/32) summarize the results of our method compared to a wide range of baselines.
Across both analyzed classifiers and both target definitions, \methodabbrev without additional refinement steps ($T{=}0$) achieves competitive or superior performance to the current state-of-the-art, with particularly strong results on the \textit{Deletion} and \textit{Positive} metrics.
We argue that the reason why our \textit{zero-shot} method can compete with, or even outperform, per-sample-optimized methods is because it aligns attribution maps more precisely with the input image.
This enables important areas to be localized more accurately than with methods that only operate at patch level.
Using only three refinement steps ($T=3$) at test time, \methodabbrev further improves on all metrics, outperforming the baselines in the majority of cases and achieving the best overall \textit{Faithfulness} scores (\textit{Ins.} - \textit{Del.} and \textit{Pos.} - \textit{Neg.}).

\input{tables/main_results_vitb16/table.tex}
\input{tables/main_results_vitb32/table.tex}

\subsection{Qualitative Evaluation}

\Cref{fig:intro} shows qualitative examples of attribution maps generated by \methodabbrev compared to best-performing baselines across the evaluated metrics for the ViT-B/16 classifier.
Even without test-time refinement ($T=0$), \methodabbrev produces attribution maps that are more precisely aligned with the input image and better capture object boundaries than the baselines.
This is particularly evident for the image classified as \textit{impala}, where our attribution map correctly highlights the animal heads, while the baselines produce more diffuse, blocky maps that also cover the background because of the patch-level attribution.
Moreover, in the case of the misclassified \textit{can opener} our method clearly highlights the bottle caps as the most important features for the model's decision, while the baselines produce more noisy maps that are less interpretable.
Additional examples for all methods and classifiers are included in the supplementary material.

\subsection{Runtime Analysis}

The runtime comparison in \Cref{tab:main_results_vitb16,tab:main_results_vitb32}, measured over 100 iterations on a single NVIDIA RTX 4090, confirms the efficiency advantage of the amortized design.
At $T{=}0$, \methodabbrev requires only a single explainer forward pass and is therefore comparable to the runtime of propagation-based methods like Grad-CAM or LeGrad, while achieving competitive or superior faithfulness compared to methods that are orders of magnitude slower.
When refining over $T{=}3$ gradient steps, the runtime increases depending on the complexity of the target model, but generally remains below the token/patch perturbation-based methods and significantly below the per-sample metric optimization of MDA.
A detailed amortization study is provided in the supplemental material.

\input{figures/TS_eval/fig.tex}

\subsection{Test-Time Refinement Influence}

\input{figures/T_eval_examples/fig.tex}

We analyze the effect of applying $T$ gradient steps of test-time refinement on the attribution quality.
As shown in \Cref{fig:ts_eval_insertion_l}, performance improves monotonically with increasing $T$, with the largest gains concentrated in the first few steps and diminishing returns beyond $T{=}3$.
Specifically, three refinement steps increase the Insertion AUC from $0.611$ to $0.649$ (predicted class).
The Deletion AUC (where lower is better) follows a similar monotonically improving trend and is reduced from $0.138$ to $0.121$ for $T=0\rightarrow3$.
The qualitative examples in \Cref{fig:t_eval_examples} confirm this: refinement progressively sharpens the attribution maps, setting the focus onto the relevant regions and suppressing background activations most clearly visible for the \textit{goldfish} and \textit{scuba diver} examples.
The number of refinement steps $T$ thus provides a flexible compute-quality tradeoff, making \methodabbrev suitable both for high-throughput settings ($T{=}0$) and for scenarios where explanation fidelity is the primary interest.

\subsection{Perturbation Steps Influence}

During training, the \textit{Deletion} and \textit{Insertion} AUC objectives are approximated by sampling $S$ uniformly spaced perturbation steps instead of evaluating the full sequence.
We ablate the effect of this hyperparameter by training models with $S \in \{4, 8, 16, 32\}$.
As shown in \Cref{fig:ts_eval_insertion_s}, the performance improves consistently from $S{=}4$ to $S{=}16$, after which gains become marginal.
This plateau indicates that $S{=}16$ already provides a sufficiently dense approximation of the AUC curve.

\subsection{Limitations}

Despite the state-of-the-art results of our approach, few limitations should be acknowledged.
First, each new target model requires a retraining of the explainer.
While inference is efficient once trained, this amortization step introduces additional computational cost and limits immediate applicability to arbitrary models.
Second, our training objective optimizes approximations of deletion and insertion metrics. 
While they are widely used to assess attribution faithfulness, they remain proxy measures of explanation quality and have known shortcomings, like sensitivity to the choice of reference image and perturbation strategy (detailed results provided in the supplemental).
And third, our method requires access to a representative training dataset for the target model, which may be unavailable for proprietary or closed-source models.
However, our optional refinement ($T>0$) can be employed to resolve such data issues.

%% file: tables/main_results_vitb16/table.tex
\begin{table*}[t]
    \caption{
        Quantitative comparison of attribution methods on ImageNet using a \mbox{ViT-B/16} classifier. 
        All metrics are averaged over three reference image modes~$\mathbf{I}_0$ (black, mean, blur). Top-3 results per metric are highlighted (dark to light).
    }
    \input{tables/main_results_vitb16/source.tex}
    \label{tab:main_results_vitb16}
\end{table*}

%% file: tables/main_results_vitb16/source.tex
\small
\setlength{\tabcolsep}{4pt}
\definecolor{rankfirst}{RGB}{133,171,214}
\definecolor{ranksecond}{RGB}{173,201,230}
\definecolor{rankthird}{RGB}{211,226,243}
\resizebox{\textwidth}{!}{%
\begin{tabular}{clccccccccr}
  \toprule
   & Method  & Deletion$\downarrow$ & Insertion$\uparrow$ & Ins. - Del.$\uparrow$ & Positive$\downarrow$ & Negative$\uparrow$ & Neg. - Pos.$\uparrow$ & ADP$\downarrow$ & PIC$\uparrow$ & Runtime\\
  \midrule
  \multirow{10}{*}{\rotatebox[origin=c]{90}{\textbf{Predicted Class}}} & Grad-CAM~\cite{expl:act:gradcam:selvaraju17}  & 0.2858 & 0.4982 & 0.2124 & 0.3346 & 0.5788 & 0.2442 & 84.3339 & 4.3577 & \cellcolor{ranksecond}\underline{0.014s}\\
   & IG~\cite{expl:grad:integratedgradients:sundararajan17}  & 0.1857 & 0.5544 & 0.3686 & 0.2248 & 0.6347 & 0.4099 & 22.9340 & 33.4353 & 0.251s\\
   & Trans-Att.~\cite{expl:vit:transatt:yuan21}  & 0.1654 & 0.5833 & 0.4178 & 0.2022 & 0.6654 & 0.4633 & 25.6994 & 30.8467 & 0.254s\\
   & Bi-Att.~\cite{expl:vit:bidirectionalatt:chen23}  & 0.1600 & 0.5958 & 0.4358 & 0.1961 & 0.6789 & 0.4828 & 28.4572 & 28.2614 & 0.254s\\
   & TIS~\cite{expl:vit:tis:englebert23}  & \cellcolor{rankthird}0.1420 & \cellcolor{ranksecond}\underline{0.6381} & \cellcolor{ranksecond}\underline{0.4961} & \cellcolor{rankthird}0.1731 & \cellcolor{ranksecond}\underline{0.7222} & \cellcolor{ranksecond}\underline{0.5492} & \cellcolor{rankfirst}\textbf{11.2479} & \cellcolor{rankfirst}\textbf{45.5658} & 1.167s\\
   & ViT-CX~\cite{expl:vit:vitcx:xie23}  & 0.1750 & 0.5714 & 0.3964 & 0.2149 & 0.6478 & 0.4328 & \cellcolor{rankthird}12.5472 & \cellcolor{ranksecond}\underline{43.4971} & 0.860s \\
   & MDA~\cite{expl:vit:mda:walker25}  & 0.1556 & \cellcolor{rankthird}0.6112 & 0.4556 & 0.1933 & \cellcolor{rankthird}0.6983 & 0.5050 & 48.6653 & 13.2691 & 13.645s \\
   & LeGrad~\cite{expl:vit:legrad:bousselham25} & 0.1666 & 0.5666 & 0.4000 & 0.2036 & 0.6461 & 0.4425 & 20.3850 & 34.3440 & \cellcolor{rankfirst}\textbf{0.006s}\\
   & \methodabbrev (Ours) & \cellcolor{ranksecond}\underline{0.1381} & 0.6112 & \cellcolor{rankthird}0.4731 & \cellcolor{ranksecond}\underline{0.1714} & 0.6947 & \cellcolor{rankthird}0.5233 & 15.8713 & 36.8246 & \cellcolor{rankthird} 0.016s\\
   & \methodabbrev $T=3$ (Ours) & \cellcolor{rankfirst}\textbf{0.1215} & \cellcolor{rankfirst}\textbf{0.6485} & \cellcolor{rankfirst}\textbf{0.5271} & \cellcolor{rankfirst}\textbf{0.1491} & \cellcolor{rankfirst}\textbf{0.7378} & \cellcolor{rankfirst}\textbf{0.5887} & \cellcolor{ranksecond}\underline{12.4658} & \cellcolor{rankthird}42.2698 & 0.547s\\
  \midrule
  \multirow{10}{*}{\rotatebox[origin=c]{90}{\textbf{Ground Truth Class}}} & Grad-CAM~\cite{expl:act:gradcam:selvaraju17}  & 0.2690 & 0.4638 & 0.1948 & 0.3068 & 0.5244 & 0.2176 & 80.4883 & 7.7366 & \cellcolor{ranksecond}\underline{0.014s}\\
   & IG~\cite{expl:grad:integratedgradients:sundararajan17}  & 0.1731 & 0.5171 & 0.3440 & 0.2028 & 0.5765 & 0.3737 & 20.0260 & 38.5759 & 0.251s\\
   & Trans-Att.~\cite{expl:vit:transatt:yuan21}  & 0.1541 & 0.5428 & 0.3887 & 0.1819 & 0.6020 & 0.4200 & 22.5843 & 36.0353 & 0.254s \\
   & Bi-Att.~\cite{expl:vit:bidirectionalatt:chen23}  & 0.1492 & 0.5534 & 0.4042 & 0.1766 & 0.6120 & 0.4354 & 25.1370 & 33.5553 & 0.254s\\
   & TIS~\cite{expl:vit:tis:englebert23} & \cellcolor{rankthird}0.1324 & \cellcolor{ranksecond}\underline{0.5966} & \cellcolor{ranksecond}\underline{0.4641} & \cellcolor{rankthird}0.1553 & \cellcolor{ranksecond}\underline{0.6595} & \cellcolor{ranksecond}\underline{0.5042} & \cellcolor{rankfirst}\textbf{9.2014} & \cellcolor{rankfirst}\textbf{50.9796} & 1.167s \\
   & ViT-CX~\cite{expl:vit:vitcx:xie23} & 0.1694 & 0.5231 & 0.3537 & 0.2024 & 0.5748 & 0.3724 & 15.2074 & \cellcolor{rankthird}42.2452 & 0.860s \\
   & MDA~\cite{expl:vit:mda:walker25}  & 0.1461 & 0.5628 & 0.4168 & 0.1756 & 0.6216 & 0.4460 & 44.8870 & 18.2510 & 13.645s\\
   & LeGrad~\cite{expl:vit:legrad:bousselham25}  & 0.1552 & 0.5280 & 0.3728 & 0.1833 & 0.5859 & 0.4027 & 17.6998 & 39.4492 & \cellcolor{rankfirst}\textbf{0.006s}\\
   & \methodabbrev (Ours) & \cellcolor{ranksecond}\underline{0.1280} & \cellcolor{rankthird}0.5657 & \cellcolor{rankthird}0.4377 & \cellcolor{ranksecond}\underline{0.1529} & \cellcolor{rankthird}0.6222 & \cellcolor{rankthird}0.4693 & \cellcolor{rankthird}14.0053 & 41.2158 & \cellcolor{rankthird}0.016s\\
   & \methodabbrev $T=3$ (Ours) &  \cellcolor{rankfirst}\textbf{0.1126} & \cellcolor{rankfirst}\textbf{0.6003} & \cellcolor{rankfirst}\textbf{0.4877} & \cellcolor{rankfirst}\textbf{0.1326} & \cellcolor{rankfirst}\textbf{0.6630} & \cellcolor{rankfirst}\textbf{0.5303} & \cellcolor{ranksecond}\underline{10.8509} & \cellcolor{ranksecond}\underline{46.7211} & 0.547s\\
  \bottomrule
\end{tabular}
}

%% file: tables/main_results_vitb32/table.tex
\begin{table*}[t]
    \caption{
        Quantitative comparison of attribution methods on ImageNet using a \mbox{ViT-B/32} classifier. 
        All metrics are averaged over three reference image modes~$\mathbf{I}_0$ (black, mean, blur). Top-3 results per metric are highlighted (dark to light).
    }
    \input{tables/main_results_vitb32/source.tex}
    \label{tab:main_results_vitb32}
\end{table*}

%% file: tables/main_results_vitb32/source.tex
\small
\setlength{\tabcolsep}{4pt}
\definecolor{rankfirst}{RGB}{133,171,214}
\definecolor{ranksecond}{RGB}{173,201,230}
\definecolor{rankthird}{RGB}{211,226,243}
\resizebox{\textwidth}{!}{%
\begin{tabular}{clccccccccr}
  \toprule
   & Method & Deletion$\downarrow$ & Insertion$\uparrow$ & Ins. - Del.$\uparrow$ & Positive$\downarrow$ & Negative$\uparrow$ & Neg. - Pos.$\uparrow$ & ADP$\downarrow$ & PIC$\uparrow$ & Runtime\\
  \midrule
  \multirow{10}{*}{\rotatebox[origin=c]{90}{\textbf{Predicted Class}}} & Grad-CAM~\cite{expl:act:gradcam:selvaraju17}  & 0.2337 & 0.5294 & 0.2956 & 0.2930 & 0.6328 & 0.3398 & 66.9314 & 8.5072 & \cellcolor{ranksecond}\underline{0.010s}\\
   & IG~\cite{expl:grad:integratedgradients:sundararajan17}  & 0.1897 & 0.5542 & 0.3645 & 0.2426 & 0.6567 & 0.4141 & 14.2831 & 39.0279 & 0.157s\\
   & Trans-Att.~\cite{expl:vit:transatt:yuan21}  & 0.2024 & 0.5507 & 0.3483 & 0.2567 & 0.6534 & 0.3967 & 17.1553 & 34.9780 & 0.157s\\
   & Bi-Att.~\cite{expl:vit:bidirectionalatt:chen23}  & 0.1887 & 0.5726 & 0.3839 & 0.2416 & 0.6782 & 0.4366 & 17.6613 & 34.3086 & 0.158s\\
   & TIS~\cite{expl:vit:tis:englebert23}  & \cellcolor{rankthird}0.1702 & \cellcolor{rankfirst}\textbf{0.6094} & \cellcolor{ranksecond}\underline{0.4392} & \cellcolor{rankthird}0.2144 & \cellcolor{ranksecond}\underline{0.7181} & \cellcolor{ranksecond}\underline{0.5037} & \cellcolor{rankthird}9.7104 & \cellcolor{ranksecond}\underline{48.3937} & 0.416s\\
   & ViT-CX~\cite{expl:vit:vitcx:xie23}  & 0.1982 & 0.5604 & 0.3622 & 0.2525 & 0.6622 & 0.4098 & \cellcolor{ranksecond}\underline{9.3272} & \cellcolor{rankfirst}\textbf{48.9950} &  0.763s\\
   & MDA~\cite{expl:vit:mda:walker25}  & 0.1973 & \cellcolor{rankthird}0.5939 & 0.3966 & 0.2511 & \cellcolor{rankthird}0.7043 & 0.4533 & 25.2806 & 26.1375 & 1.803s\\
   & LeGrad~\cite{expl:vit:legrad:bousselham25} & 0.1922 & 0.5502 & 0.3580 & 0.2457 & 0.6521 & 0.4065 & 13.4748 & 39.4919 & \cellcolor{rankfirst}\textbf{0.005s}\\
   & \methodabbrev (Ours) &  \cellcolor{ranksecond}\underline{0.1471} & 0.5649 & \cellcolor{rankthird}0.4178 & \cellcolor{ranksecond}\underline{0.1907} & 0.6703 & \cellcolor{rankthird}0.4796 & 11.5998 & 39.7140 & \cellcolor{rankthird}0.016s \\
   & \methodabbrev $T=3$ (Ours) &  \cellcolor{rankfirst}\textbf{0.1261} & \cellcolor{ranksecond}\underline{0.6080} & \cellcolor{rankfirst}\textbf{0.4819} & \cellcolor{rankfirst}\textbf{0.1600} & \cellcolor{rankfirst}\textbf{0.7224} & \cellcolor{rankfirst}\textbf{0.5624} & \cellcolor{rankfirst}\textbf{7.8339} & \cellcolor{rankthird}47.8573 & 0.212s\\
  \midrule
  \multirow{10}{*}{\rotatebox[origin=c]{90}{\textbf{Ground Truth Class}}} & Grad-CAM~\cite{expl:act:gradcam:selvaraju17}  & 0.2185 & 0.4931 & 0.2746 & 0.2647 & 0.5712 & 0.3065 & 63.3098 & 12.5057 & \cellcolor{ranksecond}\underline{0.010s}\\
   & IG~\cite{expl:grad:integratedgradients:sundararajan17}  & 0.1768 & 0.5157 & 0.3389 & 0.2181 & 0.5910 & 0.3729 & 12.7448 & 43.2325 & 0.157s\\
   & Trans-Att.~\cite{expl:vit:transatt:yuan21}  & 0.1886 & 0.5126 & 0.3240 & 0.2306 & 0.5883 & 0.3576 & 15.2576 & 39.6312 & 0.157s\\
   & Bi-Att.~\cite{expl:vit:bidirectionalatt:chen23}  & 0.1762 & 0.5316 & 0.3554 & 0.2177 & 0.6071 & 0.3894 & 15.7758 & 38.8666 & 0.158s\\
   & TIS~\cite{expl:vit:tis:englebert23}  & \cellcolor{rankthird}0.1589 & \cellcolor{rankfirst}\textbf{0.5691} & \cellcolor{ranksecond}\underline{0.4102} & \cellcolor{rankthird}0.1930 & \cellcolor{rankfirst}\textbf{0.6535} & \cellcolor{ranksecond}\underline{0.4605} & \cellcolor{ranksecond}\underline{8.0883} & \cellcolor{rankfirst}\textbf{53.2509} & 0.416s \\
   & ViT-CX~\cite{expl:vit:vitcx:xie23}  & 0.1924 & 0.5111 & 0.3187 & 0.2388 & 0.5806 & 0.3418 & 13.6714 & \cellcolor{rankthird}45.4918 & 0.763s\\
   & MDA~\cite{expl:vit:mda:walker25}  & 0.1866 & \cellcolor{rankthird}0.5443 & 0.3577 & 0.2298 & \cellcolor{rankthird}0.6191 & 0.3893 & 24.5261 & 29.0981 & 1.803s\\
   & LeGrad~\cite{expl:vit:legrad:bousselham25}  & 0.1792 & 0.5124 & 0.3331 & 0.2207 & 0.5876 & 0.3669 & 12.0416 & 43.7111 & \cellcolor{rankfirst}\textbf{0.005s}\\
   & \methodabbrev (Ours)  & \cellcolor{ranksecond}\underline{0.1363} & 0.5250 & \cellcolor{rankthird}0.3887 & \cellcolor{ranksecond}\underline{0.1694} & 0.6000 & \cellcolor{rankthird}0.4306 & \cellcolor{rankthird}10.4213 & 43.8113 & \cellcolor{rankthird}0.016s\\
   & \methodabbrev $T=3$ (Ours)  & \cellcolor{rankfirst}\textbf{0.1171} & \cellcolor{ranksecond}\underline{0.5650} & \cellcolor{rankfirst}\textbf{0.4480} & \cellcolor{rankfirst}\textbf{0.1420} & \cellcolor{ranksecond}\underline{0.6497} & \cellcolor{rankfirst}\textbf{0.5077} & \cellcolor{rankfirst}\textbf{7.0080} & \cellcolor{ranksecond}\underline{51.5807} & 0.212s\\
  \bottomrule
\end{tabular}
}

%% file: figures/TS_eval/fig.tex
\begin{figure}[tb]
    \centering
    \definecolor{tableblue}{RGB}{133,171,214}
    \definecolor{tableorange}{RGB}{214,159,100}
    \begin{subfigure}{0.5\linewidth}
    \begin{tikzpicture}[font=\scriptsize]
    \begin{axis}[
        width=\linewidth,
        height=0.6\linewidth,
        xlabel={Refinement steps $T$},
        ylabel={Insertion AUC $\uparrow$},
        xtick={0,1,2,3,4,5},
        ymin=0.5, ymax=0.7,
        ytick distance=0.05,
        every axis plot/.append style={thick, mark size=1.5pt},
        legend style={draw=none, fill=none},
        legend pos=south east,
        y tick label style={
            /pgf/number format/fixed,
            /pgf/number format/precision=2,
            /pgf/number format/fixed zerofill
        },
    ]
        \addplot[tableblue, mark=*, solid]
            coordinates {(0,0.611208) (1,0.629938) (2,0.640883) (3,0.648540) (4,0.654535) (5,0.656703)};
        \addlegendentry{Pred.}

        \addplot[tableorange, mark=*, solid]
            coordinates {(0,0.565683) (1,0.582627) (2,0.592911) (3,0.600313) (4,0.606295) (5,0.608937)};
        \addlegendentry{GT}
    \end{axis}
    \end{tikzpicture}%
    \caption{}
    \label{fig:ts_eval_insertion_l}
    \end{subfigure}%
    \begin{subfigure}{0.5\linewidth}
    \begin{tikzpicture}[font=\scriptsize]
    \begin{axis}[
        width=\linewidth,
        height=0.6\linewidth,
        xlabel={Perturbation Steps $S$},
        ylabel={Insertion AUC $\uparrow$},
        xtick={4,8,16,32},
        scaled y ticks=false,
        ymin=0.5, ymax=0.7,
        ytick distance=0.05,
        every axis plot/.append style={thick, mark size=1.5pt},
        legend style={draw=none, fill=none},
        legend pos=north east,
        y tick label style={
            /pgf/number format/fixed,
            /pgf/number format/precision=2,
            /pgf/number format/fixed zerofill
        },
    ]

        \addplot[tableblue, mark=*, solid] coordinates {(4,0.5906) (8,0.6082) (16,0.6112) (32,0.6121)};
        \addlegendentry{Pred.}
        \addplot[tableorange, mark=*, solid] coordinates {(4,0.5469) (8,0.5632) (16,0.5657) (32,0.5669)};
        \addlegendentry{GT}
    \end{axis}
    \end{tikzpicture}%
    \caption{}
    \label{fig:ts_eval_insertion_s}
    \end{subfigure}%
\caption{
    Effect of (a) test-time refinement steps $T$ and (b) training perturbation steps $S$ on the Insertion AUC.
    Results are shown for the predicted class (\textcolor{tableblue}{Pred.}) and the ground-truth class (\textcolor{tableorange}{GT}) on the ImageNet validation set (ViT-B/16).
}
    \label{fig:ts_eval_insertion}
\end{figure}
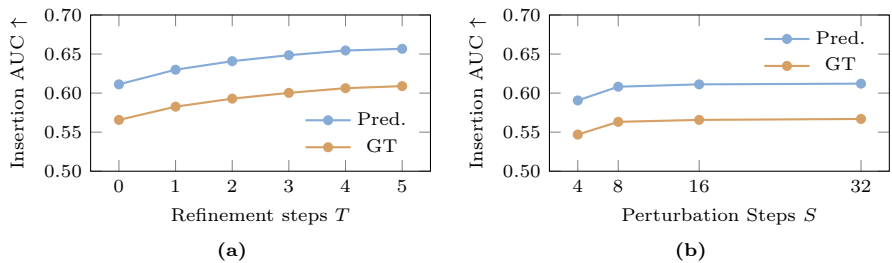

%% file: figures/T_eval_examples/fig.tex
\begin{figure}[tb]
    \centering 
        \input{figures/T_eval_examples/source.tex}
        \caption{
        \textbf{Exemplary test-time refinement.}
        Attribution maps produced by \methodabbrev without ($T{=}0$) and with ($T{=}5$) test-time refinement for the predicted class using ViT-B/16.
        Refinement progressively sharpens the heatmaps, concentrating attribution on the relevant object parts while suppressing diffuse background activations.
    }
    \label{fig:t_eval_examples}
\end{figure}
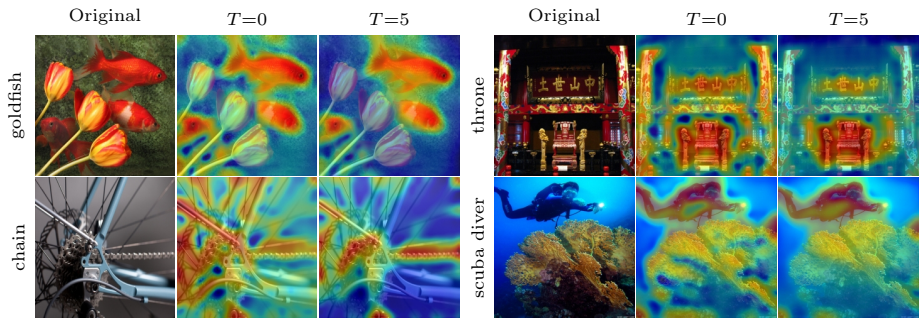

%% file: figures/T_eval_examples/source.tex
\newlength{\teximgw}
\newlength{\texcolsep}
\newlength{\texpairsep}
\setlength{\texcolsep}{1pt}
\setlength{\texpairsep}{14pt}
\setlength{\teximgw}{%
  \dimexpr(\linewidth - 4\texcolsep - \texpairsep) / 6\relax}
\resizebox{\linewidth}{!}{%
\begin{tikzpicture}[
  every node/.style={inner sep=0pt, outer sep=0pt},
]
\node[anchor=south] at (0 + 0.5*\teximgw, 0.15cm) {\scriptsize Original};
\node[anchor=south] at (1*\teximgw + 1*\texcolsep + 0.5*\teximgw, 0.15cm) {\scriptsize $T{=}0$};
\node[anchor=south] at (2*\teximgw + 2*\texcolsep + 0.5*\teximgw, 0.15cm) {\scriptsize $T{=}5$};
\node[anchor=south] at (3*\teximgw + 2*\texcolsep + 1*\texpairsep + 0.5*\teximgw, 0.15cm) {\scriptsize Original};
\node[anchor=south] at (4*\teximgw + 3*\texcolsep + 1*\texpairsep + 0.5*\teximgw, 0.15cm) {\scriptsize $T{=}0$};
\node[anchor=south] at (5*\teximgw + 4*\texcolsep + 1*\texpairsep + 0.5*\teximgw, 0.15cm) {\scriptsize $T{=}5$};
\node[anchor=north west] (img0p0c0) at (0, 0) {\includegraphics[width=\teximgw]{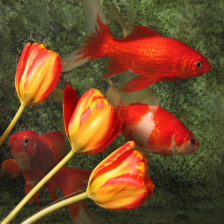}};
\node[anchor=north west] (img0p0c1) at (1*\teximgw + 1*\texcolsep, 0) {\includegraphics[width=\teximgw]{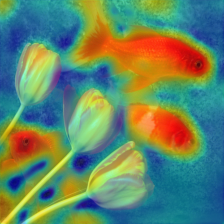}};
\node[anchor=north west] (img0p0c2) at (2*\teximgw + 2*\texcolsep, 0) {\includegraphics[width=\teximgw]{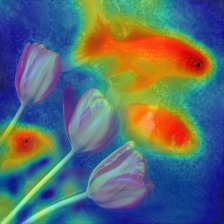}};
\node[rotate=90, anchor=south] at ([xshift=-0.15cm]img0p0c0.west) {\scriptsize goldfish};
\node[anchor=north west] (img0p1c0) at (3*\teximgw + 2*\texcolsep + 1*\texpairsep, 0) {\includegraphics[width=\teximgw]{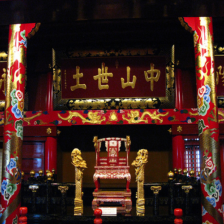}};
\node[anchor=north west] (img0p1c1) at (4*\teximgw + 3*\texcolsep + 1*\texpairsep, 0) {\includegraphics[width=\teximgw]{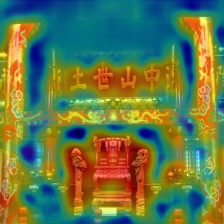}};
\node[anchor=north west] (img0p1c2) at (5*\teximgw + 4*\texcolsep + 1*\texpairsep, 0) {\includegraphics[width=\teximgw]{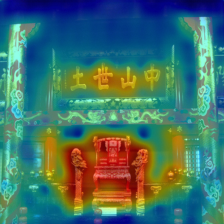}};
\node[rotate=90, anchor=south] at ([xshift=-0.15cm]img0p1c0.west) {\scriptsize throne};
\node[anchor=north west] (img1p0c0) at (0, -1*\teximgw - 1*\texcolsep) {\includegraphics[width=\teximgw]{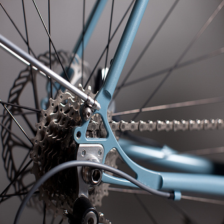}};
\node[anchor=north west] (img1p0c1) at (1*\teximgw + 1*\texcolsep, -1*\teximgw - 1*\texcolsep) {\includegraphics[width=\teximgw]{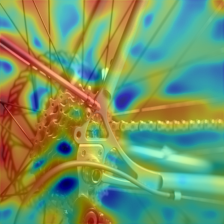}};
\node[anchor=north west] (img1p0c2) at (2*\teximgw + 2*\texcolsep, -1*\teximgw - 1*\texcolsep) {\includegraphics[width=\teximgw]{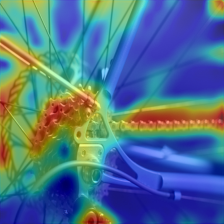}};
\node[rotate=90, anchor=south] at ([xshift=-0.15cm]img1p0c0.west) {\scriptsize chain};
\node[anchor=north west] (img1p1c0) at (3*\teximgw + 2*\texcolsep + 1*\texpairsep, -1*\teximgw - 1*\texcolsep) {\includegraphics[width=\teximgw]{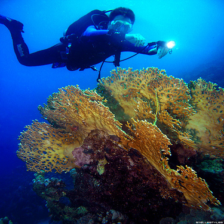}};
\node[anchor=north west] (img1p1c1) at (4*\teximgw + 3*\texcolsep + 1*\texpairsep, -1*\teximgw - 1*\texcolsep) {\includegraphics[width=\teximgw]{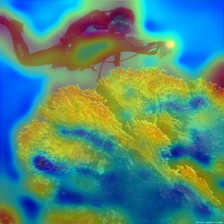}};
\node[anchor=north west] (img1p1c2) at (5*\teximgw + 4*\texcolsep + 1*\texpairsep, -1*\teximgw - 1*\texcolsep) {\includegraphics[width=\teximgw]{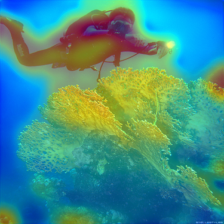}};
\node[rotate=90, anchor=south] at ([xshift=-0.15cm]img1p1c0.west) {\scriptsize scuba diver};
\end{tikzpicture}
}

%% file: sec/5_limitations_conclusion.tex
\section{Conclusion}

We introduced \methodabbrev, a metric-driven attribution framework that directly optimizes differentiable surrogates of the Deletion and Insertion metrics via a Gumbel–Sinkhorn relaxation of sorting. 
The proposed approach combines the causal grounding of perturbation-based methods with the efficiency of amortized explainers, producing dense pixel-level attribution maps in a single forward pass. 
Experiments on ImageNet with Vision Transformers demonstrate improved faithfulness metrics and sharper, boundary-aligned explanations compared to existing methods. 
These results highlight the potential of learning attribution through direct optimization of evaluation metrics for scalable and faithful model explainability.

%% file: sec/A_supp.tex
\section{Appendix}

This appendix complements the main paper with additional implementation details, evaluations, and qualitative results.
We first detail the architecture of the explainer (\cref{sec:supp:architecture}), analyze its training cost and amortization (\cref{sec:supp:training_cost}), and study out-of-distribution data via test-time refinement (\cref{sec:supp:shortcut}).
We then demonstrate the architecture-agnostic applicability by explaining a convolutional classifier (\cref{sec:supp:convnext}), evaluate \methodabbrev under the MDA protocol (\cref{sec:supp:mda_protocol}), on complementary metrics (\cref{sec:supp:complementary_metrics}), and on the ImageNet segmentation benchmark (\cref{sec:supp:segmentation}), and assess cross-architecture explainer transfer (\cref{sec:supp:transfer}).
Finally, we ablate the remaining hyperparameters (\cref{sec:supp:hyperparam_ablation}), report detailed per-reference-image results (\cref{sec:supp:detailed-main-results}), show the full AUC curves (\cref{sec:supp:auc-plots}), and present qualitative results (\cref{sec:supp:qualitative}).\\

\subsection{Explainer Architecture Details}
\label{sec:supp:architecture}

Our attribution model $\Phi_{f,\theta}$ follows an encoder-decoder design.
The encoder is a frozen DINOv3~\cite{model:dinov3:simeoni25} ViT-L/16 backbone.
We extract intermediate features from four evenly spaced transformer blocks (layers 4, 11, 17, 23), each producing spatial feature maps of dimension $1024$.
The decoder follows the Dense Prediction Transformer (DPT)~\cite{model:dpt:ranftl21} design.
At each of the four scales, the patch-level features are concatenated with a learned class embedding $\mathbf{e}_t \in \mathbb{R}^{1024}$, obtained from an embedding layer that maps the target class index $t$ to a dense vector, and projected back to 256 channels via a linear layer followed by GELU activation.
This enables the model to produce different attribution maps depending on the target class.
The four resulting feature maps are progressively fused through residual convolution units and bilinear upsampling, yielding a single 128-channel feature map at the highest resolution.
A lightweight refinement head consisting of a depthwise-separable convolution followed by an additional $3{\times}3$ convolution, each with GELU activation and batch normalization, maps this representation through a final convolutional layer to the single-channel attribution map $\mathbf{A} \in \mathbb{R}^{H \times W}$, where $H$ and $W$ are the height and width of the input image, respectively.
During training, only the decoder parameters and the class embedding are optimized while the DINOv3 backbone remains entirely frozen.\\

\subsection{Training Cost and Amortization}
\label{sec:supp:training_cost}

For each target model, the explainer requires a one-time training.
After that, it efficiently generates attributions during inference, allowing its training costs to be amortized across the generated attributions.
For ViT-B/16, training on ImageNet with $S{=}16$ perturbation steps takes $\sim$$4.5$h on four NVIDIA RTX 6000 GPUs.
\Cref{tab:training_cost} reports the resulting \textit{break-even points}, \ie~the number of attributions after which the total runtime of \methodabbrev (training plus inference) undercuts that of a more time-consuming baseline.
Against the per-sample optimization of MDA~\cite{expl:vit:mda:walker25} this happens after $\sim$$1$k generated attribution maps, against TIS~\cite{expl:vit:tis:englebert23} after $\sim$$14$k and for ViT-CX~\cite{expl:vit:vitcx:xie23} after $\sim$$19$k samples.
Reducing the perturbation steps to $S{=}8$ halves the training time to $\sim$$2$h at only a small quality drop (see Fig. 3(b) main paper), lowering all break-even points accordingly.\\

\begin{table}[t]
\centering
\caption{
\textbf{Training cost amortization.} Break-even point (number of generated attributions) after which the total runtime of \methodabbrev, including the one-time training, falls below that of the baseline. 
Results are shown for ViT-B/16 with $S{=}16$ and $S{=}8$ perturbation steps during training and with $T{=}0$ or $T{=}3$ test-time refinement steps at inference.
Lower is better.
}
\scriptsize
\label{tab:training_cost}
\setlength{\tabcolsep}{8pt}
\begin{tabular}{lcccc}
\toprule
 & \multicolumn{2}{c}{$S{=}16$ (train $\sim$$4.5$h)} & \multicolumn{2}{c}{$S{=}8$ (train $\sim$$2$h)} \\
\cmidrule(lr){2-3}\cmidrule(lr){4-5}
Baseline & $T{=}0$ & $T{=}3$ & $T{=}0$ & $T{=}3$ \\
\midrule
ViT-CX & 19.19k & 51.76k & 8.53k & 23.00k \\
TIS    & 14.07k & 26.13k & 6.25k & 11.61k \\
MDA    & \phantom{0}1.19k & \phantom{0}1.24k & 0.53k & \phantom{0}0.55k \\
\bottomrule
\end{tabular}
\end{table}

\subsection{Out-of-Distribution Attribution via Test-Time Refinement}
\label{sec:supp:shortcut}
To evaluate whether test-time refinement can detect spurious correlations in previously unseen models or data, we construct a \textit{controlled shortcut-learning scenario}:
A ViT-B/16 is fine-tuned on ImageNet for one epoch with a small white circle injected into every training image of the \textit{sunglasses} class.
We choose this class for its low baseline top-1 accuracy, which makes the shortcut effect particularly pronounced.
After fine-tuning, the accuracy on circle-augmented \textit{sunglasses} images rises to 100\%.
Crucially, the \methodabbrev explainer was trained on the original, unmodified ViT-B/16 and has never encountered the white-circle artifact, placing it in an out-of-distribution setting.
At inference, the explainer generates an initial attribution map. 
During refinement steps $T \in \{1, \dots, 5\}$, the fine-tuned model serves as the target model $f$ for backpropagation-based updates.
As shown in \Cref{fig:ood}, at $T{=}0$ the explainer highlights generic class-discriminative features.
As $T$ increases, the refinement progressively shifts the attribution map toward the white circle, revealing the shortcut the fine-tuned model relies on.\\

\input{figures/ood/fig.tex}

\subsection{Architecture-Agnostic Applicability}
\label{sec:supp:convnext}

To demonstrate the universal applicability of \methodabbrev beyond transformer-based classifiers, we apply our approach without any adaptions also to a ConvNeXt Small~\cite{model:convnext:liu22} classifier and compare against established attribution methods applicable to convolutional architectures.
\Cref{tab:main_results_convnext_small} summarizes the results.
Also in this setting, \methodabbrev achieves the best overall faithfulness.
Without refinement ($T{=}0$), our method already reduces the \textit{Deletion} significantly compared to the strongest baseline GradCAM/HiResCAM ($0.2748 \rightarrow 0.1741$, predicted class), indicating that the most-attributed regions are highly relevant for the classifier's decision.
The ADP also drops substantially, confirming a tighter foreground isolation.
With three refinement steps ($T{=}3$), \methodabbrev further improves across all metrics and achieves the highest scores on 6 out of 8 metrics for both target class definitions, including the composite faithfulness scores (\textit{Ins.} - \textit{Del.} and \textit{Neg.} - \textit{Pos.}).
The qualitative examples in \Cref{fig:qual_convnext_small} confirm our findings.
\methodabbrev produces dense, boundary-aligned attribution maps that closely follow the object silhouette, whereas the CAM-based baselines yield coarser heatmaps that often spread into the background.
This is particularly visible for the \textit{German shepherd} and \textit{whistle} example, where our method clearly aligns with the object contours.\\

\input{tables/main_results_convnext_small/table.tex}

\input{tables/mda_eval_results/table}

\subsection{Evaluation under the MDA Protocol}
\label{sec:supp:mda_protocol}

For a fair comparison with MDA~\cite{expl:vit:mda:walker25}, we additionally evaluate \methodabbrev on the exact image subset and protocol of their official codebase.
Unlike our standard evaluation on the full ImageNet validation set (50K images), MDA applies several filters to select a 5{,}000 image subset (5 high-confidence images per class).
Furthermore, while our protocol averages metrics over three reference images $\mathbf{I}_0$ (black, mean, blur), MDA suggests using a single per-image adaptive blur for Insertion and a black image for Deletion.
As shown in \Cref{tab:mda_eval_resuts}, \methodabbrev performs favorably across all metrics under this protocol, with substantially lower runtime compared to~MDA.\\

\subsection{Complementary Faithfulness Metrics}
\label{sec:supp:complementary_metrics}

To probe whether the faithfulness of \methodabbrev generalizes beyond the rank-based \textit{Deletion} and \textit{Insertion} training objective, we evaluate two complementary metrics on the same $5{,}000$-image subset and protocol as MDA~\cite{expl:vit:mda:walker25} (\Cref{sec:supp:mda_protocol}) for the ViT-B/16 classifier.
\Cref{tab:mda_mas_road} shows the magnitude-sensitive MAS~\cite{metric:mas:walker24}, which weights the perturbation by the attribution magnitudes rather than by their rank alone, and the debiased ROAD~\cite{metric:road:rong22}, which replaces removed regions by a noisy linear imputation to reduce information leakage from the mask shape.
$\text{\methodabbrev}_{T=0}$ remains competitive with the per-sample-optimized MDA on the magnitude-based MAS and achieves the best scores on the debiased ROAD metric.\\

\begin{table}[t]
\centering
\caption{
\textbf{Complementary faithfulness metrics.} Magnitude-based MAS~\cite{metric:mas:walker24} and debiased ROAD~\cite{metric:road:rong22} on the $5{,}000$-image MDA subset for the ViT-B/16 classifier. Top-3 results per metric are highlighted (dark to light).}
\label{tab:mda_mas_road}
\scriptsize
\setlength{\tabcolsep}{8pt}
\definecolor{rankfirst}{RGB}{133,171,214}
\definecolor{ranksecond}{RGB}{173,201,230}
\definecolor{rankthird}{RGB}{211,226,243}
\begin{tabular}{lccc|ccc}
\toprule
 & \multicolumn{3}{c}{MAS} & \multicolumn{3}{c}{ROAD}\\
\cmidrule(lr){2-4}\cmidrule(lr){5-7}
Method & Ins.$\uparrow$ & Del.$\downarrow$ & Diff.$\uparrow$ & LeRF$\uparrow$ & MoRF$\downarrow$ & Diff.$\uparrow$\\
\midrule
ViT-CX~\cite{expl:vit:vitcx:xie23}
& 0.582 & 0.408 & 0.174
& 0.664 & 0.201 & 0.463\\
TIS~\cite{expl:vit:tis:englebert23}
& \cellcolor{rankthird}0.615 & \cellcolor{rankthird}0.385 & \cellcolor{rankthird}0.230
& \cellcolor{ranksecond}\underline{0.712} & \cellcolor{ranksecond}\underline{0.169} & \cellcolor{ranksecond}\underline{0.543}\\
MDA~\cite{expl:vit:mda:walker25}
& \cellcolor{rankfirst}\textbf{0.775} & \cellcolor{rankfirst}\textbf{0.279} & \cellcolor{rankfirst}\textbf{0.497}
& \cellcolor{rankthird}0.687 & \cellcolor{rankthird}0.189 & \cellcolor{rankthird}0.498\\
\methodabbrev~($T{=}0$)
& \cellcolor{ranksecond}\underline{0.761} & \cellcolor{ranksecond}\underline{0.290} & \cellcolor{ranksecond}\underline{0.471}
& \cellcolor{rankfirst}\textbf{0.715} & \cellcolor{rankfirst}\textbf{0.160} & \cellcolor{rankfirst}\textbf{0.555}\\
\bottomrule
\end{tabular}
\end{table}

\subsection{Localization via ImageNet Segmentation}
\label{sec:supp:segmentation}

To demonstrate that \methodabbrev learns spatially meaningful, object-aligned attribution maps, we evaluate it on the ImageNet segmentation benchmark~\cite{dataset:imagenetseg:guillaumin14}, treating the thresholded attribution map as a foreground prediction and comparing it against the ground-truth object masks~\cite{expl:vit:transformerinterpretability:chefer21}.
We follow the same setting as MDA~\cite{expl:vit:mda:walker25} for the ViT-B/32 classifier and report the mean average precision~(MAP), the pixel-wise intersection-over-union~(IoU), and the F1 score in \Cref{tab:segmentation}.
$\text{\methodabbrev}_{T=0}$ outperforms ViT-CX~\cite{expl:vit:vitcx:xie23} and TIS~\cite{expl:vit:tis:englebert23} across all three metrics, achieving comparable performance to the per-sample-optimized MDA.
This confirms that the attribution maps generated by \methodabbrev reflect object localizations and are not an artifact of optimizing the \textit{Deletion} and \textit{Insertion} objectives.\\

\begin{table}[t]
\centering
\caption{
\textbf{ImageNet segmentation.} Localization performance on the ImageNet segmentation benchmark~\cite{dataset:imagenetseg:guillaumin14} for the ViT-B/32 classifier, following the MDA~\cite{expl:vit:mda:walker25} setting. Top-3 results per metric are highlighted (dark to light).}
\label{tab:segmentation}
\scriptsize
\setlength{\tabcolsep}{7pt}
\definecolor{rankfirst}{RGB}{133,171,214}
\definecolor{ranksecond}{RGB}{173,201,230}
\definecolor{rankthird}{RGB}{211,226,243}
\begin{tabular}{lccc}
\toprule
Method & Mean Avg. Precision$\uparrow$ & Intersection-over-Union$\uparrow$ & F1 Score$\uparrow$\\
\midrule
ViT-CX~\cite{expl:vit:vitcx:xie23} & 0.678 & 0.535 & 0.428\\
TIS~\cite{expl:vit:tis:englebert23} & \cellcolor{rankthird}0.705 & \cellcolor{rankthird}0.586 & \cellcolor{rankthird}0.460\\
MDA~\cite{expl:vit:mda:walker25} & \cellcolor{rankfirst}\textbf{0.796} & \cellcolor{rankfirst}\textbf{0.702} & \cellcolor{rankfirst}\textbf{0.487}\\
\methodabbrev~($T{=}0$) & \cellcolor{ranksecond}\underline{0.790} & \cellcolor{ranksecond}\underline{0.667} & \cellcolor{ranksecond}\underline{0.483}\\
\bottomrule
\end{tabular}
\end{table}

\subsection{Cross-Architecture Explainer Transfer}
\label{sec:supp:transfer}

Beyond applying \methodabbrev to different architectures by training a dedicated explainer for each (\Cref{sec:supp:convnext}), we further investigate whether a trained explainer transfers to a related architecture without retraining.
This is plausible for correctly-classifying architectures, as they tend to rely on similar object evidence.
In \Cref{tab:cross_arch_transfer} we evaluate, on correctly classified ImageNet samples, how the faithfulness metrics change when an explainer optimized for one ViT is applied to the other, using the per-sample-optimized MDA~\cite{expl:vit:mda:walker25} as a reference point.
$\text{\methodabbrev}_{T{=}0}$ exhibits robust zero-shot transfer, with consistently smaller drops than MDA in both directions.
Moreover, the low-cost $T{=}3$ refinement adapts the transferred explainer to the target architecture and recovers most of the remaining model-specific gap, in some cases even surpassing the native result.
This indicates that trained \methodabbrev explainers provide strong initializations for related architectures, substantially reducing the need for full retraining.\\

\begin{table}[t]
\centering
\definecolor{transfer}{RGB}{242,242,242}
\definecolor{darkred}{RGB}{185,45,45}
\definecolor{darkgreen}{RGB}{45,135,75}
\caption{
    \textbf{Cross-architecture transfer.} 
    RISE~\cite{expl:pert:rise:petsiuk18} and MAS~\cite{metric:mas:walker24} faithfulness metrics (\textit{Insertion}$-$\textit{Deletion} difference) on correctly classified ImageNet samples. 
    Gray rows are the native reference, where the explainer is optimized for the same architecture it is evaluated on.
    In the remaining transfer rows, an explainer trained for a different model was used.
}
\label{tab:cross_arch_transfer}
\scriptsize
\setlength{\tabcolsep}{4pt}
\newcommand{\native}[1]{\cellcolor{transfer}\textcolor{gray}{\textbf{#1}}}
\newcommand{\dropcell}[1]{\textcolor{darkred}{\textbf{#1}}}
\newcommand{\risecell}[1]{\textcolor{darkgreen}{\textbf{#1}}}
\begin{tabular}{l@{\hspace{6pt}}l@{\hspace{5pt}}|@{\hspace{5pt}}c@{\hspace{3pt}}c@{\hspace{3pt}}c|cc|cc}
\toprule
\multirow{2}{*}{} & \multirow{2}{*}{Method}
  & \multirow{2}{*}{Optimized for} & & \multirow{2}{*}{Evaluated on} & \multicolumn{2}{c}{RISE} & \multicolumn{2}{c}{MAS} \\
  & & & & & Diff.$\uparrow$ & $\Delta$ & Diff.$\uparrow$ & $\Delta$ \\
\midrule
\multirow{5}{*}{\rotatebox[origin=c]{90}{ViT-B/16}}
  & \native{MDA} & \native{B/16} & \native{$\rightarrow$} & \native{B/16} & \native{0.624} & \native{--} & \native{0.497} & \native{--} \\
  & MDA & B/32 & $\rightarrow$ & B/16 & 0.473 & \dropcell{-24.2\%} & 0.336 & \dropcell{-32.4\%} \\
\cmidrule{2-9}
  & \native{$\text{\methodabbrev}_{T=0}$} & \native{B/16} & \native{$\rightarrow$} & \native{B/16} & \native{0.715} & \native{--} & \native{0.471} & \native{--} \\
  & $\text{\methodabbrev}_{T=0}$ & B/32 & $\rightarrow$ & B/16 & 0.650 & \dropcell{-9.1\%} & 0.417 & \dropcell{-11.5\%} \\
  & $\text{\methodabbrev}_{T=3}$ & B/32 & $\rightarrow$ & B/16 & 0.739 & \risecell{+3.3\%} & 0.427 & \dropcell{-9.3\%} \\
\midrule
\multirow{5}{*}{\rotatebox[origin=c]{90}{ViT-B/32}}
  & \native{MDA} & \native{B/32} & \native{$\rightarrow$} & \native{B/32} & \native{0.708} & \native{--} & \native{0.616} & \native{--} \\
  & MDA & B/16 & $\rightarrow$ & B/32 & 0.612 & \dropcell{-13.6\%} & 0.509 & \dropcell{-17.4\%} \\
\cmidrule{2-9}
  & \native{$\text{\methodabbrev}_{T=0}$} & \native{B/32} & \native{$\rightarrow$} & \native{B/32} & \native{0.723} & \native{--} & \native{0.589} & \native{--} \\
  & $\text{\methodabbrev}_{T=0}$ & B/16 & $\rightarrow$ & B/32 & 0.683 & \dropcell{-5.5\%} & 0.544 & \dropcell{-7.6\%} \\
  & $\text{\methodabbrev}_{T=3}$ & B/16 & $\rightarrow$ & B/32 & 0.737 & \risecell{+1.9\%} & 0.590 & \risecell{+0.1\%} \\
\bottomrule
\end{tabular}
\end{table}

\subsection{Additional Hyperparameter Ablations}
\label{sec:supp:hyperparam_ablation}

Complementing the test-time refinement ($T$) and perturbation-step ($S$) ablations in the main paper, we ablate the remaining training hyperparameters in \Cref{tab:hyperparam_ablation}.
The grid size $G$ partitions the image into $K{=}G^2$ regions over which the region-level training perturbations are defined.
Although \methodabbrev stays largely stable \wrt the choice of $G$, we sample it uniformly from $\mathcal{U}(7,28)$ at each training iteration anyway, which avoids overfitting to a single perturbation granularity and yields the best faithfulness.
We further emphasize that the grid only defines the region-level perturbations during training: at inference, \methodabbrev always produces a dense, pixel-level attribution map.
The sorting temperature $\tau$ controls the smoothness of the Gumbel-Sinkhorn relaxation.
Here the faithfulness metrics vary only marginally across the tested values, with $\tau{=}1.0$ performing best.\\

\begin{table}[t]
\centering
\caption{
\textbf{Hyperparameter ablations.} Effect of the training grid size $G$ and the sorting temperature $\tau$ on faithfulness (RISE and MAS \textit{Insertion}$-$\textit{Deletion} difference, higher is better) for the ViT-B/16 classifier on correctly classified samples. The configuration used throughout the paper is marked \textit{(used)} and shown in bold.
}
\label{tab:hyperparam_ablation}
\scriptsize
\setlength{\tabcolsep}{5pt}
\begin{tabular}{@{}lcc|lcc@{}}
\toprule
Grid size $G$ & RISE Diff.$\uparrow$ & MAS Diff.$\uparrow$
        & Temp. $\tau$ & RISE Diff.$\uparrow$ & MAS Diff.$\uparrow$\\
\midrule
$G{\sim}\mathcal{U}(7,28)$ \textit{(used)} & \textbf{0.715} & \textbf{0.471}
    & $\tau{=}1.0$ \textit{(used)} & \textbf{0.715} & \textbf{0.471}\\
\midrule
Fixed $G{=}7$  & 0.671 & 0.427
    & $\tau{=}0.5$ & 0.703 & 0.462\\
Fixed $G{=}14$ & 0.698 & 0.453
    & $\tau{=}1.5$ & 0.710 & 0.469\\
Fixed $G{=}28$ & 0.710 & 0.470
    & $\tau{=}2.0$ & 0.708 & 0.467\\
\bottomrule
\end{tabular}
\end{table}

\subsection{Detailed Main Results}
\label{sec:supp:detailed-main-results}

\Cref{tab:main_results_vitb16_detailed,tab:main_results_vitb32_detailed,tab:main_results_convnext_small_detailed} provide the full per-reference-image results for all evaluated classifiers (ViT-B/16, ViT-B/32, ConvNeXt Small).
While the main paper reports metrics averaged over the three reference image modes $\mathbf{I}_0$ (black, mean, blur), these tables list the individual results for each mode separately, allowing for a more fine-grained analysis of the sensitivity to the choice of reference image.\\

\input{tables/main_results_vitb16_detailed/table.tex}
\input{tables/main_results_vitb32_detailed/table.tex}
\input{tables/main_results_convnext_small_detailed/table.tex}

\subsection{AUC Plots}
\label{sec:supp:auc-plots}

\Cref{fig:auc_plots} shows the full Deletion and Insertion AUC curves for ViT-B/16 and ViT-B/32, averaged over all reference images.
The curves illustrate how the target model's confidence changes as pixels are progressively removed (Deletion) or revealed (Insertion) in order of their attributed importance.
\methodabbrev consistently achieves steeper Deletion curves and steeper Insertion curves than the baselines, confirming that its ranking of pixel importance aligns more closely with the classifier's actual reliance on image regions.\\

\input{figures/AUC_plots/fig.tex}

\subsection{Qualitative Results}
\label{sec:supp:qualitative}

\Cref{fig:qual_vitb16,fig:qual_vitb32,fig:qual_convnext_small} present qualitative attribution maps for all evaluated classifiers (ViT-B/16, ViT-B/32, ConvNeXt Small) across a diverse set of ImageNet validation samples, including both correctly and incorrectly classified images.
Across all architectures, \methodabbrev produces sharper, more boundary-aligned attribution maps compared to the baselines, which tend to produce coarser, patch- or region-level activations.\\

\input{figures/qual_vitb16/fig.tex}

\input{figures/qual_vitb32/fig.tex}

\input{figures/qual_convnext_small/fig.tex}

%% file: figures/ood/fig.tex
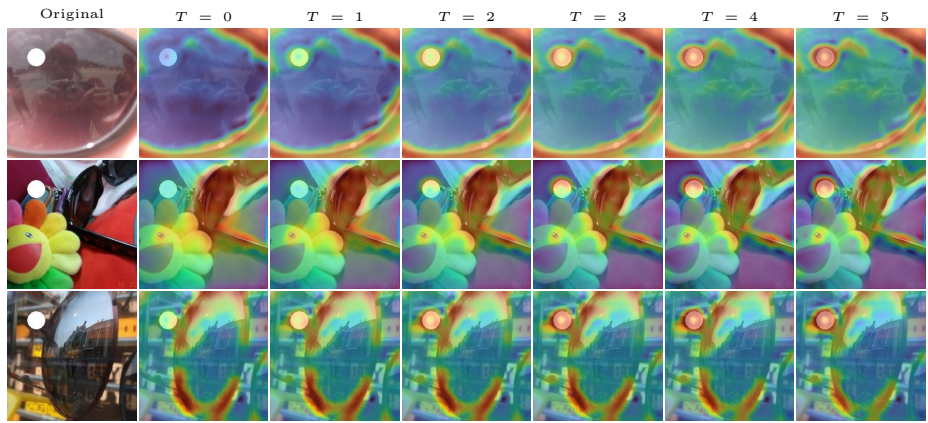
\begin{figure}[tb]
    \centering 
        \input{figures/ood/source.tex}
        \caption{\textbf{Out-of-distribution shortcut detection.} A ViT-B/16 is fine-tuned with a white circle injected into every \textit{sunglasses} training image. Using an \methodabbrev explainer trained on the \emph{clean} model, test-time refinement progressively reveals the learned shortcut: at $T{=}0$ the attribution highlights generic class features, while increasing $T$ reveals the white circle as important feature for the fine-tuned model's decision.}
    \label{fig:ood}
\end{figure}

%% file: figures/ood/source.tex
\newlength{\oodcolsep}
\setlength{\oodcolsep}{1pt}
\newlength{\oodrowsep}
\setlength{\oodrowsep}{1pt}
\newlength{\oodhdrsep}
\setlength{\oodhdrsep}{3pt}
\newlength{\oodimgw}
\setlength{\oodimgw}{\dimexpr(\linewidth - 6\oodcolsep)/7\relax}

\resizebox{\linewidth}{!}{%
\begin{tikzpicture}[
    every node/.style={inner sep=0pt, outer sep=0pt},
]
  \node[anchor=south, font=\tiny, text width=\oodimgw, align=center] at (0*\oodimgw + 0*\oodcolsep + 0.5*\oodimgw, \oodhdrsep) {Original};
  \node[anchor=south, font=\tiny, text width=\oodimgw, align=center] at (1*\oodimgw + 1*\oodcolsep + 0.5*\oodimgw, \oodhdrsep) {$T=0$};
  \node[anchor=south, font=\tiny, text width=\oodimgw, align=center] at (2*\oodimgw + 2*\oodcolsep + 0.5*\oodimgw, \oodhdrsep) {$T=1$};
  \node[anchor=south, font=\tiny, text width=\oodimgw, align=center] at (3*\oodimgw + 3*\oodcolsep + 0.5*\oodimgw, \oodhdrsep) {$T=2$};
  \node[anchor=south, font=\tiny, text width=\oodimgw, align=center] at (4*\oodimgw + 4*\oodcolsep + 0.5*\oodimgw, \oodhdrsep) {$T=3$};
  \node[anchor=south, font=\tiny, text width=\oodimgw, align=center] at (5*\oodimgw + 5*\oodcolsep + 0.5*\oodimgw, \oodhdrsep) {$T=4$};
  \node[anchor=south, font=\tiny, text width=\oodimgw, align=center] at (6*\oodimgw + 6*\oodcolsep + 0.5*\oodimgw, \oodhdrsep) {$T=5$};
  \node[anchor=north west] at (0*\oodimgw + 0*\oodcolsep, -0*\oodimgw - 0*\oodrowsep) {\includegraphics[width=\oodimgw]{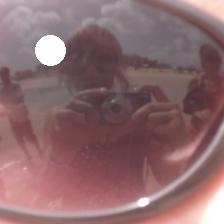}};
  \node[anchor=north west] at (1*\oodimgw + 1*\oodcolsep, -0*\oodimgw - 0*\oodrowsep) {\includegraphics[width=\oodimgw]{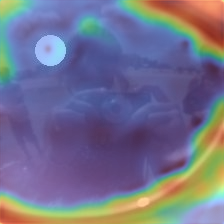}};
  \node[anchor=north west] at (2*\oodimgw + 2*\oodcolsep, -0*\oodimgw - 0*\oodrowsep) {\includegraphics[width=\oodimgw]{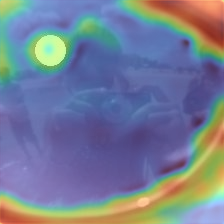}};
  \node[anchor=north west] at (3*\oodimgw + 3*\oodcolsep, -0*\oodimgw - 0*\oodrowsep) {\includegraphics[width=\oodimgw]{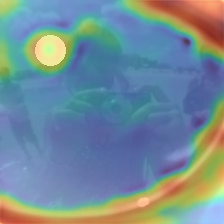}};
  \node[anchor=north west] at (4*\oodimgw + 4*\oodcolsep, -0*\oodimgw - 0*\oodrowsep) {\includegraphics[width=\oodimgw]{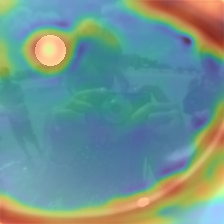}};
  \node[anchor=north west] at (5*\oodimgw + 5*\oodcolsep, -0*\oodimgw - 0*\oodrowsep) {\includegraphics[width=\oodimgw]{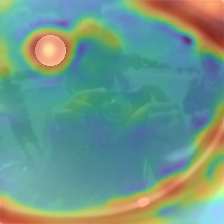}};
  \node[anchor=north west] at (6*\oodimgw + 6*\oodcolsep, -0*\oodimgw - 0*\oodrowsep) {\includegraphics[width=\oodimgw]{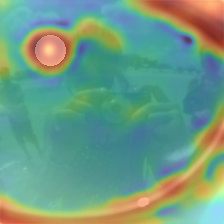}};
  \node[anchor=north west] at (0*\oodimgw + 0*\oodcolsep, -1*\oodimgw - 1*\oodrowsep) {\includegraphics[width=\oodimgw]{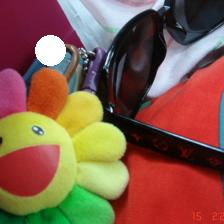}};
  \node[anchor=north west] at (1*\oodimgw + 1*\oodcolsep, -1*\oodimgw - 1*\oodrowsep) {\includegraphics[width=\oodimgw]{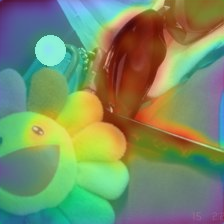}};
  \node[anchor=north west] at (2*\oodimgw + 2*\oodcolsep, -1*\oodimgw - 1*\oodrowsep) {\includegraphics[width=\oodimgw]{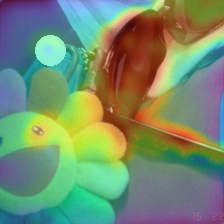}};
  \node[anchor=north west] at (3*\oodimgw + 3*\oodcolsep, -1*\oodimgw - 1*\oodrowsep) {\includegraphics[width=\oodimgw]{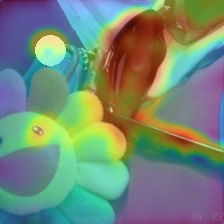}};
  \node[anchor=north west] at (4*\oodimgw + 4*\oodcolsep, -1*\oodimgw - 1*\oodrowsep) {\includegraphics[width=\oodimgw]{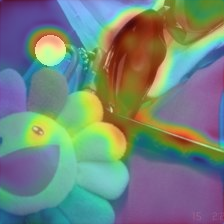}};
  \node[anchor=north west] at (5*\oodimgw + 5*\oodcolsep, -1*\oodimgw - 1*\oodrowsep) {\includegraphics[width=\oodimgw]{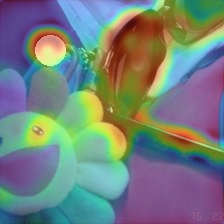}};
  \node[anchor=north west] at (6*\oodimgw + 6*\oodcolsep, -1*\oodimgw - 1*\oodrowsep) {\includegraphics[width=\oodimgw]{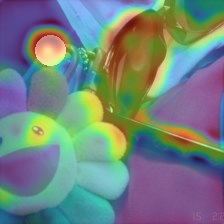}};
  \node[anchor=north west] at (0*\oodimgw + 0*\oodcolsep, -2*\oodimgw - 2*\oodrowsep) {\includegraphics[width=\oodimgw]{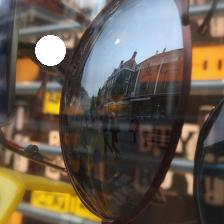}};
  \node[anchor=north west] at (1*\oodimgw + 1*\oodcolsep, -2*\oodimgw - 2*\oodrowsep) {\includegraphics[width=\oodimgw]{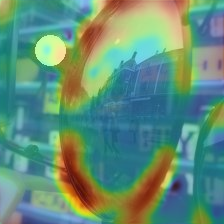}};
  \node[anchor=north west] at (2*\oodimgw + 2*\oodcolsep, -2*\oodimgw - 2*\oodrowsep) {\includegraphics[width=\oodimgw]{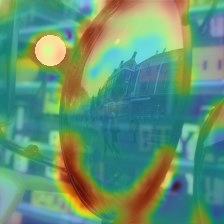}};
  \node[anchor=north west] at (3*\oodimgw + 3*\oodcolsep, -2*\oodimgw - 2*\oodrowsep) {\includegraphics[width=\oodimgw]{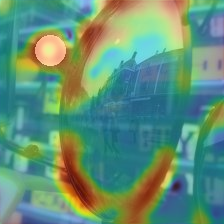}};
  \node[anchor=north west] at (4*\oodimgw + 4*\oodcolsep, -2*\oodimgw - 2*\oodrowsep) {\includegraphics[width=\oodimgw]{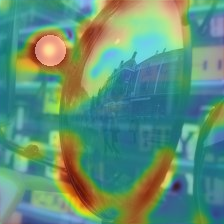}};
  \node[anchor=north west] at (5*\oodimgw + 5*\oodcolsep, -2*\oodimgw - 2*\oodrowsep) {\includegraphics[width=\oodimgw]{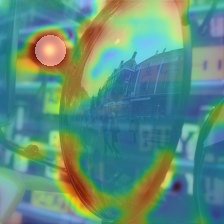}};
  \node[anchor=north west] at (6*\oodimgw + 6*\oodcolsep, -2*\oodimgw - 2*\oodrowsep) {\includegraphics[width=\oodimgw]{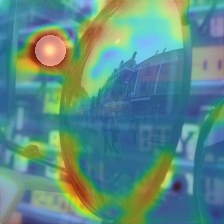}};
\end{tikzpicture}}

%% file: tables/main_results_convnext_small/table.tex
\begin{table*}[t]
    \caption{
        Quantitative comparison of attribution methods on ImageNet using a ConvNeXt Small classifier~\cite{model:convnext:liu22}. 
        All metrics are averaged over three reference image modes~$\mathbf{I}_0$ (black, mean, blur). Top-3 results per metric are highlighted (dark to light).
    }
    \input{tables/main_results_convnext_small/source.tex}
    \label{tab:main_results_convnext_small}
\end{table*}

%% file: tables/main_results_convnext_small/source.tex
\small
\setlength{\tabcolsep}{4pt}
\definecolor{rankfirst}{RGB}{133,171,214}
\definecolor{ranksecond}{RGB}{173,201,230}
\definecolor{rankthird}{RGB}{211,226,243}
\resizebox{\textwidth}{!}{%
\begin{tabular}{clccccccccr}
  \toprule
   & Method & Deletion$\downarrow$ & Insertion$\uparrow$ & Ins. - Del.$\uparrow$ & Positive$\downarrow$ & Negative$\uparrow$ & Neg. - Pos.$\uparrow$ & ADP$\downarrow$ & PIC$\uparrow$ & Runtime \\
  \midrule
  \multirow{7}{*}{\rotatebox[origin=c]{90}{\textbf{Pred. Class}}} & ScoreCAM~\cite{expl:act:scorecam:wang20} & 0.3779 & 0.5425 & 0.1646 & 0.5348 & 0.7367 & 0.2019 & 56.0312 & \cellcolor{rankthird}5.8985 & 0.821s \\
   & AblationCAM~\cite{expl:act:ablationcam:desai20} & 0.3340 & 0.5890 & 0.2550 & 0.4753 & 0.7877 & 0.3124 & 48.9473 & 4.0993 & 0.742s \\
   & GradCAM++~\cite{expl:act:gradcampp:chattopadhay18} & 0.3159 & 0.6008 & 0.2848 & 0.4635 & \cellcolor{rankthird}0.8231 & 0.3597 & 50.9349 & 4.3820 & \cellcolor{rankthird}0.019s \\
   & HiResCAM~\cite{expl:act:hirescam:draelos20} & 0.2748 & \cellcolor{ranksecond}\underline{0.6267} & 0.3520 & 0.3908 & \cellcolor{ranksecond}\underline{0.8314} & 0.4406 & 41.5140 & 5.4472 & 0.019s \\
   & GradCAM~\cite{expl:act:gradcam:selvaraju17} & \cellcolor{rankthird}0.2748 & \cellcolor{rankfirst}\textbf{0.6267} & \cellcolor{rankthird}0.3520 & \cellcolor{rankthird}0.3908 & \cellcolor{rankfirst}\textbf{0.8314} & \cellcolor{rankthird}0.4406 & \cellcolor{rankthird}41.5133 & 5.4453 & \cellcolor{rankfirst}\textbf{0.004s} \\
   & \methodabbrev (Ours) & \cellcolor{ranksecond}\underline{0.1741} & 0.5911 & \cellcolor{ranksecond}\underline{0.4170} & \cellcolor{ranksecond}\underline{0.2408} & 0.7616 & \cellcolor{ranksecond}\underline{0.5208} & \cellcolor{ranksecond}\underline{23.6857} & \cellcolor{ranksecond}\underline{6.0192} & \cellcolor{ranksecond}\underline{0.005s} \\
   & \methodabbrev $T=3$ (Ours) & \cellcolor{rankfirst}\textbf{0.1576} & \cellcolor{rankthird}0.6079 & \cellcolor{rankfirst}\textbf{0.4503} & \cellcolor{rankfirst}\textbf{0.2150} & 0.7842 & \cellcolor{rankfirst}\textbf{0.5692} & \cellcolor{rankfirst}\textbf{22.1279} & \cellcolor{rankfirst}\textbf{7.1759} & 0.313s \\
  \midrule
  \multirow{7}{*}{\rotatebox[origin=c]{90}{\textbf{GT Class}}} & ScoreCAM~\cite{expl:act:scorecam:wang20} & 0.3520 & 0.5001 & 0.1481 & 0.4856 & 0.6595 & 0.1739 & 52.7301 & 11.0598 & 0.821s \\
   & AblationCAM~\cite{expl:act:ablationcam:desai20} & 0.3117 & 0.5429 & 0.2313 & 0.4336 & 0.7023 & 0.2688 & 44.5182 & 10.6520 & 0.742s \\
   & GradCAM++~\cite{expl:act:gradcampp:chattopadhay18} & 0.2959 & 0.5533 & 0.2573 & 0.4258 & \cellcolor{rankthird}0.7331 & 0.3074 & 45.5688 & 11.9073 & \cellcolor{rankthird}0.019s \\
   & HiResCAM~\cite{expl:act:hirescam:draelos20} & 0.2578 & \cellcolor{ranksecond}\underline{0.5771} & 0.3193 & 0.3600 & \cellcolor{ranksecond}\underline{0.7415} & 0.3815 & 37.0658 & \cellcolor{ranksecond}\underline{13.1477} & 0.019s \\
   & GradCAM~\cite{expl:act:gradcam:selvaraju17} & \cellcolor{rankthird}0.2578 & \cellcolor{rankfirst}\textbf{0.5771} & \cellcolor{rankthird}0.3193 & \cellcolor{rankthird}0.3600 & \cellcolor{rankfirst}\textbf{0.7415} & \cellcolor{rankthird}0.3815 & \cellcolor{rankthird}37.0650 & \cellcolor{rankthird}13.1467 & \cellcolor{rankfirst}\textbf{0.004s} \\
   & \methodabbrev (Ours) & \cellcolor{ranksecond}\underline{0.1624} & 0.5467 & \cellcolor{ranksecond}\underline{0.3843} & \cellcolor{ranksecond}\underline{0.2191} & 0.6828 & \cellcolor{ranksecond}\underline{0.4637} & \cellcolor{ranksecond}\underline{21.2367} & 12.6424 & \cellcolor{ranksecond}\underline{0.005s} \\
   & \methodabbrev $T=3$ (Ours) & \cellcolor{rankfirst}\textbf{0.1472} & \cellcolor{rankthird}0.5613 & \cellcolor{rankfirst}\textbf{0.4141} & \cellcolor{rankfirst}\textbf{0.1957} & 0.7028 & \cellcolor{rankfirst}\textbf{0.5071} & \cellcolor{rankfirst}\textbf{19.7873} & \cellcolor{rankfirst}\textbf{14.0637} & 0.313s \\
  \bottomrule
\end{tabular}
}

%% file: tables/mda_eval_results/table.tex
\begin{table*}[t]
    \centering
    \caption{
        Quantitative comparison using the MDA evaluation framework~\cite{expl:vit:mda:walker25} for a ViT-B/16 classifier on their filtered ImageNet subset (5{,}000 images). Top-3 results per
        metric are highlighted (dark to light).
    }
    \scriptsize
    \setlength{\tabcolsep}{4pt}
    \definecolor{rankfirst}{RGB}{133,171,214}
    \definecolor{ranksecond}{RGB}{173,201,230}
    \definecolor{rankthird}{RGB}{211,226,243}
    \begin{tabular}{lccccccccr}
    \toprule
    Method  & Deletion$\downarrow$ & Insertion$\uparrow$ & Insertion - Deletion$\uparrow$ & Runtime\\
    \midrule
    Grad-CAM~\cite{expl:act:gradcam:selvaraju17} & 0.241 & 0.737 & 0.496 & \cellcolor{rankfirst}\textbf{0.014s}\\
    IG~\cite{expl:grad:integratedgradients:sundararajan17}  & 0.222 & 0.741 & 0.518 & \cellcolor{rankthird} 0.251s\\
    ViT-CX~\cite{expl:vit:vitcx:xie23}  & 0.236 & 0.722 & 0.486 & 0.860s \\
    Trans-Att.~\cite{expl:vit:transatt:yuan21}  & 0.228 & 0.748 & 0.520 & 0.254s\\
    Bi-Att.~\cite{expl:vit:bidirectionalatt:chen23}  & 0.218 & 0.760 & 0.542  & 0.254s\\
    TIS~\cite{expl:vit:tis:englebert23}  & \cellcolor{rankthird} 0.196 & 0.761 & 0.565  & 1.167s\\
    MDA~\cite{expl:vit:mda:walker25}  & 0.232 & \cellcolor{rankthird} 0.856 & \cellcolor{rankthird} 0.624  & 13.645s \\
    \methodabbrev (Ours) &  \cellcolor{ranksecond}\underline{0.147} & \cellcolor{ranksecond}\underline{0.862} & \cellcolor{ranksecond}\underline{0.716} & \cellcolor{ranksecond}\underline{0.016s}\\
    \methodabbrev $T=3$ (Ours) & \cellcolor{rankfirst}\textbf{0.131} & \cellcolor{rankfirst}\textbf{0.886} & \cellcolor{rankfirst}\textbf{0.756} & 0.547s\\
    \bottomrule
    \end{tabular}
    \label{tab:mda_eval_resuts}
\end{table*}

%% file: tables/main_results_vitb16_detailed/table.tex
\begin{table*}[t]
    \caption{
        Quantitative comparison of attribution methods on ImageNet using a \mbox{ViT-B/16} classifier~\cite{classif:vit:dosovitskiy21} for all reference image modes~$\mathbf{I}_0$ (black image, mean-colored image, blurred original image). Top-3 results per metric are highlighted (dark to light). 
    }
    \input{tables/main_results_vitb16_detailed/source.tex}
    \label{tab:main_results_vitb16_detailed}
\end{table*}

%% file: tables/main_results_vitb16_detailed/source.tex
\small
\setlength{\tabcolsep}{4pt}
\definecolor{rankfirst}{RGB}{133,171,214}
\definecolor{ranksecond}{RGB}{173,201,230}
\definecolor{rankthird}{RGB}{211,226,243}
\resizebox{\textwidth}{!}{%
\begin{tabular}{cclccccccccr}
  \toprule
   &  & Method & Deletion$\downarrow$ & Insertion$\uparrow$ & Ins. - Del.$\uparrow$ & Positive$\downarrow$ & Negative$\uparrow$ & Neg. - Pos.$\uparrow$ & ADP$\downarrow$ & PIC$\uparrow$ & Runtime \\
  \midrule
  \multirow{20}{*}{\rotatebox[origin=c]{90}{\textbf{$\mathbf{I}_0 \equiv$ Black Image}}} & \multirow{10}{*}{\rotatebox[origin=c]{90}{\textbf{Predicted Class}}} & Grad-CAM~\cite{expl:act:gradcam:selvaraju17} & 0.2542 & 0.4562 & 0.2020 & 0.3024 & 0.5336 & 0.2312 & 90.2340 & 2.6849 & \cellcolor{ranksecond}\underline{0.014s} \\
   &  & IG~\cite{expl:grad:integratedgradients:sundararajan17} & 0.1555 & 0.5063 & 0.3508 & 0.1899 & 0.5857 & 0.3958 & 26.4955 & 30.3434 & 0.251s \\
   &  & Trans-Att.~\cite{expl:vit:transatt:yuan21} & 0.1402 & 0.5380 & 0.3978 & 0.1726 & 0.6189 & 0.4463 & 29.7100 & 27.5834 & 0.254s \\
   &  & Bi-Att.~\cite{expl:vit:bidirectionalatt:chen23} & 0.1355 & 0.5533 & 0.4178 & 0.1669 & 0.6348 & 0.4678 & 32.5737 & 25.3995 & 0.254s \\
   &  & TIS~\cite{expl:vit:tis:englebert23} & \cellcolor{rankthird}0.1205 & \cellcolor{ranksecond}\underline{0.5976} & \cellcolor{ranksecond}\underline{0.4771} & \cellcolor{ranksecond}\underline{0.1474} & \cellcolor{ranksecond}\underline{0.6807} & \cellcolor{ranksecond}\underline{0.5332} & \cellcolor{rankfirst}\textbf{14.1945} & \cellcolor{rankfirst}\textbf{42.2192} & 1.167s \\
   &  & ViT-CX~\cite{expl:vit:vitcx:xie23} & 0.1488 & 0.5269 & 0.3781 & 0.1845 & 0.6019 & 0.4174 & \cellcolor{rankthird}17.1795 & \cellcolor{rankthird}37.5492 & 0.860s \\
   &  & MDA~\cite{expl:vit:mda:walker25} & 0.1336 & \cellcolor{rankthird}0.5769 & 0.4433 & 0.1669 & \cellcolor{rankthird}0.6617 & 0.4948 & 59.7938 & 8.3518 & 13.645s \\
   &  & LeGrad~\cite{expl:vit:legrad:bousselham25} & 0.1390 & 0.5189 & 0.3799 & 0.1715 & 0.5970 & 0.4254 & 24.2647 & 31.1134 & \cellcolor{rankfirst}\textbf{0.006s} \\
   &  & \methodabbrev (Ours) & \cellcolor{ranksecond}\underline{0.1195} & 0.5691 & \cellcolor{rankthird}0.4496 & \cellcolor{rankthird}0.1489 & 0.6502 & \cellcolor{rankthird}0.5013 & 18.1877 & 35.0353 & \cellcolor{rankthird}0.016s \\
   &  & \methodabbrev $T=3$ (Ours) & \cellcolor{rankfirst}\textbf{0.1079} & \cellcolor{rankfirst}\textbf{0.6005} & \cellcolor{rankfirst}\textbf{0.4926} & \cellcolor{rankfirst}\textbf{0.1336} & \cellcolor{rankfirst}\textbf{0.6883} & \cellcolor{rankfirst}\textbf{0.5547} & \cellcolor{ranksecond}\underline{14.8003} & \cellcolor{ranksecond}\underline{39.8492} & 0.547s \\
  \cmidrule{2-12}
   & \multirow{10}{*}{\rotatebox[origin=c]{90}{\textbf{Ground Truth Class}}} & Grad-CAM~\cite{expl:act:gradcam:selvaraju17} & 0.2398 & 0.4252 & 0.1854 & 0.2780 & 0.4845 & 0.2065 & 86.7297 & 5.5450 & \cellcolor{ranksecond}\underline{0.014s} \\
   &  & IG~\cite{expl:grad:integratedgradients:sundararajan17} & 0.1449 & 0.4744 & 0.3295 & 0.1711 & 0.5351 & 0.3641 & 23.2480 & 35.6353 & 0.251s \\
   &  & Trans-Att.~\cite{expl:vit:transatt:yuan21} & 0.1304 & 0.5032 & 0.3728 & 0.1548 & 0.5645 & 0.4097 & 26.3366 & 32.8113 & 0.254s \\
   &  & Bi-Att.~\cite{expl:vit:bidirectionalatt:chen23} & 0.1261 & 0.5165 & 0.3904 & 0.1497 & 0.5771 & 0.4274 & 28.9729 & 30.6954 & 0.254s \\
   &  & TIS~\cite{expl:vit:tis:englebert23} & \cellcolor{rankthird}0.1123 & \cellcolor{rankfirst}\textbf{0.5600} & \cellcolor{ranksecond}\underline{0.4478} & \cellcolor{ranksecond}\underline{0.1317} & \cellcolor{rankfirst}\textbf{0.6241} & \cellcolor{ranksecond}\underline{0.4924} & \cellcolor{rankfirst}\textbf{11.7329} & \cellcolor{rankfirst}\textbf{47.9150} & 1.167s \\
   &  & ViT-CX~\cite{expl:vit:vitcx:xie23} & 0.1436 & 0.4852 & 0.3416 & 0.1730 & 0.5384 & 0.3653 & 18.1427 & 38.4452 & 0.860s \\
   &  & MDA~\cite{expl:vit:mda:walker25} & 0.1251 & \cellcolor{rankthird}0.5332 & 0.4081 & 0.1508 & \cellcolor{rankthird}0.5928 & 0.4420 & 55.2688 & 13.5977 & 13.645s \\
   &  & LeGrad~\cite{expl:vit:legrad:bousselham25} & 0.1293 & 0.4858 & 0.3565 & 0.1538 & 0.5449 & 0.3912 & 21.1425 & 36.3753 & \cellcolor{rankfirst}\textbf{0.006s} \\
   &  & \methodabbrev (Ours) & \cellcolor{ranksecond}\underline{0.1107} & 0.5288 & \cellcolor{rankthird}0.4181 & \cellcolor{rankthird}0.1325 & 0.5865 & \cellcolor{rankthird}0.4540 & \cellcolor{rankthird}16.1274 & \cellcolor{rankthird}39.5132 & \cellcolor{rankthird}0.016s \\
   &  & \methodabbrev $T=3$ (Ours) & \cellcolor{rankfirst}\textbf{0.1000} & \cellcolor{ranksecond}\underline{0.5577} & \cellcolor{rankfirst}\textbf{0.4577} & \cellcolor{rankfirst}\textbf{0.1186} & \cellcolor{ranksecond}\underline{0.6218} & \cellcolor{rankfirst}\textbf{0.5032} & \cellcolor{ranksecond}\underline{12.9713} & \cellcolor{ranksecond}\underline{44.3711} & 0.547s \\
  \midrule
  \multirow{20}{*}{\rotatebox[origin=c]{90}{\textbf{$\mathbf{I}_0 \equiv$ Mean-colored Image}}} & \multirow{10}{*}{\rotatebox[origin=c]{90}{\textbf{Predicted Class}}} & Grad-CAM~\cite{expl:act:gradcam:selvaraju17} & 0.2802 & 0.4881 & 0.2079 & 0.3269 & 0.5684 & 0.2415 & 86.3553 & 4.0605 & \cellcolor{ranksecond}\underline{0.014s} \\
   &  & IG~\cite{expl:grad:integratedgradients:sundararajan17} & 0.1751 & 0.5429 & 0.3678 & 0.2123 & 0.6229 & 0.4106 & 23.2566 & 33.1053 & 0.251s \\
   &  & Trans-Att.~\cite{expl:vit:transatt:yuan21} & 0.1551 & 0.5730 & 0.4179 & 0.1893 & 0.6547 & 0.4654 & 25.4997 & 30.9174 & 0.254s \\
   &  & Bi-Att.~\cite{expl:vit:bidirectionalatt:chen23} & 0.1506 & 0.5853 & 0.4347 & 0.1844 & 0.6676 & 0.4832 & 28.1998 & 28.2094 & 0.254s \\
   &  & TIS~\cite{expl:vit:tis:englebert23} & \cellcolor{rankthird}0.1331 & \cellcolor{ranksecond}\underline{0.6283} & \cellcolor{ranksecond}\underline{0.4952} & \cellcolor{ranksecond}\underline{0.1619} & \cellcolor{ranksecond}\underline{0.7117} & \cellcolor{ranksecond}\underline{0.5497} & \cellcolor{ranksecond}\underline{11.5112} & \cellcolor{ranksecond}\underline{44.4791} & 1.167s \\
   &  & ViT-CX~\cite{expl:vit:vitcx:xie23} & 0.1638 & 0.5624 & 0.3987 & 0.2006 & 0.6384 & 0.4377 & \cellcolor{rankfirst}\textbf{10.7794} & \cellcolor{rankfirst}\textbf{45.6811} & 0.860s \\
   &  & MDA~\cite{expl:vit:mda:walker25} & 0.1410 & \cellcolor{rankthird}0.6006 & 0.4596 & 0.1754 & \cellcolor{rankthird}0.6869 & 0.5115 & 49.9938 & 12.5057 & 13.645s \\
   &  & LeGrad~\cite{expl:vit:legrad:bousselham25} & 0.1554 & 0.5565 & 0.4011 & 0.1902 & 0.6358 & 0.4456 & 20.4968 & 33.7513 & \cellcolor{rankfirst}\textbf{0.006s} \\
   &  & \methodabbrev (Ours) & \cellcolor{ranksecond}\underline{0.1327} & 0.5985 & \cellcolor{rankthird}0.4658 & \cellcolor{rankthird}0.1647 & 0.6808 & \cellcolor{rankthird}0.5161 & 16.3669 & 36.0993 & \cellcolor{rankthird}0.016s \\
   &  & \methodabbrev $T=3$ (Ours) & \cellcolor{rankfirst}\textbf{0.1177} & \cellcolor{rankfirst}\textbf{0.6356} & \cellcolor{rankfirst}\textbf{0.5179} & \cellcolor{rankfirst}\textbf{0.1446} & \cellcolor{rankfirst}\textbf{0.7249} & \cellcolor{rankfirst}\textbf{0.5803} & \cellcolor{rankthird}12.6456 & \cellcolor{rankthird}41.8452 & 0.547s \\
  \cmidrule{2-12}
   & \multirow{10}{*}{\rotatebox[origin=c]{90}{\textbf{Ground Truth Class}}} & Grad-CAM~\cite{expl:act:gradcam:selvaraju17} & 0.2638 & 0.4545 & 0.1907 & 0.2999 & 0.5156 & 0.2157 & 82.4823 & 7.3558 & \cellcolor{ranksecond}\underline{0.014s} \\
   &  & IG~\cite{expl:grad:integratedgradients:sundararajan17} & 0.1631 & 0.5068 & 0.3437 & 0.1910 & 0.5667 & 0.3757 & 20.2897 & 38.2752 & 0.251s \\
   &  & Trans-Att.~\cite{expl:vit:transatt:yuan21} & 0.1443 & 0.5337 & 0.3894 & 0.1698 & 0.5935 & 0.4237 & 22.3643 & 36.1313 & 0.254s \\
   &  & Bi-Att.~\cite{expl:vit:bidirectionalatt:chen23} & 0.1402 & 0.5443 & 0.4041 & 0.1655 & 0.6034 & 0.4379 & 24.8366 & 33.6413 & 0.254s \\
   &  & TIS~\cite{expl:vit:tis:englebert23} & \cellcolor{rankthird}0.1240 & \cellcolor{ranksecond}\underline{0.5877} & \cellcolor{ranksecond}\underline{0.4638} & \cellcolor{ranksecond}\underline{0.1448} & \cellcolor{ranksecond}\underline{0.6511} & \cellcolor{ranksecond}\underline{0.5063} & \cellcolor{rankfirst}\textbf{9.3110} & \cellcolor{rankfirst}\textbf{49.9430} & 1.167s \\
   &  & ViT-CX~\cite{expl:vit:vitcx:xie23} & 0.1585 & 0.5150 & 0.3565 & 0.1889 & 0.5670 & 0.3781 & 14.5135 & \cellcolor{rankthird}43.0011 & 0.860s \\
   &  & MDA~\cite{expl:vit:mda:walker25} & 0.1322 & 0.5537 & 0.4215 & 0.1590 & \cellcolor{rankthird}0.6132 & 0.4542 & 46.0686 & 17.5416 & 13.645s \\
   &  & LeGrad~\cite{expl:vit:legrad:bousselham25} & 0.1447 & 0.5191 & 0.3744 & 0.1710 & 0.5775 & 0.4065 & 17.7684 & 38.9052 & \cellcolor{rankfirst}\textbf{0.006s} \\
   &  & \methodabbrev (Ours) & \cellcolor{ranksecond}\underline{0.1226} & \cellcolor{rankthird}0.5552 & \cellcolor{rankthird}0.4326 & \cellcolor{rankthird}0.1461 & 0.6126 & \cellcolor{rankthird}0.4664 & \cellcolor{rankthird}14.3323 & 40.5832 & \cellcolor{rankthird}0.016s \\
   &  & \methodabbrev $T=3$ (Ours) & \cellcolor{rankfirst}\textbf{0.1089} & \cellcolor{rankfirst}\textbf{0.5895} & \cellcolor{rankfirst}\textbf{0.4806} & \cellcolor{rankfirst}\textbf{0.1282} & \cellcolor{rankfirst}\textbf{0.6540} & \cellcolor{rankfirst}\textbf{0.5259} & \cellcolor{ranksecond}\underline{10.8600} & \cellcolor{ranksecond}\underline{46.6231} & 0.547s \\
  \midrule
  \multirow{20}{*}{\rotatebox[origin=c]{90}{\textbf{$\mathbf{I}_0 \equiv$ Blurred Original Image}}} & \multirow{10}{*}{\rotatebox[origin=c]{90}{\textbf{Predicted Class}}} & Grad-CAM~\cite{expl:act:gradcam:selvaraju17} & 0.3230 & 0.5503 & 0.2273 & 0.3744 & 0.6343 & 0.2600 & 76.4122 & 6.3277 & \cellcolor{ranksecond}\underline{0.014s} \\
   &  & IG~\cite{expl:grad:integratedgradients:sundararajan17} & 0.2266 & 0.6139 & 0.3872 & 0.2722 & 0.6956 & 0.4234 & 19.0499 & 36.8573 & 0.251s \\
   &  & Trans-Att.~\cite{expl:vit:transatt:yuan21} & 0.2011 & 0.6388 & 0.4377 & 0.2447 & 0.7227 & 0.4781 & 21.8884 & 34.0393 & 0.254s \\
   &  & Bi-Att.~\cite{expl:vit:bidirectionalatt:chen23} & 0.1940 & 0.6489 & 0.4549 & 0.2370 & 0.7344 & 0.4974 & 24.5980 & 31.1754 & 0.254s \\
   &  & TIS~\cite{expl:vit:tis:englebert23} & \cellcolor{rankthird}0.1725 & \cellcolor{ranksecond}\underline{0.6884} & \cellcolor{ranksecond}\underline{0.5159} & \cellcolor{rankthird}0.2098 & \cellcolor{ranksecond}\underline{0.7744} & \cellcolor{ranksecond}\underline{0.5645} & \cellcolor{rankfirst}\textbf{8.0380} & \cellcolor{rankfirst}\textbf{49.9990} & 1.167s \\
   &  & ViT-CX~\cite{expl:vit:vitcx:xie23} & 0.2125 & 0.6249 & 0.4124 & 0.2596 & 0.7030 & 0.4434 & \cellcolor{ranksecond}\underline{9.6828} & \cellcolor{ranksecond}\underline{47.2611} & 0.860s \\
   &  & MDA~\cite{expl:vit:mda:walker25} & 0.1923 & 0.6561 & 0.4638 & 0.2376 & 0.7463 & 0.5087 & 36.2084 & 18.9496 & 13.645s \\
   &  & LeGrad~\cite{expl:vit:legrad:bousselham25} & 0.2053 & 0.6243 & 0.4190 & 0.2492 & 0.7057 & 0.4565 & 16.3933 & 38.1672 & \cellcolor{rankfirst}\textbf{0.006s} \\
   &  & \methodabbrev (Ours) & \cellcolor{ranksecond}\underline{0.1621} & \cellcolor{rankthird}0.6660 & \cellcolor{rankthird}0.5039 & \cellcolor{ranksecond}\underline{0.2005} & \cellcolor{rankthird}0.7531 & \cellcolor{rankthird}0.5526 & 13.0591 & 39.3392 & \cellcolor{rankthird}0.016s \\
   &  & \methodabbrev $T=3$ (Ours) & \cellcolor{rankfirst}\textbf{0.1387} & \cellcolor{rankfirst}\textbf{0.7095} & \cellcolor{rankfirst}\textbf{0.5708} & \cellcolor{rankfirst}\textbf{0.1691} & \cellcolor{rankfirst}\textbf{0.8002} & \cellcolor{rankfirst}\textbf{0.6312} & \cellcolor{rankthird}9.9514 & \cellcolor{rankthird}45.1151 & 0.547s \\
  \cmidrule{2-12}
   & \multirow{10}{*}{\rotatebox[origin=c]{90}{\textbf{Ground Truth Class}}} & Grad-CAM~\cite{expl:act:gradcam:selvaraju17} & 0.3034 & 0.5116 & 0.2082 & 0.3424 & 0.5732 & 0.2307 & 72.2530 & 10.3090 & \cellcolor{ranksecond}\underline{0.014s} \\
   &  & IG~\cite{expl:grad:integratedgradients:sundararajan17} & 0.2114 & 0.5702 & 0.3588 & 0.2463 & 0.6276 & 0.3813 & 16.5402 & 41.8172 & 0.251s \\
   &  & Trans-Att.~\cite{expl:vit:transatt:yuan21} & 0.1876 & 0.5914 & 0.4038 & 0.2212 & 0.6480 & 0.4268 & 19.0520 & 39.1632 & 0.254s \\
   &  & Bi-Att.~\cite{expl:vit:bidirectionalatt:chen23} & 0.1812 & 0.5994 & 0.4182 & 0.2146 & 0.6556 & 0.4410 & 21.6014 & 36.3293 & 0.254s \\
   &  & TIS~\cite{expl:vit:tis:englebert23} & \cellcolor{rankthird}0.1611 & \cellcolor{ranksecond}\underline{0.6419} & \cellcolor{ranksecond}\underline{0.4808} & \cellcolor{rankthird}0.1895 & \cellcolor{ranksecond}\underline{0.7032} & \cellcolor{ranksecond}\underline{0.5137} & \cellcolor{rankfirst}\textbf{6.5602} & \cellcolor{rankfirst}\textbf{55.0809} & 1.167s \\
   &  & ViT-CX~\cite{expl:vit:vitcx:xie23} & 0.2060 & 0.5692 & 0.3632 & 0.2451 & 0.6191 & 0.3739 & 12.9660 & \cellcolor{rankthird}45.2891 & 0.860s \\
   &  & MDA~\cite{expl:vit:mda:walker25} & 0.1808 & 0.6016 & 0.4207 & 0.2169 & 0.6587 & 0.4418 & 33.3235 & 23.6135 & 13.645s \\
   &  & LeGrad~\cite{expl:vit:legrad:bousselham25} & 0.1916 & 0.5792 & 0.3876 & 0.2250 & 0.6353 & 0.4103 & 14.1884 & 43.0671 & \cellcolor{rankfirst}\textbf{0.006s} \\
   &  & \methodabbrev (Ours) & \cellcolor{ranksecond}\underline{0.1507} & \cellcolor{rankthird}0.6131 & \cellcolor{rankthird}0.4624 & \cellcolor{ranksecond}\underline{0.1800} & \cellcolor{rankthird}0.6676 & \cellcolor{rankthird}0.4876 & \cellcolor{rankthird}11.5561 & 43.5511 & \cellcolor{rankthird}0.016s \\
   &  & \methodabbrev $T=3$ (Ours) & \cellcolor{rankfirst}\textbf{0.1290} & \cellcolor{rankfirst}\textbf{0.6537} & \cellcolor{rankfirst}\textbf{0.5247} & \cellcolor{rankfirst}\textbf{0.1512} & \cellcolor{rankfirst}\textbf{0.7131} & \cellcolor{rankfirst}\textbf{0.5620} & \cellcolor{ranksecond}\underline{8.7216} & \cellcolor{ranksecond}\underline{49.1690} & 0.547s \\
  \bottomrule
\end{tabular}
}

%% file: tables/main_results_vitb32_detailed/table.tex
\begin{table*}[t]
    \caption{
        Quantitative comparison of attribution methods on ImageNet using a \mbox{ViT-B/32} classifier~\cite{classif:vit:dosovitskiy21} for all reference image modes~$\mathbf{I}_0$ (black image, mean-colored image, blurred original image). Top-3 results per metric are highlighted (dark to light).  
    }
    \input{tables/main_results_vitb32_detailed/source.tex}
    \label{tab:main_results_vitb32_detailed}
\end{table*}

%% file: tables/main_results_vitb32_detailed/source.tex
\small
\setlength{\tabcolsep}{4pt}
\definecolor{rankfirst}{RGB}{133,171,214}
\definecolor{ranksecond}{RGB}{173,201,230}
\definecolor{rankthird}{RGB}{211,226,243}
\resizebox{\textwidth}{!}{%
\begin{tabular}{cclccccccccr}
  \toprule
   &  & Method & Deletion$\downarrow$ & Insertion$\uparrow$ & Ins. - Del.$\uparrow$ & Positive$\downarrow$ & Negative$\uparrow$ & Neg. - Pos.$\uparrow$ & ADP$\downarrow$ & PIC$\uparrow$ & Runtime \\
  \midrule
  \multirow{20}{*}{\rotatebox[origin=c]{90}{\textbf{$\mathbf{I}_0 \equiv$ Black Image}}} & \multirow{10}{*}{\rotatebox[origin=c]{90}{\textbf{Predicted Class}}} & Grad-CAM~\cite{expl:act:gradcam:selvaraju17} & 0.1865 & 0.4705 & 0.2840 & 0.2365 & 0.5688 & 0.3323 & 70.1550 & 7.6858 & \cellcolor{ranksecond}\underline{0.010s} \\
   &  & IG~\cite{expl:grad:integratedgradients:sundararajan17} & 0.1483 & 0.4963 & 0.3480 & 0.1905 & 0.5940 & 0.4035 & 16.2898 & 37.6572 & 0.157s \\
   &  & Trans-Att.~\cite{expl:vit:transatt:yuan21} & 0.1623 & 0.4923 & 0.3300 & 0.2064 & 0.5902 & 0.3838 & 19.2848 & 33.7333 & 0.157s \\
   &  & Bi-Att.~\cite{expl:vit:bidirectionalatt:chen23} & 0.1497 & 0.5206 & 0.3709 & 0.1914 & 0.6214 & 0.4300 & 19.6633 & 33.5833 & 0.158s \\
   &  & TIS~\cite{expl:vit:tis:englebert23} & \cellcolor{rankthird}0.1368 & \cellcolor{rankfirst}\textbf{0.5555} & \cellcolor{ranksecond}\underline{0.4186} & \cellcolor{rankthird}0.1727 & \cellcolor{rankfirst}\textbf{0.6607} & \cellcolor{ranksecond}\underline{0.4880} & \cellcolor{ranksecond}\underline{11.6713} & \cellcolor{rankfirst}\textbf{46.1231} & 0.416s \\
   &  & ViT-CX~\cite{expl:vit:vitcx:xie23} & 0.1601 & 0.5023 & 0.3422 & 0.2049 & 0.5993 & 0.3944 & \cellcolor{rankthird}12.4364 & \cellcolor{rankthird}44.0451 & 0.763s \\
   &  & MDA~\cite{expl:vit:mda:walker25} & 0.1604 & \cellcolor{ranksecond}\underline{0.5502} & \cellcolor{rankthird}0.3897 & 0.2042 & \cellcolor{ranksecond}\underline{0.6555} & \cellcolor{rankthird}0.4513 & 29.0651 & 23.3835 & 1.803s \\
   &  & LeGrad~\cite{expl:vit:legrad:bousselham25} & 0.1517 & 0.4890 & 0.3373 & 0.1952 & 0.5859 & 0.3907 & 15.4655 & 37.8712 & \cellcolor{rankfirst}\textbf{0.005s} \\
   &  & \methodabbrev (Ours) & \cellcolor{ranksecond}\underline{0.1197} & 0.4947 & 0.3750 & \cellcolor{ranksecond}\underline{0.1548} & 0.5951 & 0.4402 & 13.6634 & 37.6672 & \cellcolor{rankthird}0.016s \\
   &  & \methodabbrev $T=3$ (Ours) & \cellcolor{rankfirst}\textbf{0.1062} & \cellcolor{rankthird}0.5266 & \cellcolor{rankfirst}\textbf{0.4204} & \cellcolor{rankfirst}\textbf{0.1359} & \cellcolor{rankthird}0.6379 & \cellcolor{rankfirst}\textbf{0.5020} & \cellcolor{rankfirst}\textbf{9.6917} & \cellcolor{ranksecond}\underline{45.0871} & 0.212s \\
  \cmidrule{2-12}
   & \multirow{10}{*}{\rotatebox[origin=c]{90}{\textbf{Ground Truth Class}}} & Grad-CAM~\cite{expl:act:gradcam:selvaraju17} & 0.1745 & 0.4397 & 0.2652 & 0.2137 & 0.5163 & 0.3027 & 66.8279 & 11.2838 & \cellcolor{ranksecond}\underline{0.010s} \\
   &  & IG~\cite{expl:grad:integratedgradients:sundararajan17} & 0.1381 & 0.4635 & 0.3254 & 0.1704 & 0.5382 & 0.3679 & 14.6112 & 41.6852 & 0.157s \\
   &  & Trans-Att.~\cite{expl:vit:transatt:yuan21} & 0.1512 & 0.4600 & 0.3089 & 0.1847 & 0.5356 & 0.3508 & 17.2930 & 38.1932 & 0.157s \\
   &  & Bi-Att.~\cite{expl:vit:bidirectionalatt:chen23} & 0.1397 & 0.4851 & 0.3455 & 0.1717 & 0.5605 & 0.3888 & 17.7274 & 37.8892 & 0.158s \\
   &  & TIS~\cite{expl:vit:tis:englebert23} & \cellcolor{rankthird}0.1277 & \cellcolor{rankfirst}\textbf{0.5196} & \cellcolor{ranksecond}\underline{0.3920} & \cellcolor{rankthird}0.1545 & \cellcolor{rankfirst}\textbf{0.6035} & \cellcolor{ranksecond}\underline{0.4490} & \cellcolor{ranksecond}\underline{9.9048} & \cellcolor{rankfirst}\textbf{51.1050} & 0.416s \\
   &  & ViT-CX~\cite{expl:vit:vitcx:xie23} & 0.1545 & 0.4611 & 0.3066 & 0.1920 & 0.5311 & 0.3391 & 15.5775 & \cellcolor{rankthird}42.4012 & 0.763s \\
   &  & MDA~\cite{expl:vit:mda:walker25} & 0.1514 & \cellcolor{ranksecond}\underline{0.5057} & \cellcolor{rankthird}0.3543 & 0.1858 & \cellcolor{ranksecond}\underline{0.5801} & 0.3943 & 27.6928 & 26.7615 & 1.803s \\
   &  & LeGrad~\cite{expl:vit:legrad:bousselham25} & 0.1414 & 0.4571 & 0.3157 & 0.1748 & 0.5316 & 0.3568 & 13.9138 & 41.9092 & \cellcolor{rankfirst}\textbf{0.005s} \\
   &  & \methodabbrev (Ours) & \cellcolor{ranksecond}\underline{0.1109} & 0.4615 & 0.3506 & \cellcolor{ranksecond}\underline{0.1370} & 0.5374 & \cellcolor{rankthird}0.4004 & \cellcolor{rankthird}12.4512 & 41.6292 & \cellcolor{rankthird}0.016s \\
   &  & \methodabbrev $T=3$ (Ours) & \cellcolor{rankfirst}\textbf{0.0986} & \cellcolor{rankthird}0.4910 & \cellcolor{rankfirst}\textbf{0.3925} & \cellcolor{rankfirst}\textbf{0.1200} & \cellcolor{rankthird}0.5774 & \cellcolor{rankfirst}\textbf{0.4573} & \cellcolor{rankfirst}\textbf{8.7874} & \cellcolor{ranksecond}\underline{48.7930} & 0.212s \\
  \midrule
  \multirow{20}{*}{\rotatebox[origin=c]{90}{\textbf{$\mathbf{I}_0 \equiv$ Mean-colored Image}}} & \multirow{10}{*}{\rotatebox[origin=c]{90}{\textbf{Predicted Class}}} & Grad-CAM~\cite{expl:act:gradcam:selvaraju17} & 0.2322 & 0.5177 & 0.2855 & 0.2908 & 0.6226 & 0.3318 & 66.6890 & 7.8998 & \cellcolor{ranksecond}\underline{0.010s} \\
   &  & IG~\cite{expl:grad:integratedgradients:sundararajan17} & 0.1831 & 0.5426 & 0.3594 & 0.2338 & 0.6457 & 0.4120 & 14.7945 & 36.8733 & 0.157s \\
   &  & Trans-Att.~\cite{expl:vit:transatt:yuan21} & 0.1941 & 0.5407 & 0.3466 & 0.2459 & 0.6440 & 0.3981 & 17.7665 & 31.9514 & 0.157s \\
   &  & Bi-Att.~\cite{expl:vit:bidirectionalatt:chen23} & 0.1830 & 0.5625 & 0.3796 & 0.2336 & 0.6690 & 0.4354 & 18.1669 & 31.2494 & 0.158s \\
   &  & TIS~\cite{expl:vit:tis:englebert23} & \cellcolor{rankthird}0.1634 & \cellcolor{rankfirst}\textbf{0.6000} & \cellcolor{ranksecond}\underline{0.4366} & \cellcolor{rankthird}0.2049 & \cellcolor{ranksecond}\underline{0.7106} & \cellcolor{ranksecond}\underline{0.5057} & \cellcolor{rankthird}9.9382 & \cellcolor{rankthird}46.1071 & 0.416s \\
   &  & ViT-CX~\cite{expl:vit:vitcx:xie23} & 0.1920 & 0.5507 & 0.3587 & 0.2443 & 0.6542 & 0.4099 & \cellcolor{rankfirst}\textbf{7.9095} & \cellcolor{rankfirst}\textbf{50.5990} & 0.763s \\
   &  & MDA~\cite{expl:vit:mda:walker25} & 0.1898 & \cellcolor{rankthird}0.5811 & 0.3914 & 0.2408 & \cellcolor{rankthird}0.6927 & 0.4519 & 25.6178 & 24.6295 & 1.803s \\
   &  & LeGrad~\cite{expl:vit:legrad:bousselham25} & 0.1849 & 0.5386 & 0.3537 & 0.2362 & 0.6413 & 0.4051 & 13.8928 & 37.3753 & \cellcolor{rankfirst}\textbf{0.005s} \\
   &  & \methodabbrev (Ours) & \cellcolor{ranksecond}\underline{0.1396} & 0.5533 & \cellcolor{rankthird}0.4137 & \cellcolor{ranksecond}\underline{0.1804} & 0.6587 & \cellcolor{rankthird}0.4783 & 12.1135 & 37.8692 & \cellcolor{rankthird}0.016s \\
   &  & \methodabbrev $T=3$ (Ours) & \cellcolor{rankfirst}\textbf{0.1205} & \cellcolor{ranksecond}\underline{0.5988} & \cellcolor{rankfirst}\textbf{0.4783} & \cellcolor{rankfirst}\textbf{0.1531} & \cellcolor{rankfirst}\textbf{0.7155} & \cellcolor{rankfirst}\textbf{0.5624} & \cellcolor{ranksecond}\underline{7.9805} & \cellcolor{ranksecond}\underline{46.7151} & 0.212s \\
  \cmidrule{2-12}
   & \multirow{10}{*}{\rotatebox[origin=c]{90}{\textbf{Ground Truth Class}}} & Grad-CAM~\cite{expl:act:gradcam:selvaraju17} & 0.2171 & 0.4823 & 0.2652 & 0.2627 & 0.5626 & 0.2999 & 62.7745 & 12.2258 & \cellcolor{ranksecond}\underline{0.010s} \\
   &  & IG~\cite{expl:grad:integratedgradients:sundararajan17} & 0.1705 & 0.5054 & 0.3349 & 0.2099 & 0.5826 & 0.3728 & 13.0733 & 41.3392 & 0.157s \\
   &  & Trans-Att.~\cite{expl:vit:transatt:yuan21} & 0.1808 & 0.5037 & 0.3228 & 0.2208 & 0.5810 & 0.3602 & 15.6204 & 36.8793 & 0.157s \\
   &  & Bi-Att.~\cite{expl:vit:bidirectionalatt:chen23} & 0.1707 & 0.5229 & 0.3522 & 0.2100 & 0.6008 & 0.3908 & 16.0197 & 36.0913 & 0.158s \\
   &  & TIS~\cite{expl:vit:tis:englebert23} & \cellcolor{rankthird}0.1525 & \cellcolor{rankfirst}\textbf{0.5610} & \cellcolor{ranksecond}\underline{0.4085} & \cellcolor{rankthird}0.1843 & \cellcolor{rankfirst}\textbf{0.6482} & \cellcolor{ranksecond}\underline{0.4639} & \cellcolor{ranksecond}\underline{8.1599} & \cellcolor{rankfirst}\textbf{51.2290} & 0.416s \\
   &  & ViT-CX~\cite{expl:vit:vitcx:xie23} & 0.1863 & 0.5028 & 0.3164 & 0.2313 & 0.5746 & 0.3433 & 13.1598 & \cellcolor{rankthird}45.7631 & 0.763s \\
   &  & MDA~\cite{expl:vit:mda:walker25} & 0.1795 & \cellcolor{rankthird}0.5336 & 0.3541 & 0.2201 & \cellcolor{rankthird}0.6109 & 0.3908 & 24.7162 & 27.8754 & 1.803s \\
   &  & LeGrad~\cite{expl:vit:legrad:bousselham25} & 0.1723 & 0.5022 & 0.3299 & 0.2120 & 0.5792 & 0.3672 & 12.2789 & 41.8652 & \cellcolor{rankfirst}\textbf{0.005s} \\
   &  & \methodabbrev (Ours) & \cellcolor{ranksecond}\underline{0.1290} & 0.5155 & \cellcolor{rankthird}0.3865 & \cellcolor{ranksecond}\underline{0.1595} & 0.5933 & \cellcolor{rankthird}0.4337 & \cellcolor{rankthird}10.6786 & 42.2772 & \cellcolor{rankthird}0.016s \\
   &  & \methodabbrev $T=3$ (Ours) & \cellcolor{rankfirst}\textbf{0.1117} & \cellcolor{ranksecond}\underline{0.5579} & \cellcolor{rankfirst}\textbf{0.4462} & \cellcolor{rankfirst}\textbf{0.1354} & \cellcolor{ranksecond}\underline{0.6472} & \cellcolor{rankfirst}\textbf{0.5118} & \cellcolor{rankfirst}\textbf{6.9561} & \cellcolor{ranksecond}\underline{50.9230} & 0.212s \\
  \midrule
  \multirow{20}{*}{\rotatebox[origin=c]{90}{\textbf{$\mathbf{I}_0 \equiv$ Blurred Original Image}}} & \multirow{10}{*}{\rotatebox[origin=c]{90}{\textbf{Predicted Class}}} & Grad-CAM~\cite{expl:act:gradcam:selvaraju17} & 0.2825 & 0.5999 & 0.3174 & 0.3516 & 0.7070 & 0.3554 & 63.9501 & 9.9358 & \cellcolor{ranksecond}\underline{0.010s} \\
   &  & IG~\cite{expl:grad:integratedgradients:sundararajan17} & 0.2378 & 0.6237 & 0.3859 & 0.3036 & 0.7305 & 0.4268 & 11.7650 & 42.5531 & 0.157s \\
   &  & Trans-Att.~\cite{expl:vit:transatt:yuan21} & 0.2508 & 0.6191 & 0.3683 & 0.3178 & 0.7259 & 0.4081 & 14.4145 & 39.2492 & 0.157s \\
   &  & Bi-Att.~\cite{expl:vit:bidirectionalatt:chen23} & 0.2335 & 0.6346 & 0.4012 & 0.2998 & 0.7444 & 0.4446 & 15.1537 & 38.0932 & 0.158s \\
   &  & TIS~\cite{expl:vit:tis:englebert23} & \cellcolor{rankthird}0.2104 & \cellcolor{ranksecond}\underline{0.6727} & \cellcolor{rankthird}0.4623 & \cellcolor{rankthird}0.2656 & \cellcolor{ranksecond}\underline{0.7831} & \cellcolor{rankthird}0.5174 & \cellcolor{ranksecond}\underline{7.5218} & \cellcolor{rankfirst}\textbf{52.9509} & 0.416s \\
   &  & ViT-CX~\cite{expl:vit:vitcx:xie23} & 0.2427 & 0.6282 & 0.3856 & 0.3082 & 0.7332 & 0.4250 & \cellcolor{rankthird}7.6357 & \cellcolor{ranksecond}\underline{52.3410} & 0.763s \\
   &  & MDA~\cite{expl:vit:mda:walker25} & 0.2417 & \cellcolor{rankthird}0.6502 & 0.4086 & 0.3082 & \cellcolor{rankthird}0.7648 & 0.4566 & 21.1588 & 30.3994 & 1.803s \\
   &  & LeGrad~\cite{expl:vit:legrad:bousselham25} & 0.2401 & 0.6230 & 0.3829 & 0.3056 & 0.7292 & 0.4236 & 11.0659 & 43.2291 & \cellcolor{rankfirst}\textbf{0.005s} \\
   &  & \methodabbrev (Ours) & \cellcolor{ranksecond}\underline{0.1820} & 0.6467 & \cellcolor{ranksecond}\underline{0.4647} & \cellcolor{ranksecond}\underline{0.2370} & 0.7571 & \cellcolor{ranksecond}\underline{0.5201} & 9.0191 & 43.6451 & \cellcolor{rankthird}0.016s \\
   &  & \methodabbrev $T=3$ (Ours) & \cellcolor{rankfirst}\textbf{0.1514} & \cellcolor{rankfirst}\textbf{0.6985} & \cellcolor{rankfirst}\textbf{0.5471} & \cellcolor{rankfirst}\textbf{0.1909} & \cellcolor{rankfirst}\textbf{0.8138} & \cellcolor{rankfirst}\textbf{0.6229} & \cellcolor{rankfirst}\textbf{5.8215} & \cellcolor{rankthird}51.7770 & 0.212s \\
  \cmidrule{2-12}
   & \multirow{10}{*}{\rotatebox[origin=c]{90}{\textbf{Ground Truth Class}}} & Grad-CAM~\cite{expl:act:gradcam:selvaraju17} & 0.2640 & 0.5574 & 0.2934 & 0.3179 & 0.6346 & 0.3168 & 60.3271 & 14.0077 & \cellcolor{ranksecond}\underline{0.010s} \\
   &  & IG~\cite{expl:grad:integratedgradients:sundararajan17} & 0.2218 & 0.5783 & 0.3565 & 0.2739 & 0.6520 & 0.3781 & 10.5499 & 46.6731 & 0.157s \\
   &  & Trans-Att.~\cite{expl:vit:transatt:yuan21} & 0.2338 & 0.5742 & 0.3404 & 0.2863 & 0.6482 & 0.3619 & 12.8595 & 43.8211 & 0.157s \\
   &  & Bi-Att.~\cite{expl:vit:bidirectionalatt:chen23} & 0.2184 & 0.5868 & 0.3684 & 0.2714 & 0.6601 & 0.3887 & 13.5802 & 42.6191 & 0.158s \\
   &  & TIS~\cite{expl:vit:tis:englebert23} & \cellcolor{rankthird}0.1965 & \cellcolor{ranksecond}\underline{0.6268} & \cellcolor{ranksecond}\underline{0.4303} & \cellcolor{rankthird}0.2403 & \cellcolor{ranksecond}\underline{0.7090} & \cellcolor{ranksecond}\underline{0.4687} & \cellcolor{ranksecond}\underline{6.2001} & \cellcolor{rankfirst}\textbf{57.4189} & 0.416s \\
   &  & ViT-CX~\cite{expl:vit:vitcx:xie23} & 0.2363 & 0.5694 & 0.3332 & 0.2931 & 0.6361 & 0.3431 & 12.2770 & \cellcolor{rankthird}48.3110 & 0.763s \\
   &  & MDA~\cite{expl:vit:mda:walker25} & 0.2291 & 0.5937 & 0.3646 & 0.2834 & 0.6663 & 0.3829 & 21.1692 & 32.6573 & 1.803s \\
   &  & LeGrad~\cite{expl:vit:legrad:bousselham25} & 0.2240 & 0.5778 & 0.3539 & 0.2754 & 0.6519 & 0.3766 & 9.9322 & 47.3591 & \cellcolor{rankfirst}\textbf{0.005s} \\
   &  & \methodabbrev (Ours) & \cellcolor{ranksecond}\underline{0.1690} & \cellcolor{rankthird}0.5981 & \cellcolor{rankthird}0.4291 & \cellcolor{ranksecond}\underline{0.2117} & \cellcolor{rankthird}0.6693 & \cellcolor{rankthird}0.4576 & \cellcolor{rankthird}8.1351 & 47.5630 & \cellcolor{rankthird}0.016s \\
   &  & \methodabbrev $T=3$ (Ours) & \cellcolor{rankfirst}\textbf{0.1410} & \cellcolor{rankfirst}\textbf{0.6462} & \cellcolor{rankfirst}\textbf{0.5052} & \cellcolor{rankfirst}\textbf{0.1707} & \cellcolor{rankfirst}\textbf{0.7245} & \cellcolor{rankfirst}\textbf{0.5538} & \cellcolor{rankfirst}\textbf{5.2816} & \cellcolor{ranksecond}\underline{55.0229} & 0.212s \\
  \bottomrule
\end{tabular}
}

%% file: tables/main_results_convnext_small_detailed/table.tex
\begin{table*}[t]
    \caption{
        Quantitative comparison of attribution methods on ImageNet using a ConvNeXt Small classifier~\cite{model:convnext:liu22} for all reference image modes~$\mathbf{I}_0$ (black image, mean-colored image, blurred original image). Top-3 results per metric are highlighted (dark to light). 
    }
    \input{tables/main_results_convnext_small_detailed/source.tex}
    \label{tab:main_results_convnext_small_detailed}
\end{table*}

%% file: tables/main_results_convnext_small_detailed/source.tex
\small
\setlength{\tabcolsep}{4pt}
\definecolor{rankfirst}{RGB}{133,171,214}
\definecolor{ranksecond}{RGB}{173,201,230}
\definecolor{rankthird}{RGB}{211,226,243}
\resizebox{\textwidth}{!}{%
\begin{tabular}{cclccccccccr}
  \toprule
   &  & Method & Deletion$\downarrow$ & Insertion$\uparrow$ & Ins. - Del.$\uparrow$ & Positive$\downarrow$ & Negative$\uparrow$ & Neg. - Pos.$\uparrow$ & ADP$\downarrow$ & PIC$\uparrow$ & Runtime \\
  \midrule
  \multirow{14}{*}{\rotatebox[origin=c]{90}{\textbf{$\mathbf{I}_0 \equiv$ Black Image}}} & \multirow{7}{*}{\rotatebox[origin=c]{90}{\textbf{Pred. Class}}} & ScoreCAM~\cite{expl:act:scorecam:wang20} & 0.3405 & 0.5154 & 0.1750 & 0.5082 & 0.7185 & 0.2103 & 56.3717 & \cellcolor{ranksecond}\underline{6.7919} & 0.821s \\
   &  & AblationCAM~\cite{expl:act:ablationcam:desai20} & 0.2930 & 0.5587 & 0.2657 & 0.4337 & 0.7617 & 0.3279 & 51.8343 & 3.8980 & 0.742s \\
   &  & GradCAM++~\cite{expl:act:gradcampp:chattopadhay18} & 0.2832 & 0.5677 & 0.2845 & 0.4419 & \cellcolor{rankthird}0.8028 & 0.3608 & 55.9903 & 3.7580 & \cellcolor{rankthird}0.019s \\
   &  & HiResCAM~\cite{expl:act:hirescam:draelos20} & 0.2374 & \cellcolor{ranksecond}\underline{0.5948} & 0.3574 & \cellcolor{rankthird}0.3498 & \cellcolor{ranksecond}\underline{0.8067} & 0.4570 & 45.7085 & 4.7639 & 0.019s \\
   &  & GradCAM~\cite{expl:act:gradcam:selvaraju17} & \cellcolor{rankthird}0.2374 & \cellcolor{rankfirst}\textbf{0.5948} & \cellcolor{rankthird}0.3574 & 0.3498 & \cellcolor{rankfirst}\textbf{0.8068} & \cellcolor{rankthird}0.4570 & \cellcolor{rankthird}45.7078 & 4.7600 & \cellcolor{rankfirst}\textbf{0.004s} \\
   &  & \methodabbrev (Ours) & \cellcolor{ranksecond}\underline{0.1316} & 0.5567 & \cellcolor{ranksecond}\underline{0.4251} & \cellcolor{ranksecond}\underline{0.1669} & 0.6937 & \cellcolor{ranksecond}\underline{0.5268} & \cellcolor{ranksecond}\underline{24.2854} & \cellcolor{rankthird}6.6999 & \cellcolor{ranksecond}\underline{0.005s} \\
   &  & \methodabbrev $T=3$ (Ours) & \cellcolor{rankfirst}\textbf{0.1220} & \cellcolor{rankthird}0.5684 & \cellcolor{rankfirst}\textbf{0.4464} & \cellcolor{rankfirst}\textbf{0.1536} & 0.7101 & \cellcolor{rankfirst}\textbf{0.5565} & \cellcolor{rankfirst}\textbf{22.6663} & \cellcolor{rankfirst}\textbf{8.0138} & 0.313s \\
  \cmidrule{2-12}
   & \multirow{7}{*}{\rotatebox[origin=c]{90}{\textbf{GT Class}}} & ScoreCAM~\cite{expl:act:scorecam:wang20} & 0.3173 & 0.4748 & 0.1575 & 0.4620 & 0.6426 & 0.1806 & 53.2696 & 11.7578 & 0.821s \\
   &  & AblationCAM~\cite{expl:act:ablationcam:desai20} & 0.2738 & 0.5153 & 0.2415 & 0.3966 & 0.6801 & 0.2834 & 47.3906 & 10.1740 & 0.742s \\
   &  & GradCAM++~\cite{expl:act:gradcampp:chattopadhay18} & 0.2654 & 0.5229 & 0.2575 & 0.4064 & \cellcolor{rankthird}0.7144 & 0.3081 & 50.3194 & 11.0560 & \cellcolor{rankthird}0.019s \\
   &  & HiResCAM~\cite{expl:act:hirescam:draelos20} & 0.2229 & \cellcolor{ranksecond}\underline{0.5479} & 0.3250 & \cellcolor{rankthird}0.3226 & \cellcolor{ranksecond}\underline{0.7192} & 0.3966 & 41.0125 & \cellcolor{rankthird}12.1798 & 0.019s \\
   &  & GradCAM~\cite{expl:act:gradcam:selvaraju17} & \cellcolor{rankthird}0.2229 & \cellcolor{rankfirst}\textbf{0.5479} & \cellcolor{rankthird}0.3250 & 0.3226 & \cellcolor{rankfirst}\textbf{0.7192} & \cellcolor{rankthird}0.3966 & \cellcolor{rankthird}41.0117 & 12.1780 & \cellcolor{rankfirst}\textbf{0.004s} \\
   &  & \methodabbrev (Ours) & \cellcolor{ranksecond}\underline{0.1226} & 0.5173 & \cellcolor{ranksecond}\underline{0.3947} & \cellcolor{ranksecond}\underline{0.1508} & 0.6271 & \cellcolor{ranksecond}\underline{0.4762} & \cellcolor{ranksecond}\underline{21.8415} & \cellcolor{ranksecond}\underline{13.0457} & \cellcolor{ranksecond}\underline{0.005s} \\
   &  & \methodabbrev $T=3$ (Ours) & \cellcolor{rankfirst}\textbf{0.1138} & \cellcolor{rankthird}0.5275 & \cellcolor{rankfirst}\textbf{0.4136} & \cellcolor{rankfirst}\textbf{0.1387} & 0.6418 & \cellcolor{rankfirst}\textbf{0.5031} & \cellcolor{rankfirst}\textbf{20.3470} & \cellcolor{rankfirst}\textbf{14.7177} & 0.313s \\
  \midrule
  \multirow{14}{*}{\rotatebox[origin=c]{90}{\textbf{$\mathbf{I}_0 \equiv$ Mean-colored Image}}} & \multirow{7}{*}{\rotatebox[origin=c]{90}{\textbf{Pred. Class}}} & ScoreCAM~\cite{expl:act:scorecam:wang20} & 0.4095 & 0.5748 & 0.1653 & 0.5629 & 0.7603 & 0.1974 & 55.5316 & \cellcolor{rankfirst}\textbf{6.0919} & 0.821s \\
   &  & AblationCAM~\cite{expl:act:ablationcam:desai20} & 0.3698 & 0.6255 & 0.2557 & 0.5129 & 0.8135 & 0.3005 & 48.9017 & 4.0800 & 0.742s \\
   &  & GradCAM++~\cite{expl:act:gradcampp:chattopadhay18} & 0.3447 & \cellcolor{rankthird}0.6291 & 0.2844 & 0.4917 & \cellcolor{rankthird}0.8454 & 0.3537 & 50.7066 & 4.6420 & \cellcolor{rankthird}0.019s \\
   &  & HiResCAM~\cite{expl:act:hirescam:draelos20} & \cellcolor{rankthird}0.3076 & \cellcolor{ranksecond}\underline{0.6612} & 0.3536 & \cellcolor{rankthird}0.4273 & \cellcolor{rankfirst}\textbf{0.8547} & \cellcolor{rankthird}0.4274 & 41.1248 & 5.8179 & 0.019s \\
   &  & GradCAM~\cite{expl:act:gradcam:selvaraju17} & 0.3076 & \cellcolor{rankfirst}\textbf{0.6612} & \cellcolor{rankthird}0.3536 & 0.4273 & \cellcolor{ranksecond}\underline{0.8547} & 0.4274 & \cellcolor{rankthird}41.1240 & \cellcolor{rankthird}5.8180 & \cellcolor{rankfirst}\textbf{0.004s} \\
   &  & \methodabbrev (Ours) & \cellcolor{ranksecond}\underline{0.1944} & 0.6060 & \cellcolor{ranksecond}\underline{0.4116} & \cellcolor{ranksecond}\underline{0.2728} & 0.7918 & \cellcolor{ranksecond}\underline{0.5191} & \cellcolor{ranksecond}\underline{25.5903} & 4.8899 & \cellcolor{ranksecond}\underline{0.005s} \\
   &  & \methodabbrev $T=3$ (Ours) & \cellcolor{rankfirst}\textbf{0.1739} & 0.6237 & \cellcolor{rankfirst}\textbf{0.4498} & \cellcolor{rankfirst}\textbf{0.2406} & 0.8167 & \cellcolor{rankfirst}\textbf{0.5761} & \cellcolor{rankfirst}\textbf{24.0040} & \cellcolor{ranksecond}\underline{5.9599} & 0.313s \\
  \cmidrule{2-12}
   & \multirow{7}{*}{\rotatebox[origin=c]{90}{\textbf{GT Class}}} & ScoreCAM~\cite{expl:act:scorecam:wang20} & 0.3805 & 0.5287 & 0.1483 & 0.5087 & 0.6779 & 0.1692 & 52.3133 & 11.2478 & 0.821s \\
   &  & AblationCAM~\cite{expl:act:ablationcam:desai20} & 0.3439 & 0.5750 & 0.2310 & 0.4658 & 0.7218 & 0.2560 & 44.4529 & 10.7000 & 0.742s \\
   &  & GradCAM++~\cite{expl:act:gradcampp:chattopadhay18} & 0.3221 & \cellcolor{rankthird}0.5780 & 0.2559 & 0.4502 & \cellcolor{rankthird}0.7499 & 0.2997 & 45.4385 & 12.1920 & \cellcolor{rankthird}0.019s \\
   &  & HiResCAM~\cite{expl:act:hirescam:draelos20} & \cellcolor{rankthird}0.2880 & \cellcolor{ranksecond}\underline{0.6073} & 0.3193 & \cellcolor{rankthird}0.3922 & \cellcolor{ranksecond}\underline{0.7592} & 0.3670 & 36.7743 & \cellcolor{ranksecond}\underline{13.6277} & 0.019s \\
   &  & GradCAM~\cite{expl:act:gradcam:selvaraju17} & 0.2880 & \cellcolor{rankfirst}\textbf{0.6073} & \cellcolor{rankthird}0.3193 & 0.3922 & \cellcolor{rankfirst}\textbf{0.7592} & \cellcolor{rankthird}0.3670 & \cellcolor{rankthird}36.7733 & \cellcolor{rankfirst}\textbf{13.6280} & \cellcolor{rankfirst}\textbf{0.004s} \\
   &  & \methodabbrev (Ours) & \cellcolor{ranksecond}\underline{0.1810} & 0.5591 & \cellcolor{ranksecond}\underline{0.3781} & \cellcolor{ranksecond}\underline{0.2477} & 0.7077 & \cellcolor{ranksecond}\underline{0.4600} & \cellcolor{ranksecond}\underline{22.9150} & 11.7798 & \cellcolor{ranksecond}\underline{0.005s} \\
   &  & \methodabbrev $T=3$ (Ours) & \cellcolor{rankfirst}\textbf{0.1622} & 0.5743 & \cellcolor{rankfirst}\textbf{0.4121} & \cellcolor{rankfirst}\textbf{0.2188} & 0.7293 & \cellcolor{rankfirst}\textbf{0.5105} & \cellcolor{rankfirst}\textbf{21.4147} & \cellcolor{rankthird}13.1257 & 0.313s \\
  \midrule
  \multirow{14}{*}{\rotatebox[origin=c]{90}{\textbf{$\mathbf{I}_0 \equiv$ Blurred Original Image}}} & \multirow{7}{*}{\rotatebox[origin=c]{90}{\textbf{Pred. Class}}} & ScoreCAM~\cite{expl:act:scorecam:wang20} & 0.3836 & 0.5372 & 0.1536 & 0.5333 & 0.7314 & 0.1981 & 56.1903 & 4.8119 & 0.821s \\
   &  & AblationCAM~\cite{expl:act:ablationcam:desai20} & 0.3393 & 0.5830 & 0.2437 & 0.4792 & 0.7878 & 0.3087 & 46.1059 & 4.3200 & 0.742s \\
   &  & GradCAM++~\cite{expl:act:gradcampp:chattopadhay18} & 0.3199 & 0.6054 & 0.2855 & 0.4568 & 0.8212 & 0.3644 & 46.1080 & 4.7460 & \cellcolor{rankthird}0.019s \\
   &  & HiResCAM~\cite{expl:act:hirescam:draelos20} & 0.2793 & \cellcolor{rankthird}0.6242 & 0.3449 & 0.3954 & \cellcolor{rankfirst}\textbf{0.8327} & 0.4373 & 37.7086 & \cellcolor{rankthird}5.7599 & 0.019s \\
   &  & GradCAM~\cite{expl:act:gradcam:selvaraju17} & \cellcolor{rankthird}0.2793 & \cellcolor{ranksecond}\underline{0.6242} & \cellcolor{rankthird}0.3449 & \cellcolor{rankthird}0.3954 & \cellcolor{ranksecond}\underline{0.8327} & \cellcolor{rankthird}0.4374 & \cellcolor{rankthird}37.7080 & 5.7580 & \cellcolor{rankfirst}\textbf{0.004s} \\
   &  & \methodabbrev (Ours) & \cellcolor{ranksecond}\underline{0.1963} & 0.6105 & \cellcolor{ranksecond}\underline{0.4142} & \cellcolor{ranksecond}\underline{0.2828} & 0.7993 & \cellcolor{ranksecond}\underline{0.5164} & \cellcolor{ranksecond}\underline{21.1815} & \cellcolor{ranksecond}\underline{6.4679} & \cellcolor{ranksecond}\underline{0.005s} \\
   &  & \methodabbrev $T=3$ (Ours) & \cellcolor{rankfirst}\textbf{0.1767} & \cellcolor{rankfirst}\textbf{0.6315} & \cellcolor{rankfirst}\textbf{0.4547} & \cellcolor{rankfirst}\textbf{0.2507} & \cellcolor{rankthird}0.8257 & \cellcolor{rankfirst}\textbf{0.5749} & \cellcolor{rankfirst}\textbf{19.7132} & \cellcolor{rankfirst}\textbf{7.5538} & 0.313s \\
  \cmidrule{2-12}
   & \multirow{7}{*}{\rotatebox[origin=c]{90}{\textbf{GT Class}}} & ScoreCAM~\cite{expl:act:scorecam:wang20} & 0.3582 & 0.4968 & 0.1386 & 0.4860 & 0.6580 & 0.1719 & 52.6074 & 10.1738 & 0.821s \\
   &  & AblationCAM~\cite{expl:act:ablationcam:desai20} & 0.3173 & 0.5386 & 0.2213 & 0.4383 & 0.7052 & 0.2669 & 41.7112 & 11.0820 & 0.742s \\
   &  & GradCAM++~\cite{expl:act:gradcampp:chattopadhay18} & 0.3003 & 0.5589 & 0.2585 & 0.4208 & 0.7351 & 0.3143 & 40.9487 & 12.4740 & \cellcolor{rankthird}0.019s \\
   &  & HiResCAM~\cite{expl:act:hirescam:draelos20} & 0.2627 & \cellcolor{rankthird}0.5761 & 0.3134 & 0.3651 & \cellcolor{ranksecond}\underline{0.7460} & 0.3809 & 33.4107 & \cellcolor{ranksecond}\underline{13.6357} & 0.019s \\
   &  & GradCAM~\cite{expl:act:gradcam:selvaraju17} & \cellcolor{rankthird}0.2627 & \cellcolor{ranksecond}\underline{0.5761} & \cellcolor{rankthird}0.3134 & \cellcolor{rankthird}0.3651 & \cellcolor{rankfirst}\textbf{0.7460} & \cellcolor{rankthird}0.3809 & \cellcolor{rankthird}33.4100 & \cellcolor{rankthird}13.6340 & \cellcolor{rankfirst}\textbf{0.004s} \\
   &  & \methodabbrev (Ours) & \cellcolor{ranksecond}\underline{0.1836} & 0.5636 & \cellcolor{ranksecond}\underline{0.3801} & \cellcolor{ranksecond}\underline{0.2587} & 0.7136 & \cellcolor{ranksecond}\underline{0.4549} & \cellcolor{ranksecond}\underline{18.9536} & 13.1017 & \cellcolor{ranksecond}\underline{0.005s} \\
   &  & \methodabbrev $T=3$ (Ours) & \cellcolor{rankfirst}\textbf{0.1655} & \cellcolor{rankfirst}\textbf{0.5821} & \cellcolor{rankfirst}\textbf{0.4167} & \cellcolor{rankfirst}\textbf{0.2295} & \cellcolor{rankthird}0.7372 & \cellcolor{rankfirst}\textbf{0.5078} & \cellcolor{rankfirst}\textbf{17.6002} & \cellcolor{rankfirst}\textbf{14.3477} & 0.313s \\
  \bottomrule
\end{tabular}
}

%% file: figures/AUC_plots/fig.tex
\begin{figure}[tb]
    \centering 
        \input{figures/AUC_plots/source.tex}
        \caption{Deletion and Insertion AUC curves averaged over all reference images ($\mathbf{I}_0 \in {\text{black, mean, blur}}$) on ImageNet validation samples.
        Each row corresponds to a classifier--class combination (ViT-B/16 and ViT-B/32, predicted and ground-truth class).
        For Deletion (\emph{left}), a steeper drop indicates better attribution; for Insertion (\emph{right}), a steeper rise indicates better attribution.}
    \label{fig:auc_plots}
\end{figure}
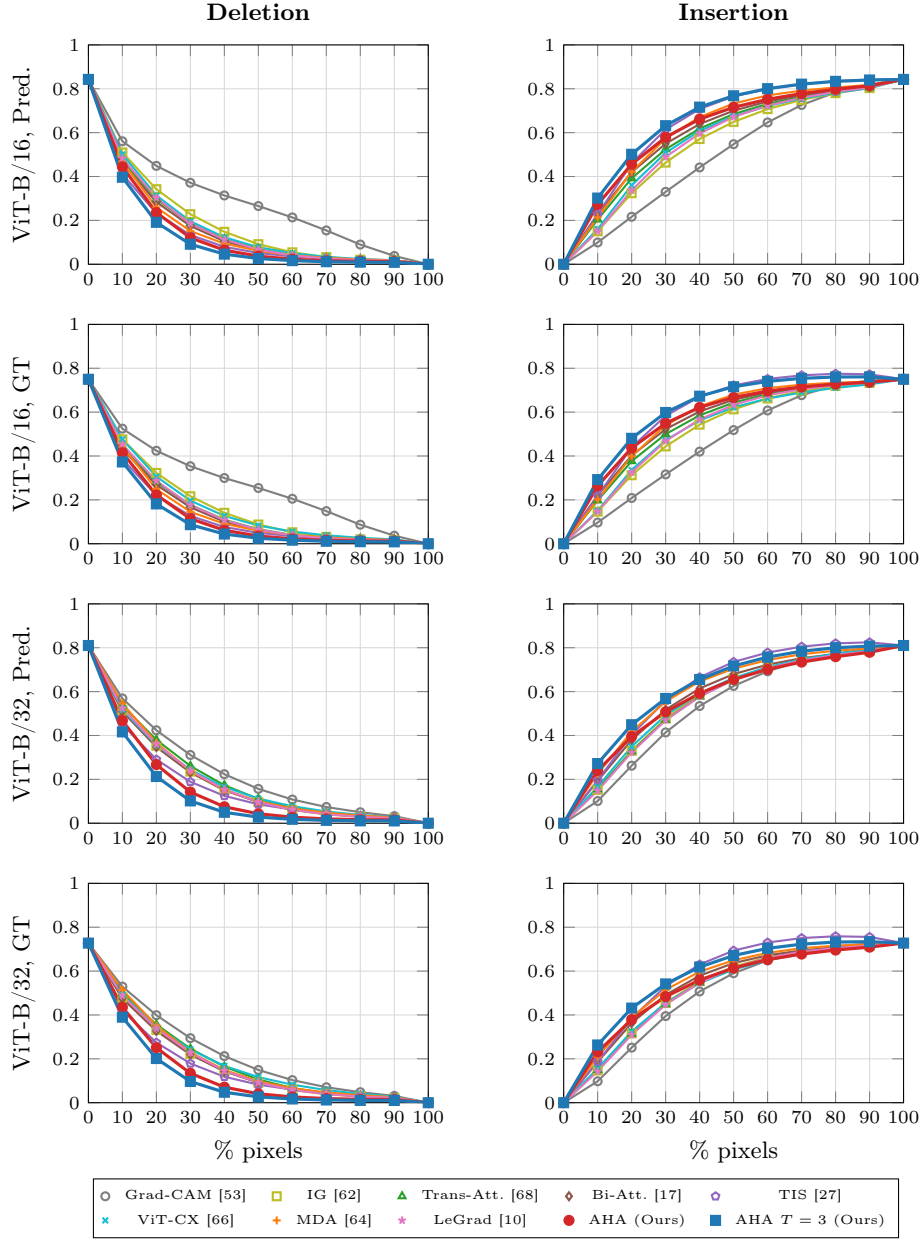

%% file: figures/AUC_plots/source.tex
\definecolor{plotclr0}{RGB}{127,127,127}
\definecolor{plotclr1}{RGB}{188,189,34}
\definecolor{plotclr2}{RGB}{44,160,44}
\definecolor{plotclr3}{RGB}{140,86,75}
\definecolor{plotclr4}{RGB}{148,103,189}
\definecolor{plotclr5}{RGB}{23,190,207}
\definecolor{plotclr6}{RGB}{255,127,14}
\definecolor{plotclr7}{RGB}{227,119,194}
\definecolor{plotclr8}{RGB}{214,39,40}
\definecolor{plotclr9}{RGB}{31,119,180}

\pgfplotsset{method0/.style={plotclr0, thick, mark=o, mark size=1.5pt}}
\pgfplotsset{method1/.style={plotclr1, thick, mark=square, mark size=1.5pt}}
\pgfplotsset{method2/.style={plotclr2, thick, mark=triangle, mark size=1.5pt}}
\pgfplotsset{method3/.style={plotclr3, thick, mark=diamond, mark size=1.5pt}}
\pgfplotsset{method4/.style={plotclr4, thick, mark=pentagon, mark size=1.5pt}}
\pgfplotsset{method5/.style={plotclr5, thick, mark=x, mark size=1.5pt}}
\pgfplotsset{method6/.style={plotclr6, thick, mark=+, mark size=1.5pt}}
\pgfplotsset{method7/.style={plotclr7, thick, mark=star, mark size=1.5pt}}
\pgfplotsset{method8/.style={plotclr8, very thick, mark=*, mark size=1.5pt}}
\pgfplotsset{method9/.style={plotclr9, very thick, mark=square*, mark size=1.5pt}}

\begin{tikzpicture}
\begin{groupplot}[
    group style={
        group size=2 by 4,
        horizontal sep=1.8cm,
        vertical sep=0.8cm,
        xlabels at=edge bottom,
        ylabels at=edge left,
    },
    width=0.5\textwidth,
    height=4.5cm,
    xmin=0, xmax=100,
    ymin=0, ymax=1,
    xtick={0,10,20,30,40,50,60,70,80,90,100},
    grid=major,
    grid style={gray!30},
    tick label style={font=\scriptsize},
    label style={font=\footnotesize},
    title style={font=\footnotesize\bfseries},
]

  \nextgroupplot[title={Deletion}, ylabel={ViT-B/16, Pred.}]
  \addplot[method0] coordinates {(0,0.8427) (10,0.5607) (20,0.4485) (30,0.3716) (40,0.3137) (50,0.2654) (60,0.2136) (70,0.1543) (80,0.0898) (90,0.0380) (100,0.0000)};
  \addplot[method1] coordinates {(0,0.8427) (10,0.5094) (20,0.3423) (30,0.2283) (40,0.1480) (50,0.0917) (60,0.0545) (70,0.0326) (80,0.0210) (90,0.0166) (100,0.0000)};
  \addplot[method2] coordinates {(0,0.8427) (10,0.4716) (20,0.2971) (30,0.1864) (40,0.1147) (50,0.0689) (60,0.0416) (70,0.0260) (80,0.0187) (90,0.0162) (100,0.0000)};
  \addplot[method3] coordinates {(0,0.8427) (10,0.4597) (20,0.2834) (30,0.1745) (40,0.1053) (50,0.0640) (60,0.0392) (70,0.0257) (80,0.0190) (90,0.0166) (100,0.0000)};
  \addplot[method4] coordinates {(0,0.8427) (10,0.4109) (20,0.2328) (30,0.1332) (40,0.0797) (50,0.0512) (60,0.0355) (70,0.0263) (80,0.0206) (90,0.0173) (100,0.0000)};
  \addplot[method5] coordinates {(0,0.8427) (10,0.5010) (20,0.3136) (30,0.1962) (40,0.1231) (50,0.0779) (60,0.0504) (70,0.0339) (80,0.0239) (90,0.0182) (100,0.0000)};
  \addplot[method6] coordinates {(0,0.8427) (10,0.4677) (20,0.2595) (30,0.1514) (40,0.0942) (50,0.0610) (60,0.0411) (70,0.0293) (80,0.0218) (90,0.0179) (100,0.0000)};
  \addplot[method7] coordinates {(0,0.8427) (10,0.4835) (20,0.3028) (30,0.1859) (40,0.1126) (50,0.0669) (60,0.0405) (70,0.0259) (80,0.0184) (90,0.0159) (100,0.0000)};
  \addplot[method8] coordinates {(0,0.8427) (10,0.4440) (20,0.2357) (30,0.1206) (40,0.0643) (50,0.0370) (60,0.0230) (70,0.0160) (80,0.0131) (90,0.0119) (100,0.0000)};
  \addplot[method9] coordinates {(0,0.8427) (10,0.3966) (20,0.1912) (30,0.0911) (40,0.0461) (50,0.0261) (60,0.0165) (70,0.0116) (80,0.0095) (90,0.0087) (100,0.0000)};

  \nextgroupplot[title={Insertion}]
  \addplot[method0] coordinates {(0,0.0000) (10,0.0993) (20,0.2164) (30,0.3302) (40,0.4414) (50,0.5474) (60,0.6466) (70,0.7262) (80,0.7834) (90,0.8192) (100,0.8427)};
  \addplot[method1] coordinates {(0,0.0000) (10,0.1503) (20,0.3238) (30,0.4635) (40,0.5709) (50,0.6490) (60,0.7074) (70,0.7494) (80,0.7801) (90,0.8032) (100,0.8427)};
  \addplot[method2] coordinates {(0,0.0000) (10,0.2040) (20,0.3919) (30,0.5244) (40,0.6187) (50,0.6848) (60,0.7311) (70,0.7640) (80,0.7877) (90,0.8068) (100,0.8427)};
  \addplot[method3] coordinates {(0,0.0000) (10,0.2281) (20,0.4198) (30,0.5498) (40,0.6392) (50,0.7002) (60,0.7419) (70,0.7712) (80,0.7922) (90,0.8086) (100,0.8427)};
  \addplot[method4] coordinates {(0,0.0000) (10,0.2275) (20,0.4611) (30,0.6142) (40,0.7079) (50,0.7653) (60,0.7996) (70,0.8211) (80,0.8348) (90,0.8420) (100,0.8427)};
  \addplot[method5] coordinates {(0,0.0000) (10,0.1606) (20,0.3600) (30,0.5083) (40,0.6082) (50,0.6749) (60,0.7216) (70,0.7551) (80,0.7814) (90,0.8031) (100,0.8427)};
  \addplot[method6] coordinates {(0,0.0000) (10,0.2132) (20,0.4216) (30,0.5735) (40,0.6711) (50,0.7328) (60,0.7698) (70,0.7918) (80,0.8061) (90,0.8176) (100,0.8427)};
  \addplot[method7] coordinates {(0,0.0000) (10,0.1507) (20,0.3372) (30,0.4893) (40,0.5965) (50,0.6715) (60,0.7231) (70,0.7593) (80,0.7857) (90,0.8062) (100,0.8427)};
  \addplot[method8] coordinates {(0,0.0000) (10,0.2729) (20,0.4562) (30,0.5790) (40,0.6617) (50,0.7156) (60,0.7522) (70,0.7781) (80,0.7977) (90,0.8139) (100,0.8427)};
  \addplot[method9] coordinates {(0,0.0000) (10,0.3018) (20,0.5014) (30,0.6318) (40,0.7164) (50,0.7681) (60,0.8006) (70,0.8211) (80,0.8338) (90,0.8400) (100,0.8427)};

  \nextgroupplot[ylabel={ViT-B/16, GT}]
  \addplot[method0] coordinates {(0,0.7498) (10,0.5247) (20,0.4239) (30,0.3534) (40,0.2996) (50,0.2543) (60,0.2051) (70,0.1487) (80,0.0868) (90,0.0367) (100,0.0000)};
  \addplot[method1] coordinates {(0,0.7498) (10,0.4760) (20,0.3229) (30,0.2166) (40,0.1411) (50,0.0877) (60,0.0523) (70,0.0314) (80,0.0202) (90,0.0161) (100,0.0000)};
  \addplot[method2] coordinates {(0,0.7498) (10,0.4417) (20,0.2810) (30,0.1774) (40,0.1093) (50,0.0659) (60,0.0398) (70,0.0251) (80,0.0180) (90,0.0158) (100,0.0000)};
  \addplot[method3] coordinates {(0,0.7498) (10,0.4317) (20,0.2685) (30,0.1661) (40,0.1005) (50,0.0612) (60,0.0375) (70,0.0248) (80,0.0185) (90,0.0162) (100,0.0000)};
  \addplot[method4] coordinates {(0,0.7498) (10,0.3862) (20,0.2208) (30,0.1269) (40,0.0763) (50,0.0495) (60,0.0347) (70,0.0260) (80,0.0204) (90,0.0171) (100,0.0000)};
  \addplot[method5] coordinates {(0,0.7498) (10,0.4759) (20,0.3062) (30,0.1965) (40,0.1268) (50,0.0829) (60,0.0555) (70,0.0384) (80,0.0269) (90,0.0198) (100,0.0000)};
  \addplot[method6] coordinates {(0,0.7498) (10,0.4430) (20,0.2489) (30,0.1455) (40,0.0909) (50,0.0591) (60,0.0398) (70,0.0284) (80,0.0212) (90,0.0174) (100,0.0000)};
  \addplot[method7] coordinates {(0,0.7498) (10,0.4527) (20,0.2866) (30,0.1767) (40,0.1076) (50,0.0642) (60,0.0389) (70,0.0248) (80,0.0177) (90,0.0155) (100,0.0000)};
  \addplot[method8] coordinates {(0,0.7498) (10,0.4150) (20,0.2226) (30,0.1143) (40,0.0612) (50,0.0353) (60,0.0222) (70,0.0156) (80,0.0128) (90,0.0117) (100,0.0000)};
  \addplot[method9] coordinates {(0,0.7498) (10,0.3728) (20,0.1815) (30,0.0869) (40,0.0442) (50,0.0251) (60,0.0161) (70,0.0113) (80,0.0093) (90,0.0086) (100,0.0000)};

  \nextgroupplot[]
  \addplot[method0] coordinates {(0,0.0000) (10,0.0967) (20,0.2085) (30,0.3161) (40,0.4200) (50,0.5182) (60,0.6076) (70,0.6773) (80,0.7226) (90,0.7443) (100,0.7498)};
  \addplot[method1] coordinates {(0,0.0000) (10,0.1466) (20,0.3123) (30,0.4437) (40,0.5429) (50,0.6127) (60,0.6623) (70,0.6970) (80,0.7192) (90,0.7330) (100,0.7498)};
  \addplot[method2] coordinates {(0,0.0000) (10,0.1974) (20,0.3764) (30,0.4997) (40,0.5852) (50,0.6436) (60,0.6822) (70,0.7080) (80,0.7243) (90,0.7347) (100,0.7498)};
  \addplot[method3] coordinates {(0,0.0000) (10,0.2205) (20,0.4023) (30,0.5222) (40,0.6032) (50,0.6562) (60,0.6903) (70,0.7127) (80,0.7265) (90,0.7355) (100,0.7498)};
  \addplot[method4] coordinates {(0,0.0000) (10,0.2183) (20,0.4397) (30,0.5834) (40,0.6702) (50,0.7219) (60,0.7513) (70,0.7677) (80,0.7750) (90,0.7726) (100,0.7498)};
  \addplot[method5] coordinates {(0,0.0000) (10,0.1507) (20,0.3363) (30,0.4722) (40,0.5626) (50,0.6218) (60,0.6619) (70,0.6893) (80,0.7106) (90,0.7264) (100,0.7498)};
  \addplot[method6] coordinates {(0,0.0000) (10,0.2018) (20,0.3989) (30,0.5396) (40,0.6265) (50,0.6797) (60,0.7093) (70,0.7252) (80,0.7340) (90,0.7394) (100,0.7498)};
  \addplot[method7] coordinates {(0,0.0000) (10,0.1471) (20,0.3251) (30,0.4682) (40,0.5668) (50,0.6332) (60,0.6765) (70,0.7047) (80,0.7227) (90,0.7346) (100,0.7498)};
  \addplot[method8] coordinates {(0,0.0000) (10,0.2640) (20,0.4367) (30,0.5490) (40,0.6218) (50,0.6671) (60,0.6960) (70,0.7148) (80,0.7277) (90,0.7368) (100,0.7498)};
  \addplot[method9] coordinates {(0,0.0000) (10,0.2924) (20,0.4806) (30,0.5983) (40,0.6726) (50,0.7156) (60,0.7401) (70,0.7540) (80,0.7603) (90,0.7607) (100,0.7498)};

  \nextgroupplot[ylabel={ViT-B/32, Pred.}]
  \addplot[method0] coordinates {(0,0.8102) (10,0.5690) (20,0.4238) (30,0.3111) (40,0.2236) (50,0.1566) (60,0.1081) (70,0.0739) (80,0.0500) (90,0.0323) (100,0.0000)};
  \addplot[method1] coordinates {(0,0.8102) (10,0.5152) (20,0.3531) (30,0.2354) (40,0.1515) (50,0.0956) (60,0.0606) (70,0.0403) (80,0.0291) (90,0.0229) (100,0.0000)};
  \addplot[method2] coordinates {(0,0.8102) (10,0.5366) (20,0.3802) (30,0.2609) (40,0.1731) (50,0.1110) (60,0.0699) (70,0.0450) (80,0.0306) (90,0.0231) (100,0.0000)};
  \addplot[method3] coordinates {(0,0.8102) (10,0.5093) (20,0.3473) (30,0.2315) (40,0.1501) (50,0.0969) (60,0.0624) (70,0.0425) (80,0.0304) (90,0.0233) (100,0.0000)};
  \addplot[method4] coordinates {(0,0.8102) (10,0.4498) (20,0.2907) (30,0.1889) (40,0.1247) (50,0.0862) (60,0.0622) (70,0.0464) (80,0.0351) (90,0.0264) (100,0.0000)};
  \addplot[method5] coordinates {(0,0.8102) (10,0.5174) (20,0.3564) (30,0.2438) (40,0.1653) (50,0.1124) (60,0.0774) (70,0.0535) (80,0.0377) (90,0.0268) (100,0.0000)};
  \addplot[method6] coordinates {(0,0.8102) (10,0.5459) (20,0.3678) (30,0.2378) (40,0.1532) (50,0.1000) (60,0.0675) (70,0.0478) (80,0.0349) (90,0.0260) (100,0.0000)};
  \addplot[method7] coordinates {(0,0.8102) (10,0.5244) (20,0.3609) (30,0.2404) (40,0.1534) (50,0.0965) (60,0.0609) (70,0.0404) (80,0.0291) (90,0.0227) (100,0.0000)};
  \addplot[method8] coordinates {(0,0.8102) (10,0.4678) (20,0.2672) (30,0.1423) (40,0.0751) (50,0.0431) (60,0.0275) (70,0.0197) (80,0.0159) (90,0.0147) (100,0.0000)};
  \addplot[method9] coordinates {(0,0.8102) (10,0.4162) (20,0.2130) (30,0.1017) (40,0.0498) (50,0.0279) (60,0.0178) (70,0.0130) (80,0.0108) (90,0.0105) (100,0.0000)};

  \nextgroupplot[]
  \addplot[method0] coordinates {(0,0.0000) (10,0.1012) (20,0.2622) (30,0.4139) (40,0.5343) (50,0.6258) (60,0.6926) (70,0.7404) (80,0.7736) (90,0.7953) (100,0.8102)};
  \addplot[method1] coordinates {(0,0.0000) (10,0.1521) (20,0.3301) (30,0.4777) (40,0.5855) (50,0.6598) (60,0.7105) (70,0.7442) (80,0.7680) (90,0.7849) (100,0.8102)};
  \addplot[method2] coordinates {(0,0.0000) (10,0.1545) (20,0.3277) (30,0.4708) (40,0.5775) (50,0.6537) (60,0.7058) (70,0.7414) (80,0.7652) (90,0.7823) (100,0.8102)};
  \addplot[method3] coordinates {(0,0.0000) (10,0.1931) (20,0.3798) (30,0.5176) (40,0.6143) (50,0.6798) (60,0.7230) (70,0.7521) (80,0.7717) (90,0.7861) (100,0.8102)};
  \addplot[method4] coordinates {(0,0.0000) (10,0.1957) (20,0.4033) (30,0.5594) (40,0.6654) (50,0.7354) (60,0.7781) (70,0.8048) (80,0.8200) (90,0.8246) (100,0.8102)};
  \addplot[method5] coordinates {(0,0.0000) (10,0.1662) (20,0.3495) (30,0.4920) (40,0.5923) (50,0.6625) (60,0.7116) (70,0.7469) (80,0.7724) (90,0.7890) (100,0.8102)};
  \addplot[method6] coordinates {(0,0.0000) (10,0.2251) (20,0.4175) (30,0.5546) (40,0.6460) (50,0.7047) (60,0.7440) (70,0.7704) (80,0.7866) (90,0.7972) (100,0.8102)};
  \addplot[method7] coordinates {(0,0.0000) (10,0.1483) (20,0.3227) (30,0.4695) (40,0.5777) (50,0.6537) (60,0.7063) (70,0.7422) (80,0.7666) (90,0.7842) (100,0.8102)};
  \addplot[method8] coordinates {(0,0.0000) (10,0.2398) (20,0.3949) (30,0.5076) (40,0.5920) (50,0.6553) (60,0.7015) (70,0.7346) (80,0.7592) (90,0.7793) (100,0.8102)};
  \addplot[method9] coordinates {(0,0.0000) (10,0.2720) (20,0.4491) (30,0.5687) (40,0.6552) (50,0.7170) (60,0.7573) (70,0.7840) (80,0.8001) (90,0.8073) (100,0.8102)};

  \nextgroupplot[ylabel={ViT-B/32, GT}, xlabel={\% pixels}]
  \addplot[method0] coordinates {(0,0.7276) (10,0.5294) (20,0.3989) (30,0.2943) (40,0.2124) (50,0.1493) (60,0.1032) (70,0.0705) (80,0.0479) (90,0.0311) (100,0.0000)};
  \addplot[method1] coordinates {(0,0.7276) (10,0.4793) (20,0.3316) (30,0.2226) (40,0.1439) (50,0.0910) (60,0.0579) (70,0.0386) (80,0.0281) (90,0.0222) (100,0.0000)};
  \addplot[method2] coordinates {(0,0.7276) (10,0.4986) (20,0.3566) (30,0.2465) (40,0.1644) (50,0.1058) (60,0.0668) (70,0.0431) (80,0.0294) (90,0.0224) (100,0.0000)};
  \addplot[method3] coordinates {(0,0.7276) (10,0.4752) (20,0.3271) (30,0.2194) (40,0.1430) (50,0.0925) (60,0.0598) (70,0.0407) (80,0.0293) (90,0.0226) (100,0.0000)};
  \addplot[method4] coordinates {(0,0.7276) (10,0.4204) (20,0.2742) (30,0.1789) (40,0.1184) (50,0.0823) (60,0.0597) (70,0.0446) (80,0.0338) (90,0.0256) (100,0.0000)};
  \addplot[method5] coordinates {(0,0.7276) (10,0.4916) (20,0.3471) (30,0.2428) (40,0.1679) (50,0.1166) (60,0.0822) (70,0.0575) (80,0.0403) (90,0.0276) (100,0.0000)};
  \addplot[method6] coordinates {(0,0.7276) (10,0.5138) (20,0.3522) (30,0.2303) (40,0.1493) (50,0.0978) (60,0.0661) (70,0.0466) (80,0.0340) (90,0.0252) (100,0.0000)};
  \addplot[method7] coordinates {(0,0.7276) (10,0.4883) (20,0.3391) (30,0.2275) (40,0.1457) (50,0.0920) (60,0.0582) (70,0.0388) (80,0.0281) (90,0.0219) (100,0.0000)};
  \addplot[method8] coordinates {(0,0.7276) (10,0.4349) (20,0.2506) (30,0.1342) (40,0.0710) (50,0.0407) (60,0.0262) (70,0.0189) (80,0.0153) (90,0.0143) (100,0.0000)};
  \addplot[method9] coordinates {(0,0.7276) (10,0.3898) (20,0.2014) (30,0.0966) (40,0.0476) (50,0.0266) (60,0.0170) (70,0.0125) (80,0.0104) (90,0.0102) (100,0.0000)};

  \nextgroupplot[xlabel={\% pixels}, legend to name=grouplegend, legend columns=5, legend style={font=\tiny, column sep=1ex, /tikz/every even column/.append style={column sep=2ex}}, legend image post style={only marks}]
  \addplot[method0] coordinates {(0,0.0000) (10,0.0977) (20,0.2511) (30,0.3948) (40,0.5071) (50,0.5902) (60,0.6486) (70,0.6880) (80,0.7131) (90,0.7256) (100,0.7276)};
  \addlegendentry{Grad-CAM~\cite{expl:act:gradcam:selvaraju17}}
  \addplot[method1] coordinates {(0,0.0000) (10,0.1472) (20,0.3164) (30,0.4554) (40,0.5542) (50,0.6203) (60,0.6630) (70,0.6893) (80,0.7059) (90,0.7155) (100,0.7276)};
  \addlegendentry{IG~\cite{expl:grad:integratedgradients:sundararajan17}}
  \addplot[method2] coordinates {(0,0.0000) (10,0.1492) (20,0.3141) (30,0.4485) (40,0.5466) (50,0.6147) (60,0.6591) (70,0.6873) (80,0.7041) (90,0.7137) (100,0.7276)};
  \addlegendentry{Trans-Att.~\cite{expl:vit:transatt:yuan21}}
  \addplot[method3] coordinates {(0,0.0000) (10,0.1862) (20,0.3630) (30,0.4918) (40,0.5789) (50,0.6364) (60,0.6722) (70,0.6941) (80,0.7073) (90,0.7154) (100,0.7276)};
  \addlegendentry{Bi-Att.~\cite{expl:vit:bidirectionalatt:chen23}}
  \addplot[method4] coordinates {(0,0.0000) (10,0.1871) (20,0.3843) (30,0.5312) (40,0.6301) (50,0.6935) (60,0.7301) (70,0.7505) (80,0.7588) (90,0.7556) (100,0.7276)};
  \addlegendentry{TIS~\cite{expl:vit:tis:englebert23}}
  \addplot[method5] coordinates {(0,0.0000) (10,0.1556) (20,0.3246) (30,0.4552) (40,0.5452) (50,0.6071) (60,0.6490) (70,0.6780) (80,0.6986) (90,0.7119) (100,0.7276)};
  \addlegendentry{ViT-CX~\cite{expl:vit:vitcx:xie23}}
  \addplot[method6] coordinates {(0,0.0000) (10,0.2094) (20,0.3884) (30,0.5146) (40,0.5974) (50,0.6492) (60,0.6831) (70,0.7042) (80,0.7157) (90,0.7219) (100,0.7276)};
  \addlegendentry{MDA~\cite{expl:vit:mda:walker25}}
  \addplot[method7] coordinates {(0,0.0000) (10,0.1431) (20,0.3096) (30,0.4476) (40,0.5472) (50,0.6151) (60,0.6600) (70,0.6884) (80,0.7053) (90,0.7152) (100,0.7276)};
  \addlegendentry{LeGrad~\cite{expl:vit:legrad:bousselham25}}
  \addplot[method8] coordinates {(0,0.0000) (10,0.2315) (20,0.3788) (30,0.4834) (40,0.5593) (50,0.6140) (60,0.6524) (70,0.6776) (80,0.6959) (90,0.7092) (100,0.7276)};
  \addlegendentry{\methodabbrev (Ours)}
  \addplot[method9] coordinates {(0,0.0000) (10,0.2632) (20,0.4313) (30,0.5416) (40,0.6183) (50,0.6711) (60,0.7036) (70,0.7227) (80,0.7324) (90,0.7340) (100,0.7276)};
  \addlegendentry{\methodabbrev $T=3$ (Ours)}

\end{groupplot}

\node[anchor=north, inner sep=0pt, text width=\textwidth, align=center] at
    ($(group c1r4.south)!0.5!(group c2r4.south) + (0,-1cm)$)
    {\pgfplotslegendfromname{grouplegend}};

\end{tikzpicture}

%% file: figures/qual_vitb16/fig.tex
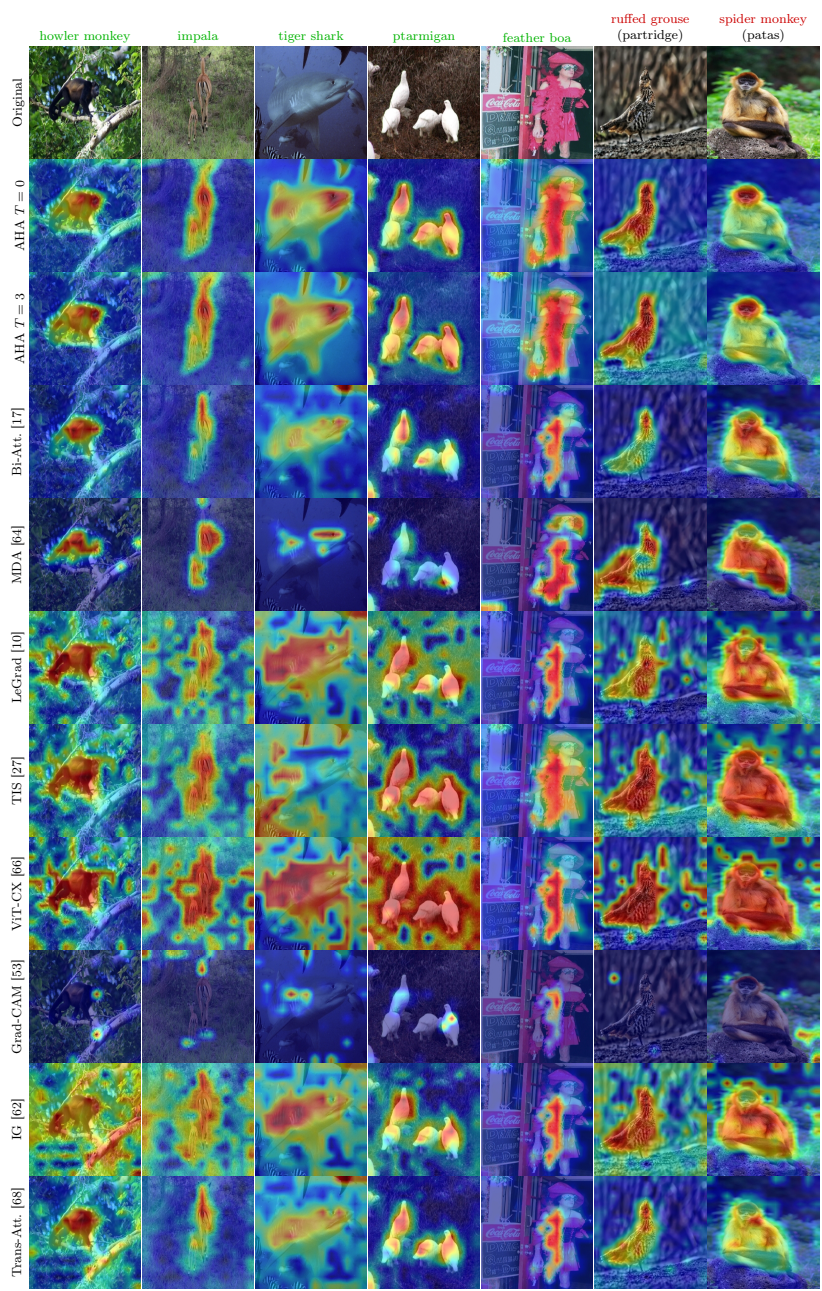
\begin{figure*}[tb]
    \centering
    \resizebox{\textwidth}{!}{
        \input{figures/qual_vitb16/source.tex}
    }
    \caption{
        \textbf{Qualitative comparison of attribution maps} for a ViT-B/16 classifier~\cite{classif:vit:dosovitskiy21} on ImageNet~\cite{dataset:imagenet:deng09} samples.
        The \textcolor{green!70!black}{green} / \textcolor{red!80!black}{red} labels mark correct / incorrect predictions (ground truth in parentheses).
        }
    \label{fig:qual_vitb16}
\end{figure*}

%% file: figures/qual_vitb16/source.tex
\makeatletter
\@ifundefined{qualcolsep}{\newlength{\qualcolsep}}{}
\setlength{\qualcolsep}{1pt}
\@ifundefined{qualrowsep}{\newlength{\qualrowsep}}{}
\setlength{\qualrowsep}{1pt}
\@ifundefined{qualhdrsep}{\newlength{\qualhdrsep}}{}
\setlength{\qualhdrsep}{3pt}
\@ifundefined{quallabelsep}{\newlength{\quallabelsep}}{}
\setlength{\quallabelsep}{3pt}
\@ifundefined{quallabelw}{\newlength{\quallabelw}}{}
\setlength{\quallabelw}{0.35cm}
\@ifundefined{qualimgw}{\newlength{\qualimgw}}{}
\setlength{\qualimgw}{2.4000cm}
\makeatother
\pgfmathsetlengthmacro{\qualfs}{0.115*\qualimgw}
\pgfmathsetlengthmacro{\qualbls}{1.2*\qualfs}

\begin{tikzpicture}[
    every node/.style={inner sep=0pt, outer sep=0pt},
]
  \hypersetup{hidelinks}
  \path[use as bounding box] (0, \qualhdrsep + 2.4*\qualbls) rectangle (\quallabelw + \quallabelsep + 7*\qualimgw + 6*\qualcolsep + \quallabelw + \quallabelsep, -11*\qualimgw - 10*\qualrowsep);
  \node[anchor=south, font={\fontsize{\qualfs}{\qualbls}\selectfont}, text width=\qualimgw, align=center] at (\quallabelw + \quallabelsep + 0*\qualimgw + 0*\qualcolsep + 0.5*\qualimgw, \qualhdrsep) {\textcolor{green!70!black}{howler monkey}};
  \node[anchor=south, font={\fontsize{\qualfs}{\qualbls}\selectfont}, text width=\qualimgw, align=center] at (\quallabelw + \quallabelsep + 1*\qualimgw + 1*\qualcolsep + 0.5*\qualimgw, \qualhdrsep) {\textcolor{green!70!black}{impala}};
  \node[anchor=south, font={\fontsize{\qualfs}{\qualbls}\selectfont}, text width=\qualimgw, align=center] at (\quallabelw + \quallabelsep + 2*\qualimgw + 2*\qualcolsep + 0.5*\qualimgw, \qualhdrsep) {\textcolor{green!70!black}{tiger shark}};
  \node[anchor=south, font={\fontsize{\qualfs}{\qualbls}\selectfont}, text width=\qualimgw, align=center] at (\quallabelw + \quallabelsep + 3*\qualimgw + 3*\qualcolsep + 0.5*\qualimgw, \qualhdrsep) {\textcolor{green!70!black}{ptarmigan}};
  \node[anchor=south, font={\fontsize{\qualfs}{\qualbls}\selectfont}, text width=\qualimgw, align=center] at (\quallabelw + \quallabelsep + 4*\qualimgw + 4*\qualcolsep + 0.5*\qualimgw, \qualhdrsep) {\textcolor{green!70!black}{feather boa}};
  \node[anchor=south, font={\fontsize{\qualfs}{\qualbls}\selectfont}, text width=\qualimgw, align=center] at (\quallabelw + \quallabelsep + 5*\qualimgw + 5*\qualcolsep + 0.5*\qualimgw, \qualhdrsep) {\textcolor{red!80!black}{ruffed grouse} \\ (partridge)};
  \node[anchor=south, font={\fontsize{\qualfs}{\qualbls}\selectfont}, text width=\qualimgw, align=center] at (\quallabelw + \quallabelsep + 6*\qualimgw + 6*\qualcolsep + 0.5*\qualimgw, \qualhdrsep) {\textcolor{red!80!black}{spider monkey} \\ (patas)};
  \node[anchor=south, overlay, font={\fontsize{\qualfs}{\qualbls}\selectfont}, rotate=90] at (\quallabelw, -0*\qualimgw - 0*\qualrowsep - 0.5*\qualimgw) {Original};
  \node[anchor=north west] at (\quallabelw + \quallabelsep + 0*\qualimgw + 0*\qualcolsep, -0*\qualimgw - 0*\qualrowsep) {\includegraphics[width=\qualimgw]{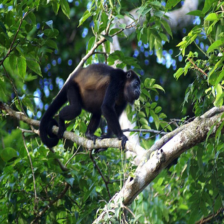}};
  \node[anchor=north west] at (\quallabelw + \quallabelsep + 1*\qualimgw + 1*\qualcolsep, -0*\qualimgw - 0*\qualrowsep) {\includegraphics[width=\qualimgw]{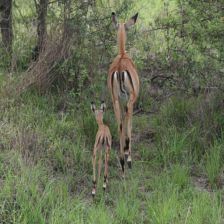}};
  \node[anchor=north west] at (\quallabelw + \quallabelsep + 2*\qualimgw + 2*\qualcolsep, -0*\qualimgw - 0*\qualrowsep) {\includegraphics[width=\qualimgw]{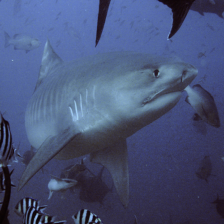}};
  \node[anchor=north west] at (\quallabelw + \quallabelsep + 3*\qualimgw + 3*\qualcolsep, -0*\qualimgw - 0*\qualrowsep) {\includegraphics[width=\qualimgw]{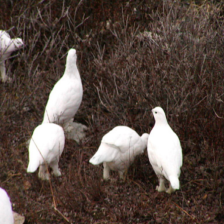}};
  \node[anchor=north west] at (\quallabelw + \quallabelsep + 4*\qualimgw + 4*\qualcolsep, -0*\qualimgw - 0*\qualrowsep) {\includegraphics[width=\qualimgw]{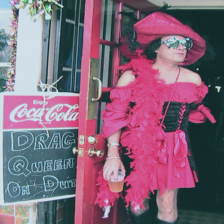}};
  \node[anchor=north west] at (\quallabelw + \quallabelsep + 5*\qualimgw + 5*\qualcolsep, -0*\qualimgw - 0*\qualrowsep) {\includegraphics[width=\qualimgw]{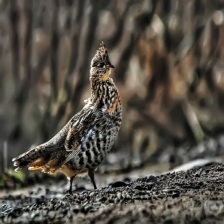}};
  \node[anchor=north west] at (\quallabelw + \quallabelsep + 6*\qualimgw + 6*\qualcolsep, -0*\qualimgw - 0*\qualrowsep) {\includegraphics[width=\qualimgw]{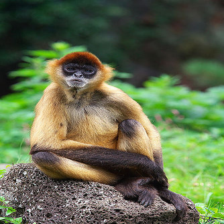}};
  \node[anchor=south, overlay, font={\fontsize{\qualfs}{\qualbls}\selectfont}, rotate=90] at (\quallabelw, -1*\qualimgw - 1*\qualrowsep - 0.5*\qualimgw) {\methodabbrev $T=0$};
  \node[anchor=north west] at (\quallabelw + \quallabelsep + 0*\qualimgw + 0*\qualcolsep, -1*\qualimgw - 1*\qualrowsep) {\includegraphics[width=\qualimgw]{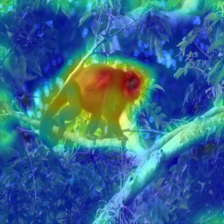}};
  \node[anchor=north west] at (\quallabelw + \quallabelsep + 1*\qualimgw + 1*\qualcolsep, -1*\qualimgw - 1*\qualrowsep) {\includegraphics[width=\qualimgw]{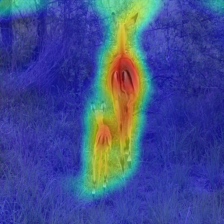}};
  \node[anchor=north west] at (\quallabelw + \quallabelsep + 2*\qualimgw + 2*\qualcolsep, -1*\qualimgw - 1*\qualrowsep) {\includegraphics[width=\qualimgw]{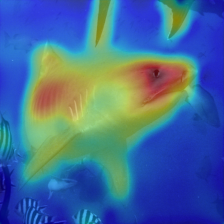}};
  \node[anchor=north west] at (\quallabelw + \quallabelsep + 3*\qualimgw + 3*\qualcolsep, -1*\qualimgw - 1*\qualrowsep) {\includegraphics[width=\qualimgw]{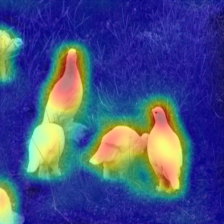}};
  \node[anchor=north west] at (\quallabelw + \quallabelsep + 4*\qualimgw + 4*\qualcolsep, -1*\qualimgw - 1*\qualrowsep) {\includegraphics[width=\qualimgw]{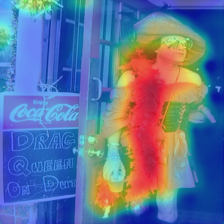}};
  \node[anchor=north west] at (\quallabelw + \quallabelsep + 5*\qualimgw + 5*\qualcolsep, -1*\qualimgw - 1*\qualrowsep) {\includegraphics[width=\qualimgw]{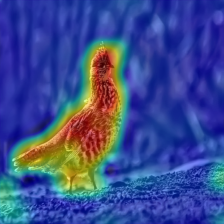}};
  \node[anchor=north west] at (\quallabelw + \quallabelsep + 6*\qualimgw + 6*\qualcolsep, -1*\qualimgw - 1*\qualrowsep) {\includegraphics[width=\qualimgw]{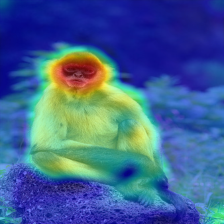}};
  \node[anchor=south, overlay, font={\fontsize{\qualfs}{\qualbls}\selectfont}, rotate=90] at (\quallabelw, -2*\qualimgw - 2*\qualrowsep - 0.5*\qualimgw) {\methodabbrev $T=3$};
  \node[anchor=north west] at (\quallabelw + \quallabelsep + 0*\qualimgw + 0*\qualcolsep, -2*\qualimgw - 2*\qualrowsep) {\includegraphics[width=\qualimgw]{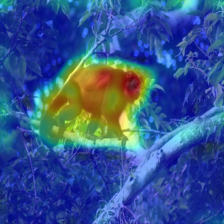}};
  \node[anchor=north west] at (\quallabelw + \quallabelsep + 1*\qualimgw + 1*\qualcolsep, -2*\qualimgw - 2*\qualrowsep) {\includegraphics[width=\qualimgw]{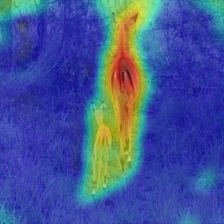}};
  \node[anchor=north west] at (\quallabelw + \quallabelsep + 2*\qualimgw + 2*\qualcolsep, -2*\qualimgw - 2*\qualrowsep) {\includegraphics[width=\qualimgw]{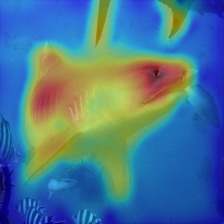}};
  \node[anchor=north west] at (\quallabelw + \quallabelsep + 3*\qualimgw + 3*\qualcolsep, -2*\qualimgw - 2*\qualrowsep) {\includegraphics[width=\qualimgw]{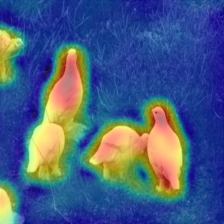}};
  \node[anchor=north west] at (\quallabelw + \quallabelsep + 4*\qualimgw + 4*\qualcolsep, -2*\qualimgw - 2*\qualrowsep) {\includegraphics[width=\qualimgw]{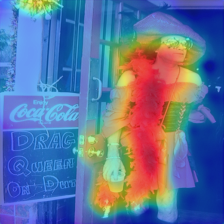}};
  \node[anchor=north west] at (\quallabelw + \quallabelsep + 5*\qualimgw + 5*\qualcolsep, -2*\qualimgw - 2*\qualrowsep) {\includegraphics[width=\qualimgw]{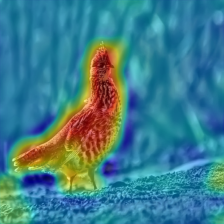}};
  \node[anchor=north west] at (\quallabelw + \quallabelsep + 6*\qualimgw + 6*\qualcolsep, -2*\qualimgw - 2*\qualrowsep) {\includegraphics[width=\qualimgw]{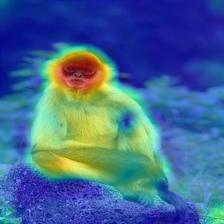}};
  \node[anchor=south, overlay, font={\fontsize{\qualfs}{\qualbls}\selectfont}, rotate=90] at (\quallabelw, -3*\qualimgw - 3*\qualrowsep - 0.5*\qualimgw) {Bi-Att.~\cite{expl:vit:bidirectionalatt:chen23}};
  \node[anchor=north west] at (\quallabelw + \quallabelsep + 0*\qualimgw + 0*\qualcolsep, -3*\qualimgw - 3*\qualrowsep) {\includegraphics[width=\qualimgw]{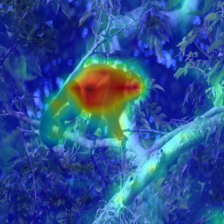}};
  \node[anchor=north west] at (\quallabelw + \quallabelsep + 1*\qualimgw + 1*\qualcolsep, -3*\qualimgw - 3*\qualrowsep) {\includegraphics[width=\qualimgw]{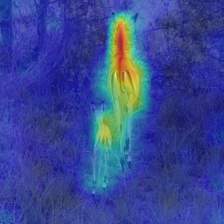}};
  \node[anchor=north west] at (\quallabelw + \quallabelsep + 2*\qualimgw + 2*\qualcolsep, -3*\qualimgw - 3*\qualrowsep) {\includegraphics[width=\qualimgw]{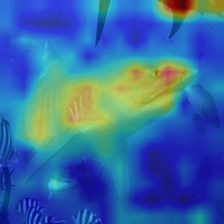}};
  \node[anchor=north west] at (\quallabelw + \quallabelsep + 3*\qualimgw + 3*\qualcolsep, -3*\qualimgw - 3*\qualrowsep) {\includegraphics[width=\qualimgw]{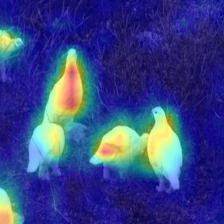}};
  \node[anchor=north west] at (\quallabelw + \quallabelsep + 4*\qualimgw + 4*\qualcolsep, -3*\qualimgw - 3*\qualrowsep) {\includegraphics[width=\qualimgw]{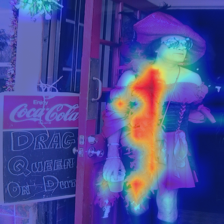}};
  \node[anchor=north west] at (\quallabelw + \quallabelsep + 5*\qualimgw + 5*\qualcolsep, -3*\qualimgw - 3*\qualrowsep) {\includegraphics[width=\qualimgw]{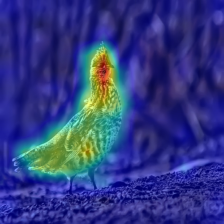}};
  \node[anchor=north west] at (\quallabelw + \quallabelsep + 6*\qualimgw + 6*\qualcolsep, -3*\qualimgw - 3*\qualrowsep) {\includegraphics[width=\qualimgw]{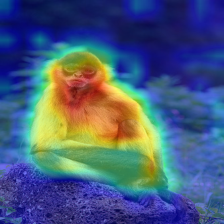}};
  \node[anchor=south, overlay, font={\fontsize{\qualfs}{\qualbls}\selectfont}, rotate=90] at (\quallabelw, -4*\qualimgw - 4*\qualrowsep - 0.5*\qualimgw) {MDA~\cite{expl:vit:mda:walker25}};
  \node[anchor=north west] at (\quallabelw + \quallabelsep + 0*\qualimgw + 0*\qualcolsep, -4*\qualimgw - 4*\qualrowsep) {\includegraphics[width=\qualimgw]{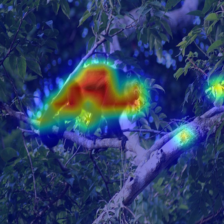}};
  \node[anchor=north west] at (\quallabelw + \quallabelsep + 1*\qualimgw + 1*\qualcolsep, -4*\qualimgw - 4*\qualrowsep) {\includegraphics[width=\qualimgw]{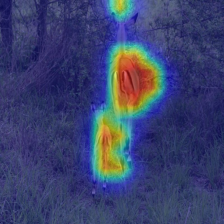}};
  \node[anchor=north west] at (\quallabelw + \quallabelsep + 2*\qualimgw + 2*\qualcolsep, -4*\qualimgw - 4*\qualrowsep) {\includegraphics[width=\qualimgw]{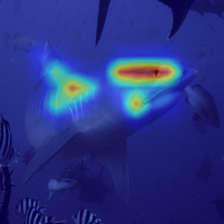}};
  \node[anchor=north west] at (\quallabelw + \quallabelsep + 3*\qualimgw + 3*\qualcolsep, -4*\qualimgw - 4*\qualrowsep) {\includegraphics[width=\qualimgw]{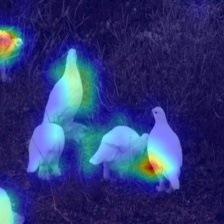}};
  \node[anchor=north west] at (\quallabelw + \quallabelsep + 4*\qualimgw + 4*\qualcolsep, -4*\qualimgw - 4*\qualrowsep) {\includegraphics[width=\qualimgw]{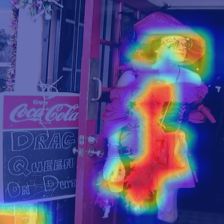}};
  \node[anchor=north west] at (\quallabelw + \quallabelsep + 5*\qualimgw + 5*\qualcolsep, -4*\qualimgw - 4*\qualrowsep) {\includegraphics[width=\qualimgw]{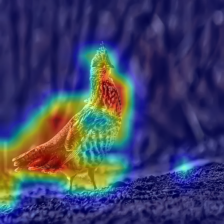}};
  \node[anchor=north west] at (\quallabelw + \quallabelsep + 6*\qualimgw + 6*\qualcolsep, -4*\qualimgw - 4*\qualrowsep) {\includegraphics[width=\qualimgw]{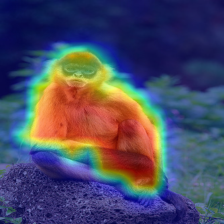}};
  \node[anchor=south, overlay, font={\fontsize{\qualfs}{\qualbls}\selectfont}, rotate=90] at (\quallabelw, -5*\qualimgw - 5*\qualrowsep - 0.5*\qualimgw) {LeGrad~\cite{expl:vit:legrad:bousselham25}};
  \node[anchor=north west] at (\quallabelw + \quallabelsep + 0*\qualimgw + 0*\qualcolsep, -5*\qualimgw - 5*\qualrowsep) {\includegraphics[width=\qualimgw]{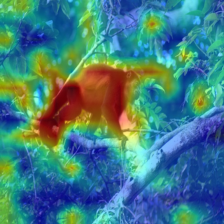}};
  \node[anchor=north west] at (\quallabelw + \quallabelsep + 1*\qualimgw + 1*\qualcolsep, -5*\qualimgw - 5*\qualrowsep) {\includegraphics[width=\qualimgw]{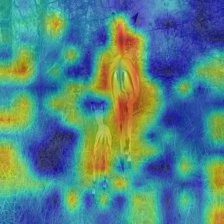}};
  \node[anchor=north west] at (\quallabelw + \quallabelsep + 2*\qualimgw + 2*\qualcolsep, -5*\qualimgw - 5*\qualrowsep) {\includegraphics[width=\qualimgw]{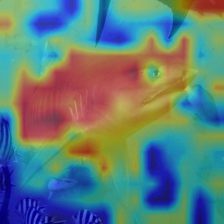}};
  \node[anchor=north west] at (\quallabelw + \quallabelsep + 3*\qualimgw + 3*\qualcolsep, -5*\qualimgw - 5*\qualrowsep) {\includegraphics[width=\qualimgw]{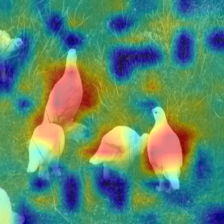}};
  \node[anchor=north west] at (\quallabelw + \quallabelsep + 4*\qualimgw + 4*\qualcolsep, -5*\qualimgw - 5*\qualrowsep) {\includegraphics[width=\qualimgw]{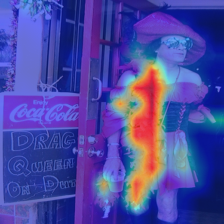}};
  \node[anchor=north west] at (\quallabelw + \quallabelsep + 5*\qualimgw + 5*\qualcolsep, -5*\qualimgw - 5*\qualrowsep) {\includegraphics[width=\qualimgw]{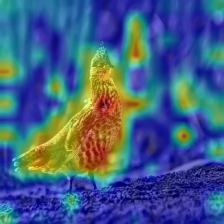}};
  \node[anchor=north west] at (\quallabelw + \quallabelsep + 6*\qualimgw + 6*\qualcolsep, -5*\qualimgw - 5*\qualrowsep) {\includegraphics[width=\qualimgw]{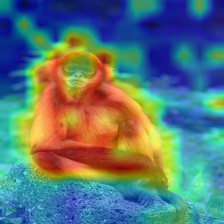}};
  \node[anchor=south, overlay, font={\fontsize{\qualfs}{\qualbls}\selectfont}, rotate=90] at (\quallabelw, -6*\qualimgw - 6*\qualrowsep - 0.5*\qualimgw) {TIS~\cite{expl:vit:tis:englebert23}};
  \node[anchor=north west] at (\quallabelw + \quallabelsep + 0*\qualimgw + 0*\qualcolsep, -6*\qualimgw - 6*\qualrowsep) {\includegraphics[width=\qualimgw]{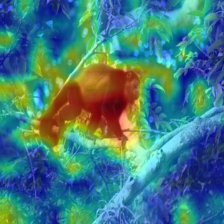}};
  \node[anchor=north west] at (\quallabelw + \quallabelsep + 1*\qualimgw + 1*\qualcolsep, -6*\qualimgw - 6*\qualrowsep) {\includegraphics[width=\qualimgw]{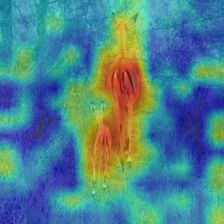}};
  \node[anchor=north west] at (\quallabelw + \quallabelsep + 2*\qualimgw + 2*\qualcolsep, -6*\qualimgw - 6*\qualrowsep) {\includegraphics[width=\qualimgw]{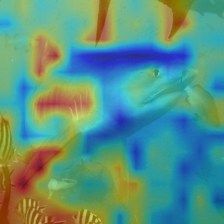}};
  \node[anchor=north west] at (\quallabelw + \quallabelsep + 3*\qualimgw + 3*\qualcolsep, -6*\qualimgw - 6*\qualrowsep) {\includegraphics[width=\qualimgw]{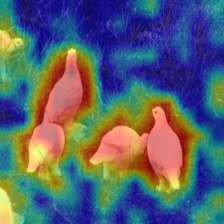}};
  \node[anchor=north west] at (\quallabelw + \quallabelsep + 4*\qualimgw + 4*\qualcolsep, -6*\qualimgw - 6*\qualrowsep) {\includegraphics[width=\qualimgw]{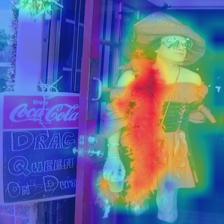}};
  \node[anchor=north west] at (\quallabelw + \quallabelsep + 5*\qualimgw + 5*\qualcolsep, -6*\qualimgw - 6*\qualrowsep) {\includegraphics[width=\qualimgw]{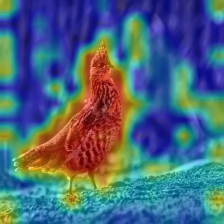}};
  \node[anchor=north west] at (\quallabelw + \quallabelsep + 6*\qualimgw + 6*\qualcolsep, -6*\qualimgw - 6*\qualrowsep) {\includegraphics[width=\qualimgw]{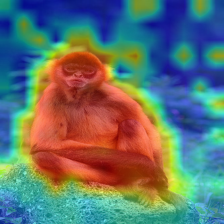}};
  \node[anchor=south, overlay, font={\fontsize{\qualfs}{\qualbls}\selectfont}, rotate=90] at (\quallabelw, -7*\qualimgw - 7*\qualrowsep - 0.5*\qualimgw) {ViT-CX~\cite{expl:vit:vitcx:xie23}};
  \node[anchor=north west] at (\quallabelw + \quallabelsep + 0*\qualimgw + 0*\qualcolsep, -7*\qualimgw - 7*\qualrowsep) {\includegraphics[width=\qualimgw]{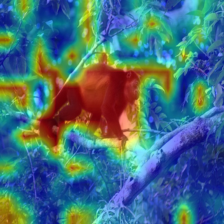}};
  \node[anchor=north west] at (\quallabelw + \quallabelsep + 1*\qualimgw + 1*\qualcolsep, -7*\qualimgw - 7*\qualrowsep) {\includegraphics[width=\qualimgw]{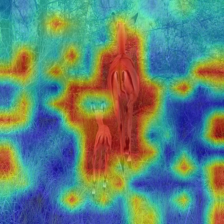}};
  \node[anchor=north west] at (\quallabelw + \quallabelsep + 2*\qualimgw + 2*\qualcolsep, -7*\qualimgw - 7*\qualrowsep) {\includegraphics[width=\qualimgw]{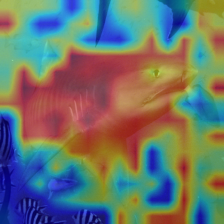}};
  \node[anchor=north west] at (\quallabelw + \quallabelsep + 3*\qualimgw + 3*\qualcolsep, -7*\qualimgw - 7*\qualrowsep) {\includegraphics[width=\qualimgw]{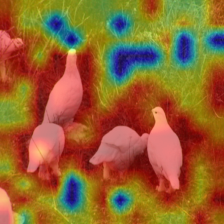}};
  \node[anchor=north west] at (\quallabelw + \quallabelsep + 4*\qualimgw + 4*\qualcolsep, -7*\qualimgw - 7*\qualrowsep) {\includegraphics[width=\qualimgw]{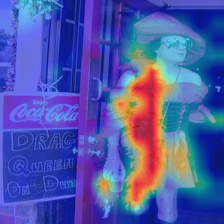}};
  \node[anchor=north west] at (\quallabelw + \quallabelsep + 5*\qualimgw + 5*\qualcolsep, -7*\qualimgw - 7*\qualrowsep) {\includegraphics[width=\qualimgw]{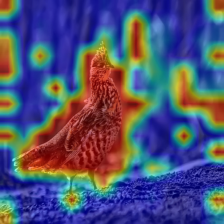}};
  \node[anchor=north west] at (\quallabelw + \quallabelsep + 6*\qualimgw + 6*\qualcolsep, -7*\qualimgw - 7*\qualrowsep) {\includegraphics[width=\qualimgw]{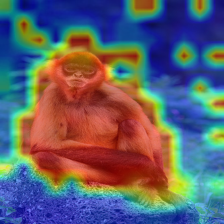}};
  \node[anchor=south, overlay, font={\fontsize{\qualfs}{\qualbls}\selectfont}, rotate=90] at (\quallabelw, -8*\qualimgw - 8*\qualrowsep - 0.5*\qualimgw) {Grad-CAM~\cite{expl:act:gradcam:selvaraju17}};
  \node[anchor=north west] at (\quallabelw + \quallabelsep + 0*\qualimgw + 0*\qualcolsep, -8*\qualimgw - 8*\qualrowsep) {\includegraphics[width=\qualimgw]{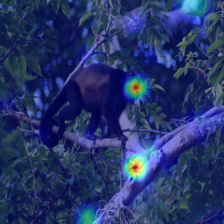}};
  \node[anchor=north west] at (\quallabelw + \quallabelsep + 1*\qualimgw + 1*\qualcolsep, -8*\qualimgw - 8*\qualrowsep) {\includegraphics[width=\qualimgw]{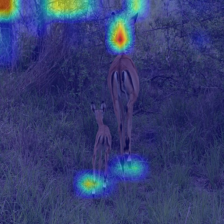}};
  \node[anchor=north west] at (\quallabelw + \quallabelsep + 2*\qualimgw + 2*\qualcolsep, -8*\qualimgw - 8*\qualrowsep) {\includegraphics[width=\qualimgw]{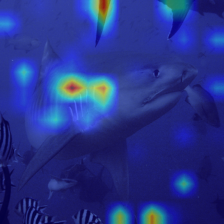}};
  \node[anchor=north west] at (\quallabelw + \quallabelsep + 3*\qualimgw + 3*\qualcolsep, -8*\qualimgw - 8*\qualrowsep) {\includegraphics[width=\qualimgw]{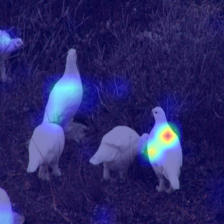}};
  \node[anchor=north west] at (\quallabelw + \quallabelsep + 4*\qualimgw + 4*\qualcolsep, -8*\qualimgw - 8*\qualrowsep) {\includegraphics[width=\qualimgw]{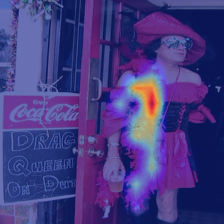}};
  \node[anchor=north west] at (\quallabelw + \quallabelsep + 5*\qualimgw + 5*\qualcolsep, -8*\qualimgw - 8*\qualrowsep) {\includegraphics[width=\qualimgw]{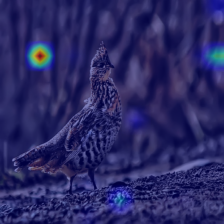}};
  \node[anchor=north west] at (\quallabelw + \quallabelsep + 6*\qualimgw + 6*\qualcolsep, -8*\qualimgw - 8*\qualrowsep) {\includegraphics[width=\qualimgw]{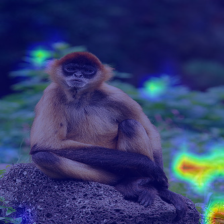}};
  \node[anchor=south, overlay, font={\fontsize{\qualfs}{\qualbls}\selectfont}, rotate=90] at (\quallabelw, -9*\qualimgw - 9*\qualrowsep - 0.5*\qualimgw) {IG~\cite{expl:grad:integratedgradients:sundararajan17}};
  \node[anchor=north west] at (\quallabelw + \quallabelsep + 0*\qualimgw + 0*\qualcolsep, -9*\qualimgw - 9*\qualrowsep) {\includegraphics[width=\qualimgw]{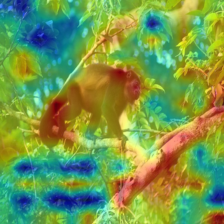}};
  \node[anchor=north west] at (\quallabelw + \quallabelsep + 1*\qualimgw + 1*\qualcolsep, -9*\qualimgw - 9*\qualrowsep) {\includegraphics[width=\qualimgw]{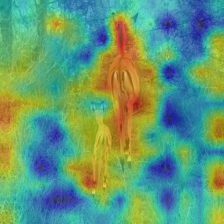}};
  \node[anchor=north west] at (\quallabelw + \quallabelsep + 2*\qualimgw + 2*\qualcolsep, -9*\qualimgw - 9*\qualrowsep) {\includegraphics[width=\qualimgw]{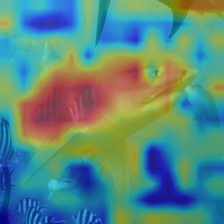}};
  \node[anchor=north west] at (\quallabelw + \quallabelsep + 3*\qualimgw + 3*\qualcolsep, -9*\qualimgw - 9*\qualrowsep) {\includegraphics[width=\qualimgw]{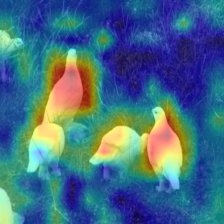}};
  \node[anchor=north west] at (\quallabelw + \quallabelsep + 4*\qualimgw + 4*\qualcolsep, -9*\qualimgw - 9*\qualrowsep) {\includegraphics[width=\qualimgw]{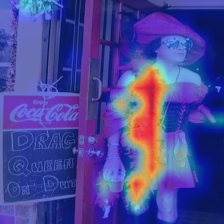}};
  \node[anchor=north west] at (\quallabelw + \quallabelsep + 5*\qualimgw + 5*\qualcolsep, -9*\qualimgw - 9*\qualrowsep) {\includegraphics[width=\qualimgw]{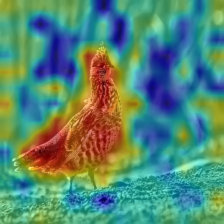}};
  \node[anchor=north west] at (\quallabelw + \quallabelsep + 6*\qualimgw + 6*\qualcolsep, -9*\qualimgw - 9*\qualrowsep) {\includegraphics[width=\qualimgw]{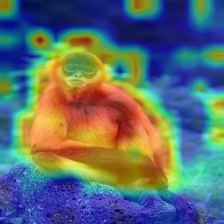}};
  \node[anchor=south, overlay, font={\fontsize{\qualfs}{\qualbls}\selectfont}, rotate=90] at (\quallabelw, -10*\qualimgw - 10*\qualrowsep - 0.5*\qualimgw) {Trans-Att.~\cite{expl:vit:transatt:yuan21}};
  \node[anchor=north west] at (\quallabelw + \quallabelsep + 0*\qualimgw + 0*\qualcolsep, -10*\qualimgw - 10*\qualrowsep) {\includegraphics[width=\qualimgw]{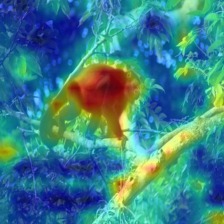}};
  \node[anchor=north west] at (\quallabelw + \quallabelsep + 1*\qualimgw + 1*\qualcolsep, -10*\qualimgw - 10*\qualrowsep) {\includegraphics[width=\qualimgw]{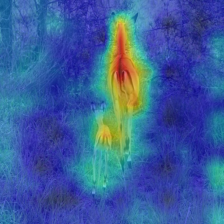}};
  \node[anchor=north west] at (\quallabelw + \quallabelsep + 2*\qualimgw + 2*\qualcolsep, -10*\qualimgw - 10*\qualrowsep) {\includegraphics[width=\qualimgw]{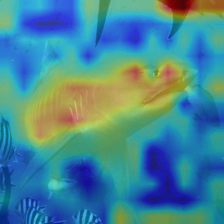}};
  \node[anchor=north west] at (\quallabelw + \quallabelsep + 3*\qualimgw + 3*\qualcolsep, -10*\qualimgw - 10*\qualrowsep) {\includegraphics[width=\qualimgw]{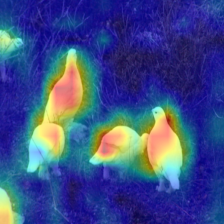}};
  \node[anchor=north west] at (\quallabelw + \quallabelsep + 4*\qualimgw + 4*\qualcolsep, -10*\qualimgw - 10*\qualrowsep) {\includegraphics[width=\qualimgw]{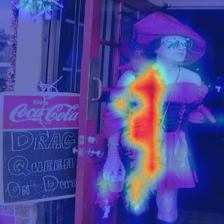}};
  \node[anchor=north west] at (\quallabelw + \quallabelsep + 5*\qualimgw + 5*\qualcolsep, -10*\qualimgw - 10*\qualrowsep) {\includegraphics[width=\qualimgw]{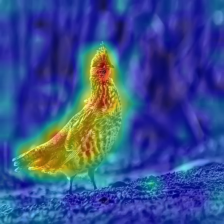}};
  \node[anchor=north west] at (\quallabelw + \quallabelsep + 6*\qualimgw + 6*\qualcolsep, -10*\qualimgw - 10*\qualrowsep) {\includegraphics[width=\qualimgw]{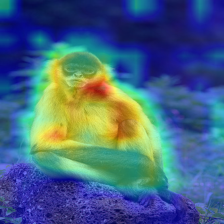}};
\end{tikzpicture}

%% file: figures/qual_vitb32/fig.tex
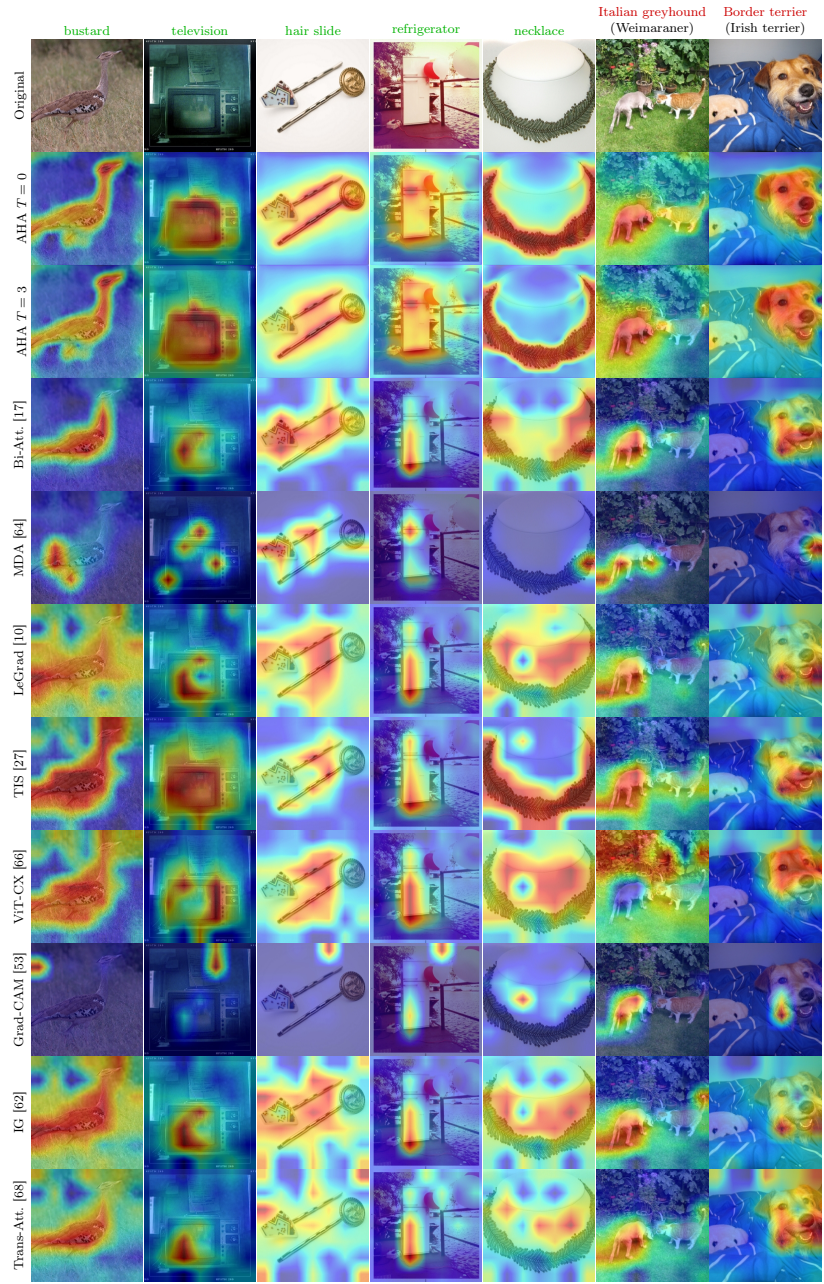
\begin{figure*}[tb]
    \centering
    \resizebox{\textwidth}{!}{
        \input{figures/qual_vitb32/source.tex}
    }
    \caption{
        \textbf{Qualitative comparison of attribution maps} for a ViT-B/32 classifier~\cite{classif:vit:dosovitskiy21} on ImageNet~\cite{dataset:imagenet:deng09} samples.
        The \textcolor{green!70!black}{green} / \textcolor{red!80!black}{red} labels mark correct / incorrect predictions (ground truth in parentheses).
        }
    \label{fig:qual_vitb32}
\end{figure*}

%% file: figures/qual_vitb32/source.tex
\makeatletter
\@ifundefined{qualcolsep}{\newlength{\qualcolsep}}{}
\setlength{\qualcolsep}{1pt}
\@ifundefined{qualrowsep}{\newlength{\qualrowsep}}{}
\setlength{\qualrowsep}{1pt}
\@ifundefined{qualhdrsep}{\newlength{\qualhdrsep}}{}
\setlength{\qualhdrsep}{3pt}
\@ifundefined{quallabelsep}{\newlength{\quallabelsep}}{}
\setlength{\quallabelsep}{3pt}
\@ifundefined{quallabelw}{\newlength{\quallabelw}}{}
\setlength{\quallabelw}{0.35cm}
\@ifundefined{qualimgw}{\newlength{\qualimgw}}{}
\setlength{\qualimgw}{2.4000cm}
\makeatother
\pgfmathsetlengthmacro{\qualfs}{0.115*\qualimgw}
\pgfmathsetlengthmacro{\qualbls}{1.2*\qualfs}

\begin{tikzpicture}[
    every node/.style={inner sep=0pt, outer sep=0pt},
]
  \hypersetup{hidelinks}
  \path[use as bounding box] (0, \qualhdrsep + 2.4*\qualbls) rectangle (\quallabelw + \quallabelsep + 7*\qualimgw + 6*\qualcolsep + \quallabelw + \quallabelsep, -11*\qualimgw - 10*\qualrowsep);
  \node[anchor=south, font={\fontsize{\qualfs}{\qualbls}\selectfont}, text width=\qualimgw, align=center] at (\quallabelw + \quallabelsep + 0*\qualimgw + 0*\qualcolsep + 0.5*\qualimgw, \qualhdrsep) {\textcolor{green!70!black}{bustard}};
  \node[anchor=south, font={\fontsize{\qualfs}{\qualbls}\selectfont}, text width=\qualimgw, align=center] at (\quallabelw + \quallabelsep + 1*\qualimgw + 1*\qualcolsep + 0.5*\qualimgw, \qualhdrsep) {\textcolor{green!70!black}{television}};
  \node[anchor=south, font={\fontsize{\qualfs}{\qualbls}\selectfont}, text width=\qualimgw, align=center] at (\quallabelw + \quallabelsep + 2*\qualimgw + 2*\qualcolsep + 0.5*\qualimgw, \qualhdrsep) {\textcolor{green!70!black}{hair slide}};
  \node[anchor=south, font={\fontsize{\qualfs}{\qualbls}\selectfont}, text width=\qualimgw, align=center] at (\quallabelw + \quallabelsep + 3*\qualimgw + 3*\qualcolsep + 0.5*\qualimgw, \qualhdrsep) {\textcolor{green!70!black}{refrigerator}};
  \node[anchor=south, font={\fontsize{\qualfs}{\qualbls}\selectfont}, text width=\qualimgw, align=center] at (\quallabelw + \quallabelsep + 4*\qualimgw + 4*\qualcolsep + 0.5*\qualimgw, \qualhdrsep) {\textcolor{green!70!black}{necklace}};
  \node[anchor=south, font={\fontsize{\qualfs}{\qualbls}\selectfont}, text width=\qualimgw, align=center] at (\quallabelw + \quallabelsep + 5*\qualimgw + 5*\qualcolsep + 0.5*\qualimgw, \qualhdrsep) {\textcolor{red!80!black}{Italian greyhound} \\ (Weimaraner)};
  \node[anchor=south, font={\fontsize{\qualfs}{\qualbls}\selectfont}, text width=\qualimgw, align=center] at (\quallabelw + \quallabelsep + 6*\qualimgw + 6*\qualcolsep + 0.5*\qualimgw, \qualhdrsep) {\textcolor{red!80!black}{Border terrier} \\ (Irish terrier)};
  \node[anchor=south, overlay, font={\fontsize{\qualfs}{\qualbls}\selectfont}, rotate=90] at (\quallabelw, -0*\qualimgw - 0*\qualrowsep - 0.5*\qualimgw) {Original};
  \node[anchor=north west] at (\quallabelw + \quallabelsep + 0*\qualimgw + 0*\qualcolsep, -0*\qualimgw - 0*\qualrowsep) {\includegraphics[width=\qualimgw]{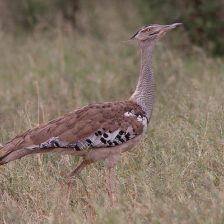}};
  \node[anchor=north west] at (\quallabelw + \quallabelsep + 1*\qualimgw + 1*\qualcolsep, -0*\qualimgw - 0*\qualrowsep) {\includegraphics[width=\qualimgw]{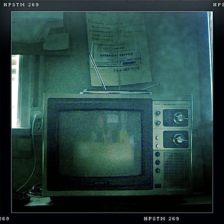}};
  \node[anchor=north west] at (\quallabelw + \quallabelsep + 2*\qualimgw + 2*\qualcolsep, -0*\qualimgw - 0*\qualrowsep) {\includegraphics[width=\qualimgw]{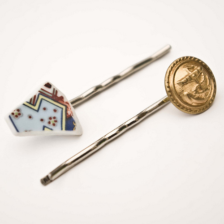}};
  \node[anchor=north west] at (\quallabelw + \quallabelsep + 3*\qualimgw + 3*\qualcolsep, -0*\qualimgw - 0*\qualrowsep) {\includegraphics[width=\qualimgw]{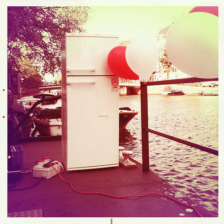}};
  \node[anchor=north west] at (\quallabelw + \quallabelsep + 4*\qualimgw + 4*\qualcolsep, -0*\qualimgw - 0*\qualrowsep) {\includegraphics[width=\qualimgw]{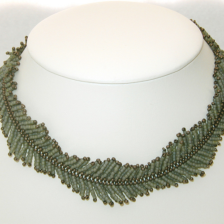}};
  \node[anchor=north west] at (\quallabelw + \quallabelsep + 5*\qualimgw + 5*\qualcolsep, -0*\qualimgw - 0*\qualrowsep) {\includegraphics[width=\qualimgw]{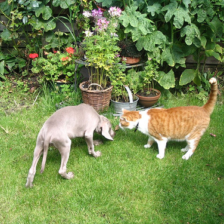}};
  \node[anchor=north west] at (\quallabelw + \quallabelsep + 6*\qualimgw + 6*\qualcolsep, -0*\qualimgw - 0*\qualrowsep) {\includegraphics[width=\qualimgw]{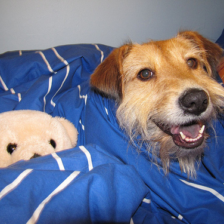}};
  \node[anchor=south, overlay, font={\fontsize{\qualfs}{\qualbls}\selectfont}, rotate=90] at (\quallabelw, -1*\qualimgw - 1*\qualrowsep - 0.5*\qualimgw) {\methodabbrev $T=0$};
  \node[anchor=north west] at (\quallabelw + \quallabelsep + 0*\qualimgw + 0*\qualcolsep, -1*\qualimgw - 1*\qualrowsep) {\includegraphics[width=\qualimgw]{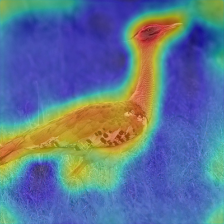}};
  \node[anchor=north west] at (\quallabelw + \quallabelsep + 1*\qualimgw + 1*\qualcolsep, -1*\qualimgw - 1*\qualrowsep) {\includegraphics[width=\qualimgw]{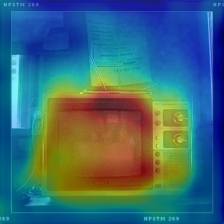}};
  \node[anchor=north west] at (\quallabelw + \quallabelsep + 2*\qualimgw + 2*\qualcolsep, -1*\qualimgw - 1*\qualrowsep) {\includegraphics[width=\qualimgw]{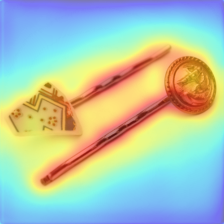}};
  \node[anchor=north west] at (\quallabelw + \quallabelsep + 3*\qualimgw + 3*\qualcolsep, -1*\qualimgw - 1*\qualrowsep) {\includegraphics[width=\qualimgw]{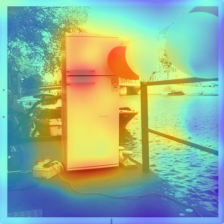}};
  \node[anchor=north west] at (\quallabelw + \quallabelsep + 4*\qualimgw + 4*\qualcolsep, -1*\qualimgw - 1*\qualrowsep) {\includegraphics[width=\qualimgw]{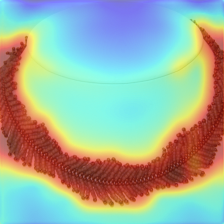}};
  \node[anchor=north west] at (\quallabelw + \quallabelsep + 5*\qualimgw + 5*\qualcolsep, -1*\qualimgw - 1*\qualrowsep) {\includegraphics[width=\qualimgw]{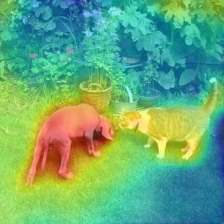}};
  \node[anchor=north west] at (\quallabelw + \quallabelsep + 6*\qualimgw + 6*\qualcolsep, -1*\qualimgw - 1*\qualrowsep) {\includegraphics[width=\qualimgw]{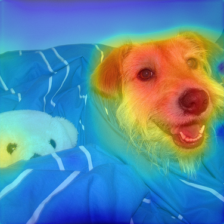}};
  \node[anchor=south, overlay, font={\fontsize{\qualfs}{\qualbls}\selectfont}, rotate=90] at (\quallabelw, -2*\qualimgw - 2*\qualrowsep - 0.5*\qualimgw) {\methodabbrev $T=3$};
  \node[anchor=north west] at (\quallabelw + \quallabelsep + 0*\qualimgw + 0*\qualcolsep, -2*\qualimgw - 2*\qualrowsep) {\includegraphics[width=\qualimgw]{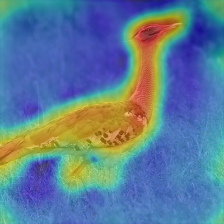}};
  \node[anchor=north west] at (\quallabelw + \quallabelsep + 1*\qualimgw + 1*\qualcolsep, -2*\qualimgw - 2*\qualrowsep) {\includegraphics[width=\qualimgw]{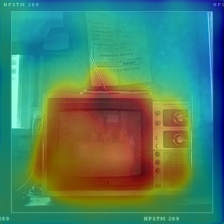}};
  \node[anchor=north west] at (\quallabelw + \quallabelsep + 2*\qualimgw + 2*\qualcolsep, -2*\qualimgw - 2*\qualrowsep) {\includegraphics[width=\qualimgw]{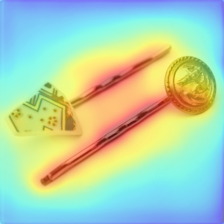}};
  \node[anchor=north west] at (\quallabelw + \quallabelsep + 3*\qualimgw + 3*\qualcolsep, -2*\qualimgw - 2*\qualrowsep) {\includegraphics[width=\qualimgw]{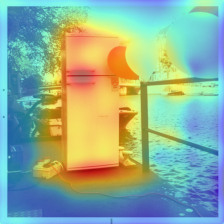}};
  \node[anchor=north west] at (\quallabelw + \quallabelsep + 4*\qualimgw + 4*\qualcolsep, -2*\qualimgw - 2*\qualrowsep) {\includegraphics[width=\qualimgw]{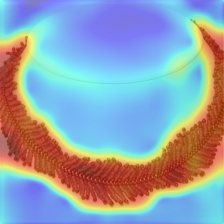}};
  \node[anchor=north west] at (\quallabelw + \quallabelsep + 5*\qualimgw + 5*\qualcolsep, -2*\qualimgw - 2*\qualrowsep) {\includegraphics[width=\qualimgw]{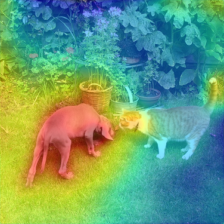}};
  \node[anchor=north west] at (\quallabelw + \quallabelsep + 6*\qualimgw + 6*\qualcolsep, -2*\qualimgw - 2*\qualrowsep) {\includegraphics[width=\qualimgw]{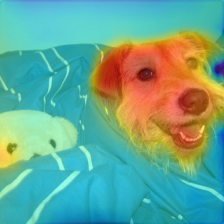}};
  \node[anchor=south, overlay, font={\fontsize{\qualfs}{\qualbls}\selectfont}, rotate=90] at (\quallabelw, -3*\qualimgw - 3*\qualrowsep - 0.5*\qualimgw) {Bi-Att.~\cite{expl:vit:bidirectionalatt:chen23}};
  \node[anchor=north west] at (\quallabelw + \quallabelsep + 0*\qualimgw + 0*\qualcolsep, -3*\qualimgw - 3*\qualrowsep) {\includegraphics[width=\qualimgw]{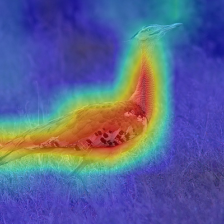}};
  \node[anchor=north west] at (\quallabelw + \quallabelsep + 1*\qualimgw + 1*\qualcolsep, -3*\qualimgw - 3*\qualrowsep) {\includegraphics[width=\qualimgw]{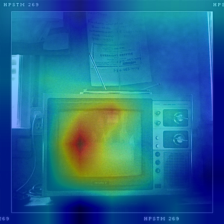}};
  \node[anchor=north west] at (\quallabelw + \quallabelsep + 2*\qualimgw + 2*\qualcolsep, -3*\qualimgw - 3*\qualrowsep) {\includegraphics[width=\qualimgw]{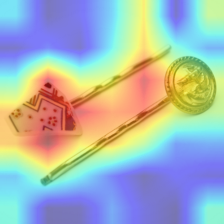}};
  \node[anchor=north west] at (\quallabelw + \quallabelsep + 3*\qualimgw + 3*\qualcolsep, -3*\qualimgw - 3*\qualrowsep) {\includegraphics[width=\qualimgw]{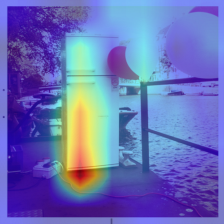}};
  \node[anchor=north west] at (\quallabelw + \quallabelsep + 4*\qualimgw + 4*\qualcolsep, -3*\qualimgw - 3*\qualrowsep) {\includegraphics[width=\qualimgw]{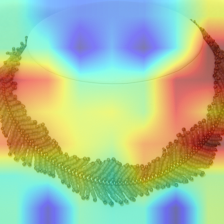}};
  \node[anchor=north west] at (\quallabelw + \quallabelsep + 5*\qualimgw + 5*\qualcolsep, -3*\qualimgw - 3*\qualrowsep) {\includegraphics[width=\qualimgw]{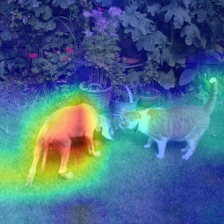}};
  \node[anchor=north west] at (\quallabelw + \quallabelsep + 6*\qualimgw + 6*\qualcolsep, -3*\qualimgw - 3*\qualrowsep) {\includegraphics[width=\qualimgw]{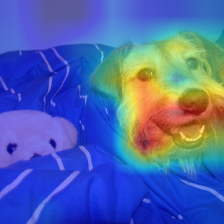}};
  \node[anchor=south, overlay, font={\fontsize{\qualfs}{\qualbls}\selectfont}, rotate=90] at (\quallabelw, -4*\qualimgw - 4*\qualrowsep - 0.5*\qualimgw) {MDA~\cite{expl:vit:mda:walker25}};
  \node[anchor=north west] at (\quallabelw + \quallabelsep + 0*\qualimgw + 0*\qualcolsep, -4*\qualimgw - 4*\qualrowsep) {\includegraphics[width=\qualimgw]{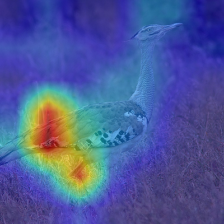}};
  \node[anchor=north west] at (\quallabelw + \quallabelsep + 1*\qualimgw + 1*\qualcolsep, -4*\qualimgw - 4*\qualrowsep) {\includegraphics[width=\qualimgw]{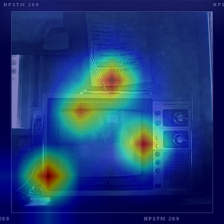}};
  \node[anchor=north west] at (\quallabelw + \quallabelsep + 2*\qualimgw + 2*\qualcolsep, -4*\qualimgw - 4*\qualrowsep) {\includegraphics[width=\qualimgw]{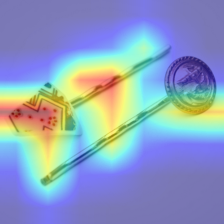}};
  \node[anchor=north west] at (\quallabelw + \quallabelsep + 3*\qualimgw + 3*\qualcolsep, -4*\qualimgw - 4*\qualrowsep) {\includegraphics[width=\qualimgw]{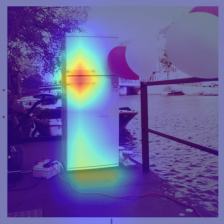}};
  \node[anchor=north west] at (\quallabelw + \quallabelsep + 4*\qualimgw + 4*\qualcolsep, -4*\qualimgw - 4*\qualrowsep) {\includegraphics[width=\qualimgw]{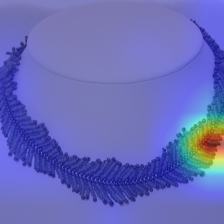}};
  \node[anchor=north west] at (\quallabelw + \quallabelsep + 5*\qualimgw + 5*\qualcolsep, -4*\qualimgw - 4*\qualrowsep) {\includegraphics[width=\qualimgw]{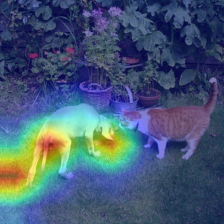}};
  \node[anchor=north west] at (\quallabelw + \quallabelsep + 6*\qualimgw + 6*\qualcolsep, -4*\qualimgw - 4*\qualrowsep) {\includegraphics[width=\qualimgw]{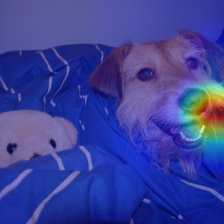}};
  \node[anchor=south, overlay, font={\fontsize{\qualfs}{\qualbls}\selectfont}, rotate=90] at (\quallabelw, -5*\qualimgw - 5*\qualrowsep - 0.5*\qualimgw) {LeGrad~\cite{expl:vit:legrad:bousselham25}};
  \node[anchor=north west] at (\quallabelw + \quallabelsep + 0*\qualimgw + 0*\qualcolsep, -5*\qualimgw - 5*\qualrowsep) {\includegraphics[width=\qualimgw]{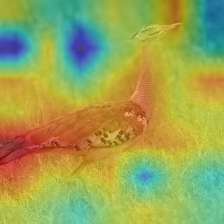}};
  \node[anchor=north west] at (\quallabelw + \quallabelsep + 1*\qualimgw + 1*\qualcolsep, -5*\qualimgw - 5*\qualrowsep) {\includegraphics[width=\qualimgw]{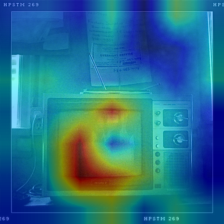}};
  \node[anchor=north west] at (\quallabelw + \quallabelsep + 2*\qualimgw + 2*\qualcolsep, -5*\qualimgw - 5*\qualrowsep) {\includegraphics[width=\qualimgw]{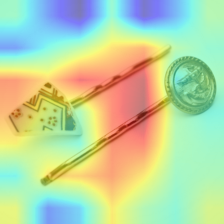}};
  \node[anchor=north west] at (\quallabelw + \quallabelsep + 3*\qualimgw + 3*\qualcolsep, -5*\qualimgw - 5*\qualrowsep) {\includegraphics[width=\qualimgw]{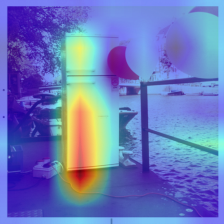}};
  \node[anchor=north west] at (\quallabelw + \quallabelsep + 4*\qualimgw + 4*\qualcolsep, -5*\qualimgw - 5*\qualrowsep) {\includegraphics[width=\qualimgw]{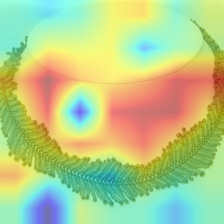}};
  \node[anchor=north west] at (\quallabelw + \quallabelsep + 5*\qualimgw + 5*\qualcolsep, -5*\qualimgw - 5*\qualrowsep) {\includegraphics[width=\qualimgw]{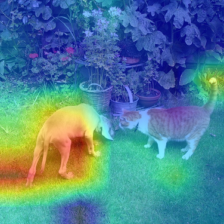}};
  \node[anchor=north west] at (\quallabelw + \quallabelsep + 6*\qualimgw + 6*\qualcolsep, -5*\qualimgw - 5*\qualrowsep) {\includegraphics[width=\qualimgw]{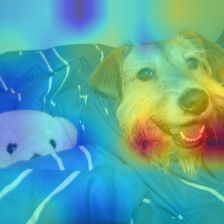}};
  \node[anchor=south, overlay, font={\fontsize{\qualfs}{\qualbls}\selectfont}, rotate=90] at (\quallabelw, -6*\qualimgw - 6*\qualrowsep - 0.5*\qualimgw) {TIS~\cite{expl:vit:tis:englebert23}};
  \node[anchor=north west] at (\quallabelw + \quallabelsep + 0*\qualimgw + 0*\qualcolsep, -6*\qualimgw - 6*\qualrowsep) {\includegraphics[width=\qualimgw]{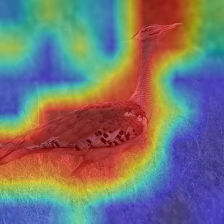}};
  \node[anchor=north west] at (\quallabelw + \quallabelsep + 1*\qualimgw + 1*\qualcolsep, -6*\qualimgw - 6*\qualrowsep) {\includegraphics[width=\qualimgw]{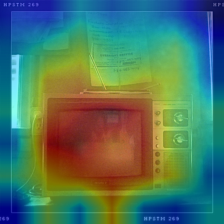}};
  \node[anchor=north west] at (\quallabelw + \quallabelsep + 2*\qualimgw + 2*\qualcolsep, -6*\qualimgw - 6*\qualrowsep) {\includegraphics[width=\qualimgw]{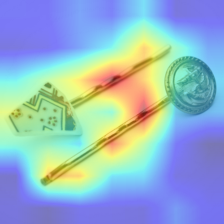}};
  \node[anchor=north west] at (\quallabelw + \quallabelsep + 3*\qualimgw + 3*\qualcolsep, -6*\qualimgw - 6*\qualrowsep) {\includegraphics[width=\qualimgw]{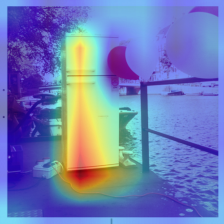}};
  \node[anchor=north west] at (\quallabelw + \quallabelsep + 4*\qualimgw + 4*\qualcolsep, -6*\qualimgw - 6*\qualrowsep) {\includegraphics[width=\qualimgw]{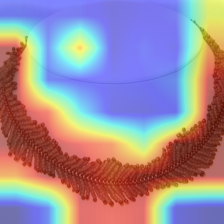}};
  \node[anchor=north west] at (\quallabelw + \quallabelsep + 5*\qualimgw + 5*\qualcolsep, -6*\qualimgw - 6*\qualrowsep) {\includegraphics[width=\qualimgw]{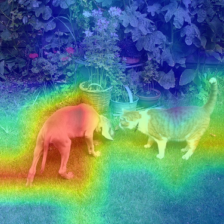}};
  \node[anchor=north west] at (\quallabelw + \quallabelsep + 6*\qualimgw + 6*\qualcolsep, -6*\qualimgw - 6*\qualrowsep) {\includegraphics[width=\qualimgw]{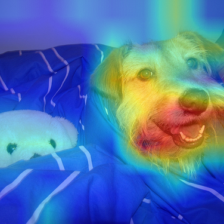}};
  \node[anchor=south, overlay, font={\fontsize{\qualfs}{\qualbls}\selectfont}, rotate=90] at (\quallabelw, -7*\qualimgw - 7*\qualrowsep - 0.5*\qualimgw) {ViT-CX~\cite{expl:vit:vitcx:xie23}};
  \node[anchor=north west] at (\quallabelw + \quallabelsep + 0*\qualimgw + 0*\qualcolsep, -7*\qualimgw - 7*\qualrowsep) {\includegraphics[width=\qualimgw]{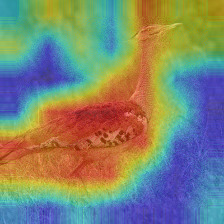}};
  \node[anchor=north west] at (\quallabelw + \quallabelsep + 1*\qualimgw + 1*\qualcolsep, -7*\qualimgw - 7*\qualrowsep) {\includegraphics[width=\qualimgw]{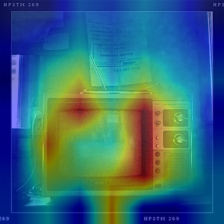}};
  \node[anchor=north west] at (\quallabelw + \quallabelsep + 2*\qualimgw + 2*\qualcolsep, -7*\qualimgw - 7*\qualrowsep) {\includegraphics[width=\qualimgw]{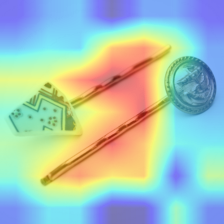}};
  \node[anchor=north west] at (\quallabelw + \quallabelsep + 3*\qualimgw + 3*\qualcolsep, -7*\qualimgw - 7*\qualrowsep) {\includegraphics[width=\qualimgw]{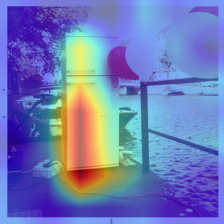}};
  \node[anchor=north west] at (\quallabelw + \quallabelsep + 4*\qualimgw + 4*\qualcolsep, -7*\qualimgw - 7*\qualrowsep) {\includegraphics[width=\qualimgw]{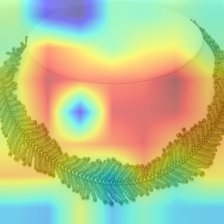}};
  \node[anchor=north west] at (\quallabelw + \quallabelsep + 5*\qualimgw + 5*\qualcolsep, -7*\qualimgw - 7*\qualrowsep) {\includegraphics[width=\qualimgw]{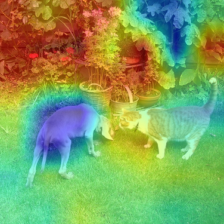}};
  \node[anchor=north west] at (\quallabelw + \quallabelsep + 6*\qualimgw + 6*\qualcolsep, -7*\qualimgw - 7*\qualrowsep) {\includegraphics[width=\qualimgw]{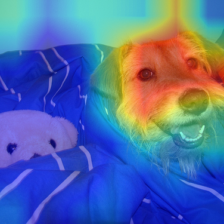}};
  \node[anchor=south, overlay, font={\fontsize{\qualfs}{\qualbls}\selectfont}, rotate=90] at (\quallabelw, -8*\qualimgw - 8*\qualrowsep - 0.5*\qualimgw) {Grad-CAM~\cite{expl:act:gradcam:selvaraju17}};
  \node[anchor=north west] at (\quallabelw + \quallabelsep + 0*\qualimgw + 0*\qualcolsep, -8*\qualimgw - 8*\qualrowsep) {\includegraphics[width=\qualimgw]{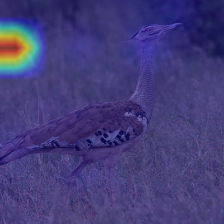}};
  \node[anchor=north west] at (\quallabelw + \quallabelsep + 1*\qualimgw + 1*\qualcolsep, -8*\qualimgw - 8*\qualrowsep) {\includegraphics[width=\qualimgw]{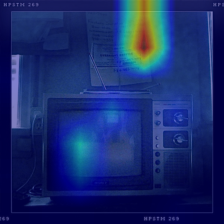}};
  \node[anchor=north west] at (\quallabelw + \quallabelsep + 2*\qualimgw + 2*\qualcolsep, -8*\qualimgw - 8*\qualrowsep) {\includegraphics[width=\qualimgw]{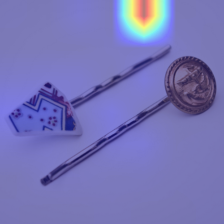}};
  \node[anchor=north west] at (\quallabelw + \quallabelsep + 3*\qualimgw + 3*\qualcolsep, -8*\qualimgw - 8*\qualrowsep) {\includegraphics[width=\qualimgw]{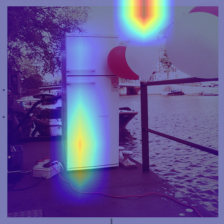}};
  \node[anchor=north west] at (\quallabelw + \quallabelsep + 4*\qualimgw + 4*\qualcolsep, -8*\qualimgw - 8*\qualrowsep) {\includegraphics[width=\qualimgw]{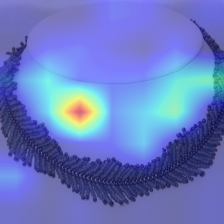}};
  \node[anchor=north west] at (\quallabelw + \quallabelsep + 5*\qualimgw + 5*\qualcolsep, -8*\qualimgw - 8*\qualrowsep) {\includegraphics[width=\qualimgw]{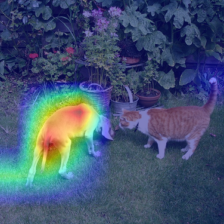}};
  \node[anchor=north west] at (\quallabelw + \quallabelsep + 6*\qualimgw + 6*\qualcolsep, -8*\qualimgw - 8*\qualrowsep) {\includegraphics[width=\qualimgw]{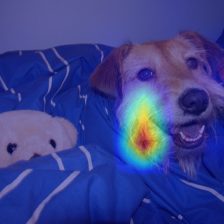}};
  \node[anchor=south, overlay, font={\fontsize{\qualfs}{\qualbls}\selectfont}, rotate=90] at (\quallabelw, -9*\qualimgw - 9*\qualrowsep - 0.5*\qualimgw) {IG~\cite{expl:grad:integratedgradients:sundararajan17}};
  \node[anchor=north west] at (\quallabelw + \quallabelsep + 0*\qualimgw + 0*\qualcolsep, -9*\qualimgw - 9*\qualrowsep) {\includegraphics[width=\qualimgw]{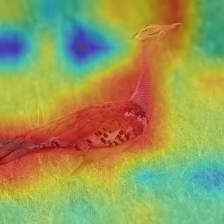}};
  \node[anchor=north west] at (\quallabelw + \quallabelsep + 1*\qualimgw + 1*\qualcolsep, -9*\qualimgw - 9*\qualrowsep) {\includegraphics[width=\qualimgw]{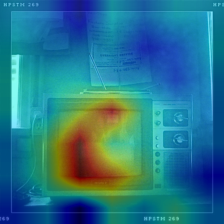}};
  \node[anchor=north west] at (\quallabelw + \quallabelsep + 2*\qualimgw + 2*\qualcolsep, -9*\qualimgw - 9*\qualrowsep) {\includegraphics[width=\qualimgw]{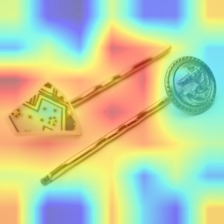}};
  \node[anchor=north west] at (\quallabelw + \quallabelsep + 3*\qualimgw + 3*\qualcolsep, -9*\qualimgw - 9*\qualrowsep) {\includegraphics[width=\qualimgw]{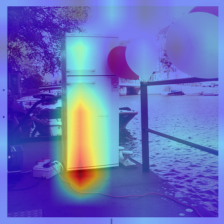}};
  \node[anchor=north west] at (\quallabelw + \quallabelsep + 4*\qualimgw + 4*\qualcolsep, -9*\qualimgw - 9*\qualrowsep) {\includegraphics[width=\qualimgw]{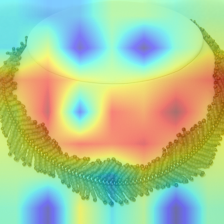}};
  \node[anchor=north west] at (\quallabelw + \quallabelsep + 5*\qualimgw + 5*\qualcolsep, -9*\qualimgw - 9*\qualrowsep) {\includegraphics[width=\qualimgw]{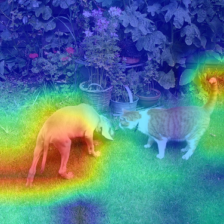}};
  \node[anchor=north west] at (\quallabelw + \quallabelsep + 6*\qualimgw + 6*\qualcolsep, -9*\qualimgw - 9*\qualrowsep) {\includegraphics[width=\qualimgw]{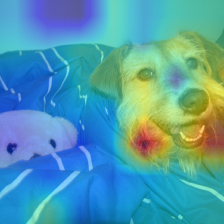}};
  \node[anchor=south, overlay, font={\fontsize{\qualfs}{\qualbls}\selectfont}, rotate=90] at (\quallabelw, -10*\qualimgw - 10*\qualrowsep - 0.5*\qualimgw) {Trans-Att.~\cite{expl:vit:transatt:yuan21}};
  \node[anchor=north west] at (\quallabelw + \quallabelsep + 0*\qualimgw + 0*\qualcolsep, -10*\qualimgw - 10*\qualrowsep) {\includegraphics[width=\qualimgw]{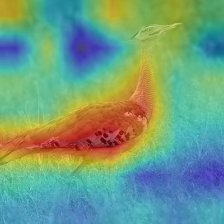}};
  \node[anchor=north west] at (\quallabelw + \quallabelsep + 1*\qualimgw + 1*\qualcolsep, -10*\qualimgw - 10*\qualrowsep) {\includegraphics[width=\qualimgw]{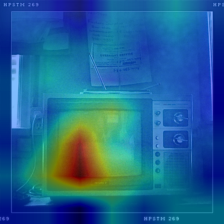}};
  \node[anchor=north west] at (\quallabelw + \quallabelsep + 2*\qualimgw + 2*\qualcolsep, -10*\qualimgw - 10*\qualrowsep) {\includegraphics[width=\qualimgw]{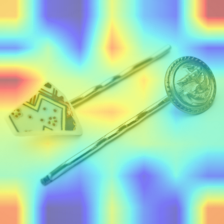}};
  \node[anchor=north west] at (\quallabelw + \quallabelsep + 3*\qualimgw + 3*\qualcolsep, -10*\qualimgw - 10*\qualrowsep) {\includegraphics[width=\qualimgw]{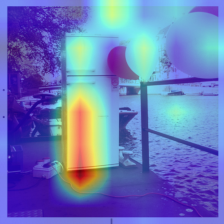}};
  \node[anchor=north west] at (\quallabelw + \quallabelsep + 4*\qualimgw + 4*\qualcolsep, -10*\qualimgw - 10*\qualrowsep) {\includegraphics[width=\qualimgw]{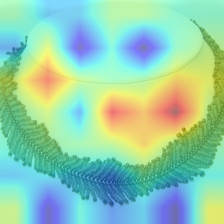}};
  \node[anchor=north west] at (\quallabelw + \quallabelsep + 5*\qualimgw + 5*\qualcolsep, -10*\qualimgw - 10*\qualrowsep) {\includegraphics[width=\qualimgw]{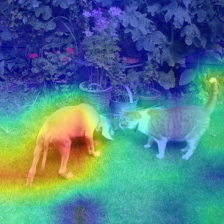}};
  \node[anchor=north west] at (\quallabelw + \quallabelsep + 6*\qualimgw + 6*\qualcolsep, -10*\qualimgw - 10*\qualrowsep) {\includegraphics[width=\qualimgw]{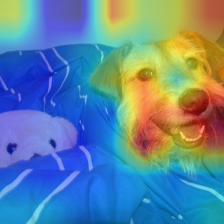}};
\end{tikzpicture}

%% file: figures/qual_convnext_small/fig.tex
\begin{figure*}[tb]
    \centering
    \resizebox{\textwidth}{!}{
        \input{figures/qual_convnext_small/source.tex}
    }
    \caption{
        \textbf{Qualitative comparison of attribution maps} for a ConvNeXt Small classifier~\cite{model:convnext:liu22} on ImageNet~\cite{dataset:imagenet:deng09} samples.
        The \textcolor{green!70!black}{green} / \textcolor{red!80!black}{red} labels mark correct / incorrect predictions (ground truth in parentheses).
        }
        \label{fig:qual_convnext_small}
\end{figure*}
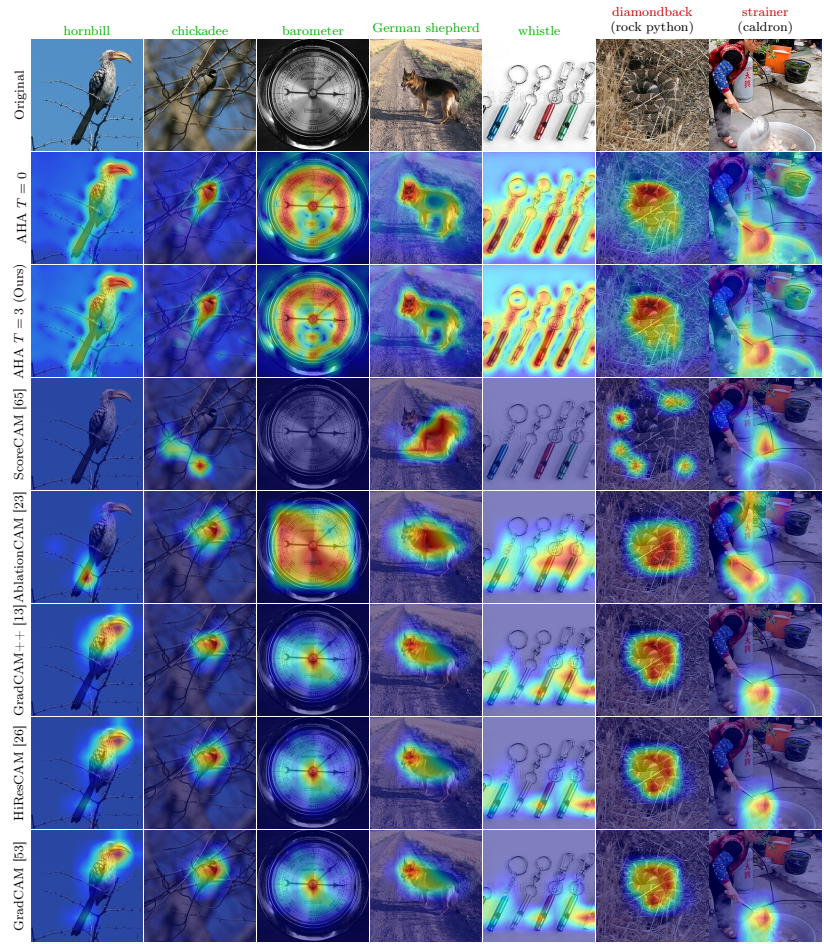

%% file: figures/qual_convnext_small/source.tex
\makeatletter
\@ifundefined{qualcolsep}{\newlength{\qualcolsep}}{}
\setlength{\qualcolsep}{1pt}
\@ifundefined{qualrowsep}{\newlength{\qualrowsep}}{}
\setlength{\qualrowsep}{1pt}
\@ifundefined{qualhdrsep}{\newlength{\qualhdrsep}}{}
\setlength{\qualhdrsep}{3pt}
\@ifundefined{quallabelsep}{\newlength{\quallabelsep}}{}
\setlength{\quallabelsep}{3pt}
\@ifundefined{quallabelw}{\newlength{\quallabelw}}{}
\setlength{\quallabelw}{0.35cm}
\@ifundefined{qualimgw}{\newlength{\qualimgw}}{}
\setlength{\qualimgw}{2.4000cm}
\makeatother
\pgfmathsetlengthmacro{\qualfs}{0.115*\qualimgw}
\pgfmathsetlengthmacro{\qualbls}{1.2*\qualfs}

\begin{tikzpicture}[
    every node/.style={inner sep=0pt, outer sep=0pt},
]
  \hypersetup{hidelinks}
  \path[use as bounding box] (0, \qualhdrsep + 2.4*\qualbls) rectangle (\quallabelw + \quallabelsep + 7*\qualimgw + 6*\qualcolsep + \quallabelw + \quallabelsep, -8*\qualimgw - 7*\qualrowsep);
  \node[anchor=south, font={\fontsize{\qualfs}{\qualbls}\selectfont}, text width=\qualimgw, align=center] at (\quallabelw + \quallabelsep + 0*\qualimgw + 0*\qualcolsep + 0.5*\qualimgw, \qualhdrsep) {\textcolor{green!70!black}{hornbill}};
  \node[anchor=south, font={\fontsize{\qualfs}{\qualbls}\selectfont}, text width=\qualimgw, align=center] at (\quallabelw + \quallabelsep + 1*\qualimgw + 1*\qualcolsep + 0.5*\qualimgw, \qualhdrsep) {\textcolor{green!70!black}{chickadee}};
  \node[anchor=south, font={\fontsize{\qualfs}{\qualbls}\selectfont}, text width=\qualimgw, align=center] at (\quallabelw + \quallabelsep + 2*\qualimgw + 2*\qualcolsep + 0.5*\qualimgw, \qualhdrsep) {\textcolor{green!70!black}{barometer}};
  \node[anchor=south, font={\fontsize{\qualfs}{\qualbls}\selectfont}, text width=\qualimgw, align=center] at (\quallabelw + \quallabelsep + 3*\qualimgw + 3*\qualcolsep + 0.5*\qualimgw, \qualhdrsep) {\textcolor{green!70!black}{German shepherd}};
  \node[anchor=south, font={\fontsize{\qualfs}{\qualbls}\selectfont}, text width=\qualimgw, align=center] at (\quallabelw + \quallabelsep + 4*\qualimgw + 4*\qualcolsep + 0.5*\qualimgw, \qualhdrsep) {\textcolor{green!70!black}{whistle}};
  \node[anchor=south, font={\fontsize{\qualfs}{\qualbls}\selectfont}, text width=\qualimgw, align=center] at (\quallabelw + \quallabelsep + 5*\qualimgw + 5*\qualcolsep + 0.5*\qualimgw, \qualhdrsep) {\textcolor{red!80!black}{diamondback} \\ (rock python)};
  \node[anchor=south, font={\fontsize{\qualfs}{\qualbls}\selectfont}, text width=\qualimgw, align=center] at (\quallabelw + \quallabelsep + 6*\qualimgw + 6*\qualcolsep + 0.5*\qualimgw, \qualhdrsep) {\textcolor{red!80!black}{strainer} \\ (caldron)};
  \node[anchor=south, overlay, font={\fontsize{\qualfs}{\qualbls}\selectfont}, rotate=90] at (\quallabelw, -0*\qualimgw - 0*\qualrowsep - 0.5*\qualimgw) {Original};
  \node[anchor=north west] at (\quallabelw + \quallabelsep + 0*\qualimgw + 0*\qualcolsep, -0*\qualimgw - 0*\qualrowsep) {\includegraphics[width=\qualimgw]{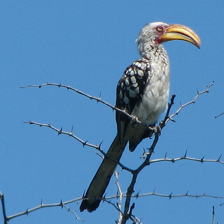}};
  \node[anchor=north west] at (\quallabelw + \quallabelsep + 1*\qualimgw + 1*\qualcolsep, -0*\qualimgw - 0*\qualrowsep) {\includegraphics[width=\qualimgw]{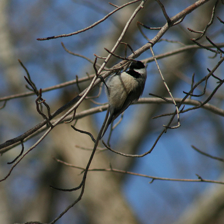}};
  \node[anchor=north west] at (\quallabelw + \quallabelsep + 2*\qualimgw + 2*\qualcolsep, -0*\qualimgw - 0*\qualrowsep) {\includegraphics[width=\qualimgw]{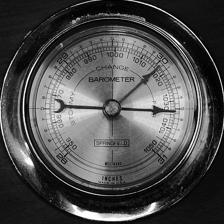}};
  \node[anchor=north west] at (\quallabelw + \quallabelsep + 3*\qualimgw + 3*\qualcolsep, -0*\qualimgw - 0*\qualrowsep) {\includegraphics[width=\qualimgw]{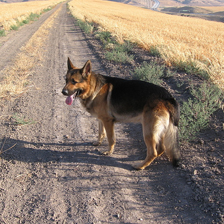}};
  \node[anchor=north west] at (\quallabelw + \quallabelsep + 4*\qualimgw + 4*\qualcolsep, -0*\qualimgw - 0*\qualrowsep) {\includegraphics[width=\qualimgw]{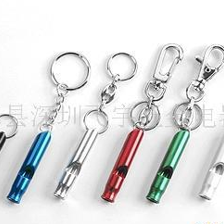}};
  \node[anchor=north west] at (\quallabelw + \quallabelsep + 5*\qualimgw + 5*\qualcolsep, -0*\qualimgw - 0*\qualrowsep) {\includegraphics[width=\qualimgw]{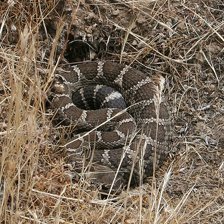}};
  \node[anchor=north west] at (\quallabelw + \quallabelsep + 6*\qualimgw + 6*\qualcolsep, -0*\qualimgw - 0*\qualrowsep) {\includegraphics[width=\qualimgw]{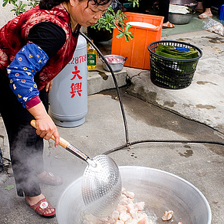}};
  \node[anchor=south, overlay, font={\fontsize{\qualfs}{\qualbls}\selectfont}, rotate=90] at (\quallabelw, -1*\qualimgw - 1*\qualrowsep - 0.5*\qualimgw) {\methodabbrev $T=0$};
  \node[anchor=north west] at (\quallabelw + \quallabelsep + 0*\qualimgw + 0*\qualcolsep, -1*\qualimgw - 1*\qualrowsep) {\includegraphics[width=\qualimgw]{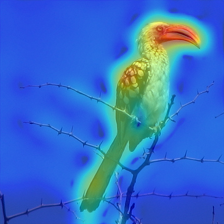}};
  \node[anchor=north west] at (\quallabelw + \quallabelsep + 1*\qualimgw + 1*\qualcolsep, -1*\qualimgw - 1*\qualrowsep) {\includegraphics[width=\qualimgw]{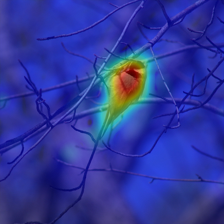}};
  \node[anchor=north west] at (\quallabelw + \quallabelsep + 2*\qualimgw + 2*\qualcolsep, -1*\qualimgw - 1*\qualrowsep) {\includegraphics[width=\qualimgw]{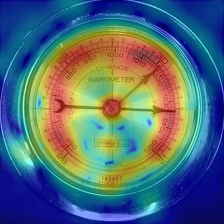}};
  \node[anchor=north west] at (\quallabelw + \quallabelsep + 3*\qualimgw + 3*\qualcolsep, -1*\qualimgw - 1*\qualrowsep) {\includegraphics[width=\qualimgw]{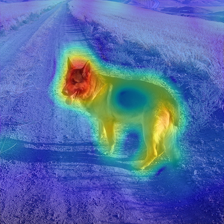}};
  \node[anchor=north west] at (\quallabelw + \quallabelsep + 4*\qualimgw + 4*\qualcolsep, -1*\qualimgw - 1*\qualrowsep) {\includegraphics[width=\qualimgw]{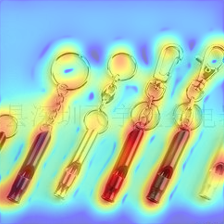}};
  \node[anchor=north west] at (\quallabelw + \quallabelsep + 5*\qualimgw + 5*\qualcolsep, -1*\qualimgw - 1*\qualrowsep) {\includegraphics[width=\qualimgw]{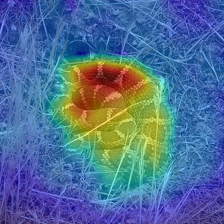}};
  \node[anchor=north west] at (\quallabelw + \quallabelsep + 6*\qualimgw + 6*\qualcolsep, -1*\qualimgw - 1*\qualrowsep) {\includegraphics[width=\qualimgw]{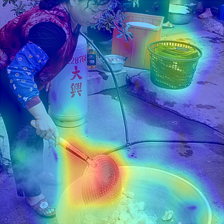}};
  \node[anchor=south, overlay, font={\fontsize{\qualfs}{\qualbls}\selectfont}, rotate=90] at (\quallabelw, -2*\qualimgw - 2*\qualrowsep - 0.5*\qualimgw) {\methodabbrev $T=3$ (Ours)};
  \node[anchor=north west] at (\quallabelw + \quallabelsep + 0*\qualimgw + 0*\qualcolsep, -2*\qualimgw - 2*\qualrowsep) {\includegraphics[width=\qualimgw]{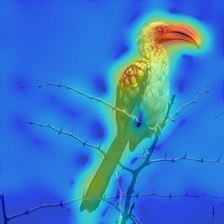}};
  \node[anchor=north west] at (\quallabelw + \quallabelsep + 1*\qualimgw + 1*\qualcolsep, -2*\qualimgw - 2*\qualrowsep) {\includegraphics[width=\qualimgw]{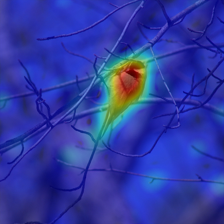}};
  \node[anchor=north west] at (\quallabelw + \quallabelsep + 2*\qualimgw + 2*\qualcolsep, -2*\qualimgw - 2*\qualrowsep) {\includegraphics[width=\qualimgw]{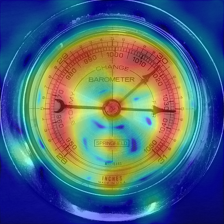}};
  \node[anchor=north west] at (\quallabelw + \quallabelsep + 3*\qualimgw + 3*\qualcolsep, -2*\qualimgw - 2*\qualrowsep) {\includegraphics[width=\qualimgw]{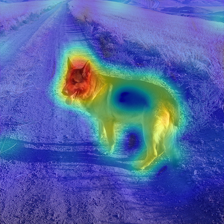}};
  \node[anchor=north west] at (\quallabelw + \quallabelsep + 4*\qualimgw + 4*\qualcolsep, -2*\qualimgw - 2*\qualrowsep) {\includegraphics[width=\qualimgw]{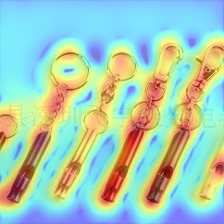}};
  \node[anchor=north west] at (\quallabelw + \quallabelsep + 5*\qualimgw + 5*\qualcolsep, -2*\qualimgw - 2*\qualrowsep) {\includegraphics[width=\qualimgw]{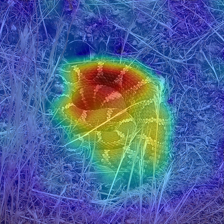}};
  \node[anchor=north west] at (\quallabelw + \quallabelsep + 6*\qualimgw + 6*\qualcolsep, -2*\qualimgw - 2*\qualrowsep) {\includegraphics[width=\qualimgw]{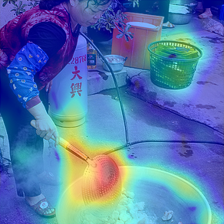}};
  \node[anchor=south, overlay, font={\fontsize{\qualfs}{\qualbls}\selectfont}, rotate=90] at (\quallabelw, -3*\qualimgw - 3*\qualrowsep - 0.5*\qualimgw) {ScoreCAM~\cite{expl:act:scorecam:wang20}};
  \node[anchor=north west] at (\quallabelw + \quallabelsep + 0*\qualimgw + 0*\qualcolsep, -3*\qualimgw - 3*\qualrowsep) {\includegraphics[width=\qualimgw]{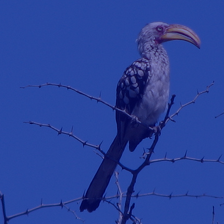}};
  \node[anchor=north west] at (\quallabelw + \quallabelsep + 1*\qualimgw + 1*\qualcolsep, -3*\qualimgw - 3*\qualrowsep) {\includegraphics[width=\qualimgw]{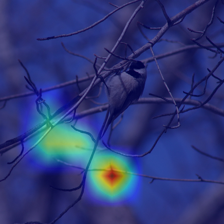}};
  \node[anchor=north west] at (\quallabelw + \quallabelsep + 2*\qualimgw + 2*\qualcolsep, -3*\qualimgw - 3*\qualrowsep) {\includegraphics[width=\qualimgw]{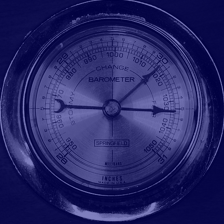}};
  \node[anchor=north west] at (\quallabelw + \quallabelsep + 3*\qualimgw + 3*\qualcolsep, -3*\qualimgw - 3*\qualrowsep) {\includegraphics[width=\qualimgw]{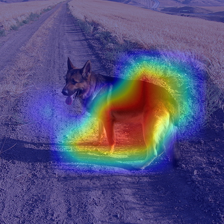}};
  \node[anchor=north west] at (\quallabelw + \quallabelsep + 4*\qualimgw + 4*\qualcolsep, -3*\qualimgw - 3*\qualrowsep) {\includegraphics[width=\qualimgw]{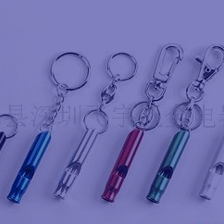}};
  \node[anchor=north west] at (\quallabelw + \quallabelsep + 5*\qualimgw + 5*\qualcolsep, -3*\qualimgw - 3*\qualrowsep) {\includegraphics[width=\qualimgw]{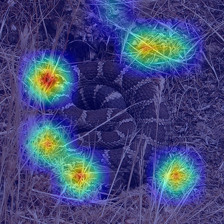}};
  \node[anchor=north west] at (\quallabelw + \quallabelsep + 6*\qualimgw + 6*\qualcolsep, -3*\qualimgw - 3*\qualrowsep) {\includegraphics[width=\qualimgw]{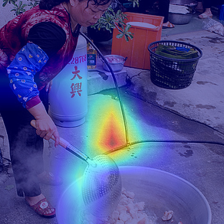}};
  \node[anchor=south, overlay, font={\fontsize{\qualfs}{\qualbls}\selectfont}, rotate=90] at (\quallabelw, -4*\qualimgw - 4*\qualrowsep - 0.5*\qualimgw) {AblationCAM~\cite{expl:act:ablationcam:desai20}};
  \node[anchor=north west] at (\quallabelw + \quallabelsep + 0*\qualimgw + 0*\qualcolsep, -4*\qualimgw - 4*\qualrowsep) {\includegraphics[width=\qualimgw]{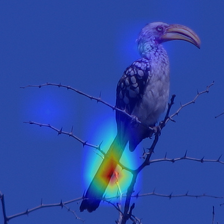}};
  \node[anchor=north west] at (\quallabelw + \quallabelsep + 1*\qualimgw + 1*\qualcolsep, -4*\qualimgw - 4*\qualrowsep) {\includegraphics[width=\qualimgw]{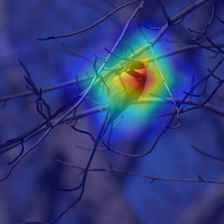}};
  \node[anchor=north west] at (\quallabelw + \quallabelsep + 2*\qualimgw + 2*\qualcolsep, -4*\qualimgw - 4*\qualrowsep) {\includegraphics[width=\qualimgw]{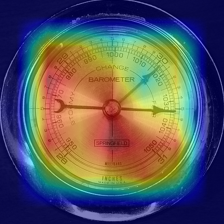}};
  \node[anchor=north west] at (\quallabelw + \quallabelsep + 3*\qualimgw + 3*\qualcolsep, -4*\qualimgw - 4*\qualrowsep) {\includegraphics[width=\qualimgw]{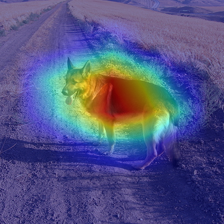}};
  \node[anchor=north west] at (\quallabelw + \quallabelsep + 4*\qualimgw + 4*\qualcolsep, -4*\qualimgw - 4*\qualrowsep) {\includegraphics[width=\qualimgw]{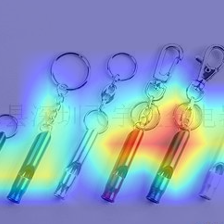}};
  \node[anchor=north west] at (\quallabelw + \quallabelsep + 5*\qualimgw + 5*\qualcolsep, -4*\qualimgw - 4*\qualrowsep) {\includegraphics[width=\qualimgw]{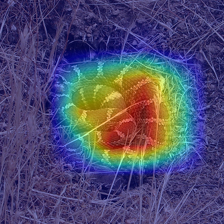}};
  \node[anchor=north west] at (\quallabelw + \quallabelsep + 6*\qualimgw + 6*\qualcolsep, -4*\qualimgw - 4*\qualrowsep) {\includegraphics[width=\qualimgw]{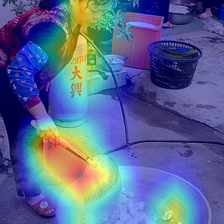}};
  \node[anchor=south, overlay, font={\fontsize{\qualfs}{\qualbls}\selectfont}, rotate=90] at (\quallabelw, -5*\qualimgw - 5*\qualrowsep - 0.5*\qualimgw) {GradCAM++~\cite{expl:act:gradcampp:chattopadhay18}};
  \node[anchor=north west] at (\quallabelw + \quallabelsep + 0*\qualimgw + 0*\qualcolsep, -5*\qualimgw - 5*\qualrowsep) {\includegraphics[width=\qualimgw]{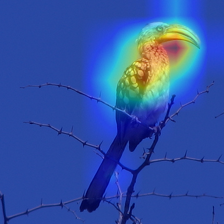}};
  \node[anchor=north west] at (\quallabelw + \quallabelsep + 1*\qualimgw + 1*\qualcolsep, -5*\qualimgw - 5*\qualrowsep) {\includegraphics[width=\qualimgw]{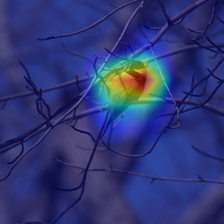}};
  \node[anchor=north west] at (\quallabelw + \quallabelsep + 2*\qualimgw + 2*\qualcolsep, -5*\qualimgw - 5*\qualrowsep) {\includegraphics[width=\qualimgw]{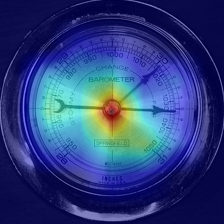}};
  \node[anchor=north west] at (\quallabelw + \quallabelsep + 3*\qualimgw + 3*\qualcolsep, -5*\qualimgw - 5*\qualrowsep) {\includegraphics[width=\qualimgw]{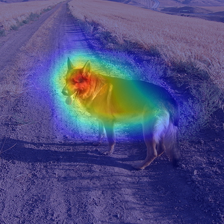}};
  \node[anchor=north west] at (\quallabelw + \quallabelsep + 4*\qualimgw + 4*\qualcolsep, -5*\qualimgw - 5*\qualrowsep) {\includegraphics[width=\qualimgw]{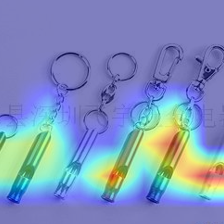}};
  \node[anchor=north west] at (\quallabelw + \quallabelsep + 5*\qualimgw + 5*\qualcolsep, -5*\qualimgw - 5*\qualrowsep) {\includegraphics[width=\qualimgw]{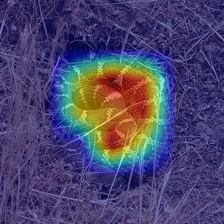}};
  \node[anchor=north west] at (\quallabelw + \quallabelsep + 6*\qualimgw + 6*\qualcolsep, -5*\qualimgw - 5*\qualrowsep) {\includegraphics[width=\qualimgw]{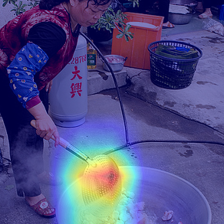}};
  \node[anchor=south, overlay, font={\fontsize{\qualfs}{\qualbls}\selectfont}, rotate=90] at (\quallabelw, -6*\qualimgw - 6*\qualrowsep - 0.5*\qualimgw) {HiResCAM~\cite{expl:act:hirescam:draelos20}};
  \node[anchor=north west] at (\quallabelw + \quallabelsep + 0*\qualimgw + 0*\qualcolsep, -6*\qualimgw - 6*\qualrowsep) {\includegraphics[width=\qualimgw]{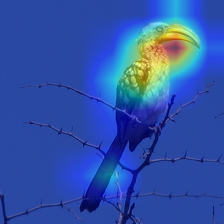}};
  \node[anchor=north west] at (\quallabelw + \quallabelsep + 1*\qualimgw + 1*\qualcolsep, -6*\qualimgw - 6*\qualrowsep) {\includegraphics[width=\qualimgw]{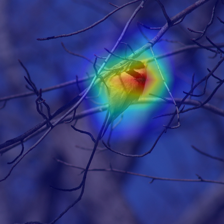}};
  \node[anchor=north west] at (\quallabelw + \quallabelsep + 2*\qualimgw + 2*\qualcolsep, -6*\qualimgw - 6*\qualrowsep) {\includegraphics[width=\qualimgw]{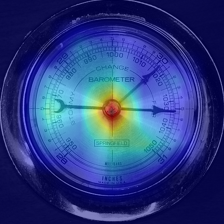}};
  \node[anchor=north west] at (\quallabelw + \quallabelsep + 3*\qualimgw + 3*\qualcolsep, -6*\qualimgw - 6*\qualrowsep) {\includegraphics[width=\qualimgw]{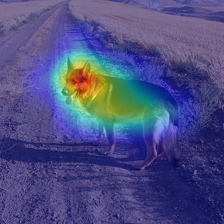}};
  \node[anchor=north west] at (\quallabelw + \quallabelsep + 4*\qualimgw + 4*\qualcolsep, -6*\qualimgw - 6*\qualrowsep) {\includegraphics[width=\qualimgw]{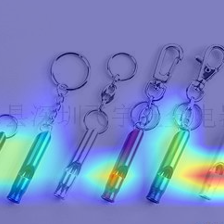}};
  \node[anchor=north west] at (\quallabelw + \quallabelsep + 5*\qualimgw + 5*\qualcolsep, -6*\qualimgw - 6*\qualrowsep) {\includegraphics[width=\qualimgw]{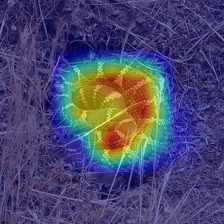}};
  \node[anchor=north west] at (\quallabelw + \quallabelsep + 6*\qualimgw + 6*\qualcolsep, -6*\qualimgw - 6*\qualrowsep) {\includegraphics[width=\qualimgw]{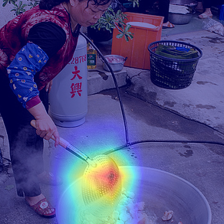}};
  \node[anchor=south, overlay, font={\fontsize{\qualfs}{\qualbls}\selectfont}, rotate=90] at (\quallabelw, -7*\qualimgw - 7*\qualrowsep - 0.5*\qualimgw) {GradCAM~\cite{expl:act:gradcam:selvaraju17}};
  \node[anchor=north west] at (\quallabelw + \quallabelsep + 0*\qualimgw + 0*\qualcolsep, -7*\qualimgw - 7*\qualrowsep) {\includegraphics[width=\qualimgw]{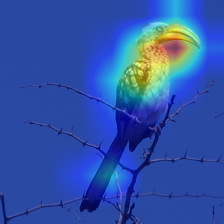}};
  \node[anchor=north west] at (\quallabelw + \quallabelsep + 1*\qualimgw + 1*\qualcolsep, -7*\qualimgw - 7*\qualrowsep) {\includegraphics[width=\qualimgw]{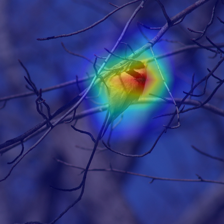}};
  \node[anchor=north west] at (\quallabelw + \quallabelsep + 2*\qualimgw + 2*\qualcolsep, -7*\qualimgw - 7*\qualrowsep) {\includegraphics[width=\qualimgw]{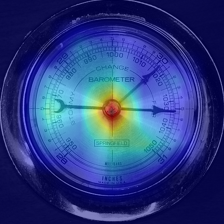}};
  \node[anchor=north west] at (\quallabelw + \quallabelsep + 3*\qualimgw + 3*\qualcolsep, -7*\qualimgw - 7*\qualrowsep) {\includegraphics[width=\qualimgw]{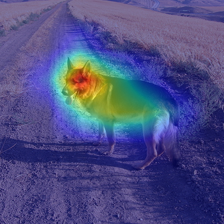}};
  \node[anchor=north west] at (\quallabelw + \quallabelsep + 4*\qualimgw + 4*\qualcolsep, -7*\qualimgw - 7*\qualrowsep) {\includegraphics[width=\qualimgw]{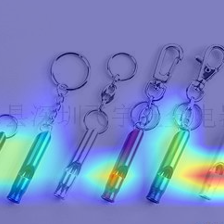}};
  \node[anchor=north west] at (\quallabelw + \quallabelsep + 5*\qualimgw + 5*\qualcolsep, -7*\qualimgw - 7*\qualrowsep) {\includegraphics[width=\qualimgw]{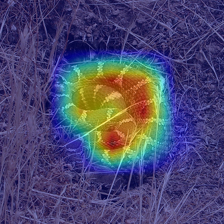}};
  \node[anchor=north west] at (\quallabelw + \quallabelsep + 6*\qualimgw + 6*\qualcolsep, -7*\qualimgw - 7*\qualrowsep) {\includegraphics[width=\qualimgw]{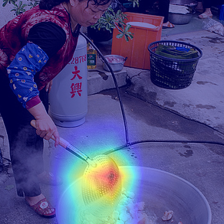}};
\end{tikzpicture}